\RequirePackage{fix-cm}
\documentclass[twocolumn]{svjour3}     

\smartqed  

\usepackage[latin1]{inputenc}
\usepackage[authoryear]{natbib}
\usepackage{graphicx}
\usepackage{epsfig}
\usepackage{subfigure}
\usepackage[active]{srcltx}
\usepackage{amsfonts,amsmath,amssymb}
\usepackage{color}
\usepackage{enumitem}
\usepackage{pifont}
\usepackage{appendix}
\usepackage{hyperref}

\usepackage[lined,ruled,linesnumbered]{algorithm2e}
\usepackage{rotating}
\usepackage{afterpage}
\usepackage{adjustbox}

\usepackage{bm}

\DeclareMathOperator*{\argmin}{\arg\,\min}

\newcommand{\bS}{\boldsymbol{S}}
\newcommand{\bA}{\boldsymbol{A}}
\newcommand{\bs}{\boldsymbol{s}}

\newcommand{\bR}{\boldsymbol{R}}

\newcommand{\bt}{\boldsymbol{t}}
\newcommand{\bI}{\boldsymbol{I}}
\newcommand{\nrsfm}[0]{{{\sc nrs}{\rm\em f}{\sc m}}}
\newcommand{\sfs}[0]{{\sc s{\rm\em f}\sc s}}
\newcommand{\sfm}[0]{{\sc s{\rm\em f}\sc m}}

\newcommand{\mvs}[0]{{\sc mvs}}

\usepackage{array}
\newcolumntype{L}[1]{>{\raggedright\let\newline\\\arraybackslash\hspace{0pt}}m{#1}}
\newcolumntype{C}[1]{>{\centering\let\newline\\\arraybackslash\hspace{0pt}}m{#1}}
\newcolumntype{R}[1]{>{\raggedleft\let\newline\\\arraybackslash\hspace{0pt}}m{#1}}

\newcommand*{\SHOWCOMMENTS}{}
\newcommand{\etal}{{\em et al.}}
\ifdefined\SHOWCOMMENTS
\newcommand{\notee}[3]{{\color{#2}\textbf{{\sc #1}: #3}}}
\newcommand{\CHRIS}[1]{\notee{Chris}{red}{#1}}
\newcommand{\LOURDES}[1]{\notee{Lourdes}{blue}{#1}}
\newcommand{\QI}[1]{\notee{Qi}{green}{#1}}

\else
\newcommand{\notee}[1]{}
\newcommand{\CHRIS}[1]{}
\newcommand{\QI}[1]{}
\newcommand{\LOURDES}[1]{}
\fi

\usepackage{times}
\usepackage{upgreek}
\usepackage{mathrsfs}
\usepackage{balance}

\usepackage{color}

\DeclareMathAlphabet{\mathpzc}{OT1}{pzc}{m}{it}

\journalname{International Journal of Computer Vision}
\begin{document}

\title{Better Together: Joint Reasoning for Non-rigid 3D Reconstruction with Specularities and Shading}

\titlerunning{Joint Reasoning for Non-rigid 3D Reconstruction}        

\author{Qi Liu-Yin* \and Rui Yu* \and Lourdes Agapito \and Andrew Fitzgibbon \and
  Chris Russell}
\authorrunning{Qi Liu-Yin and Rui Yu and Lourdes Agapito and Andrew Fitzgibbon and
  Chris Russell}

\institute{
  * The first two authors contribute equally.
  \\\\
  Qi Liu-Yin \and Rui Yu \and Lourdes Agapito \at
  University College London, London, UK \\
  \email{\{Qi.Liu,R.Yu,L.Agapito\}@cs.ucl.ac.uk} \\ \\
    Andrew Fitzgibbon \at
    Microsoft Research Cambridge, Cambridge, UK \\
    \email{awf@microsoft.com} \\ \\
    Chris Russell \at
    University of Surrey \\
    and\\
    Alan Turing Institute, London, UK \\
    \email{cr0040@surrey.ac.uk}
  }

\date{Received: date / Accepted: date}

\maketitle
\begin{abstract}
 We demonstrate the use of shape-from-shading (\sfs) to improve
both the quality and the robustness of 3D reconstruction of dynamic objects
captured by a single camera. Unlike previous approaches that made use of
\sfs~as a post-processing step, we offer a principled integrated approach that
solves dynamic object tracking and reconstruction and \sfs~as a single unified
cost function.
Moving beyond Lambertian \sfs, we propose a general approach that models both
specularities and shading while simultaneously tracking and reconstructing
general dynamic objects. Solving these problems jointly prevents the kinds of
tracking failures which can not be recovered from by pipeline approaches. We
show state-of-the-art results both qualitatively and quantitatively.

\end{abstract}




\section{Introduction}

\begin{figure*}[t!]
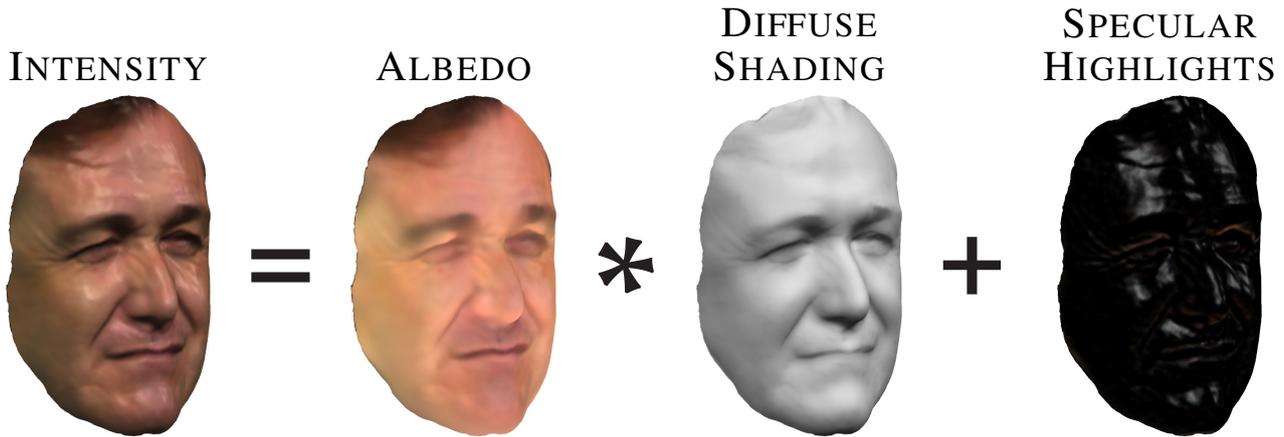

	\centering
	\bgroup
	\setlength\tabcolsep{4pt}
\resizebox{\linewidth}{!}{
	\begin{tabular}{ccccccc}
		\sc{Intensity} && \sc{Albedo} && \shortstack{\sc{Diffuse} \\ \sc{Shading}} && \shortstack{\sc{Specular}\\ \sc{Highlights}}\\
		\adjustimage{width=0.17\columnwidth, valign=m}{intensity000-trim}&
		\adjustimage{width=0.05\columnwidth, valign=m}{bold_equal}&
		\adjustimage{width=0.17\columnwidth, valign=m}{albedo000-trim}&
		\adjustimage{width=0.05\columnwidth, valign=m}{bold_mult}&
		\adjustimage{width=0.17\columnwidth, valign=m}{shading000-trim}&
		\adjustimage{width=0.05\columnwidth, valign=m}{bold_sum}&
		\adjustimage{width=0.17\columnwidth, valign=m}{specular000-trim}
		\\
		\multicolumn{7}{c}{}
	\end{tabular}}
	\egroup
	\caption{\label{fig:decomposition} Intensity decomposition into the product of
    albedo and diffuse shading (as a function of spherical harmonics and 3D
    shape estimation) plus the specular component.
	}
\end{figure*}

As the quality of 3D reconstructions of dynamic and deformable objects 
such as animals and faces has improved, robustness and
the reconstruction of semantically meaningful details like smile and frown
lines become more important. These transient fine details can not be recovered
by tracking alone, and require an understanding of the lighting in the environment
and a knowledge of how the surface normals of the object affect its
illumination.

While these shading artifacts can inform highly-detailed reconstructions, they
can also prevent the tracking of objects. In homogeneously textured regions,
such as human skin, the variance in the appearance of a patch due to
lighting changes can be much greater than the difference in appearance between
one patch and the next. A combination of these effects makes it vital that we
model illumination changes if we wish to correctly capture facial
deformations particularly those of the brow and cheeks.

The instability of color as a tracking cue is well known and much
remarked upon in the literature. Focusing on recent works in dynamic 3D
reconstruction using depth or multi-camera capture, it is noticeable how papers
such as~\cite{Dou_2015_CVPR,Newcombe:etal:CVPR2015} make use of raw depth maps
without color information in reconstruction.
Similarly, although the RGB-D based
work~\cite{kinect_nr:siggraph2014} made use of color information they
only matched appearance between pairs of adjacent frames as over long
sequences changes in shadow and illumination made color matching unreliable.
These problems can largely be ignored in the reconstruction of rigid scenes that
do not move relative to the lighting environment. Here shading
artifacts remain constant throughout the sequence, while specularities
typically occur sparsely and can be handled without being explicitly modeled
through the use of robust statistics~\citep{Newcombe:etal:ICCV2011}.

In the field of non-rigid monocular reconstruction from RGB video, we are not so
fortunate. With only a single RGB camera as input, we must make use of
color information.  However, without depth information, matching
color only between pairs of frames is prone to drift, with many
tracks gradually diffusing away from an object over long
sequences~\citep{Sundaram:etal:ECCV2010}. Similarly, moving objects can
no longer be assumed to be static with respect to the lighting
environment, and outside of a controlled studio-lit environment, changes
in the orientation of objects lead to significant changes in
appearance. Such changes often lead to the failure of direct image
intensity based trackers such as~\cite{Yu_2015_ICCV}.

Building from cutting edge approaches to non-rigid monocular
reconstruction from RGB video and \sfs, we propose a unified framework
for jointly reasoning about shape-from-shading and reconstructing
arbitrary deforming objects. Unlike existing methods, our general
approach is not object specific and targets non-Lambertian surfaces
such as skin while modeling both specularities and shading.  The
importance of a unified framework becomes even more apparent when handling video
shot outside the controlled lighting of a studio, where small
rotations of an object can induce significant changes of appearance
across most of the object, leading to a loss of tracking.  Further, we
empirically demonstrate that modeling the non-Lambertian properties of
surfaces such as skin, and capturing both specularities and shading is
vital for the joint integration of \sfs~with non-rigid reconstruction.

\begin{figure*}[t!]
	\centering
	\bgroup
	\setlength\tabcolsep{15pt}
\resizebox{\linewidth}{!}{
  \begin{tabular}{cccc}
    \emph{i)} Rendered Input&\emph{ii)} Yu~\etal&\emph{iii)} Ours (no spec)&\emph{iv)} Ours \\
    \includegraphics[width=0.2\columnwidth]{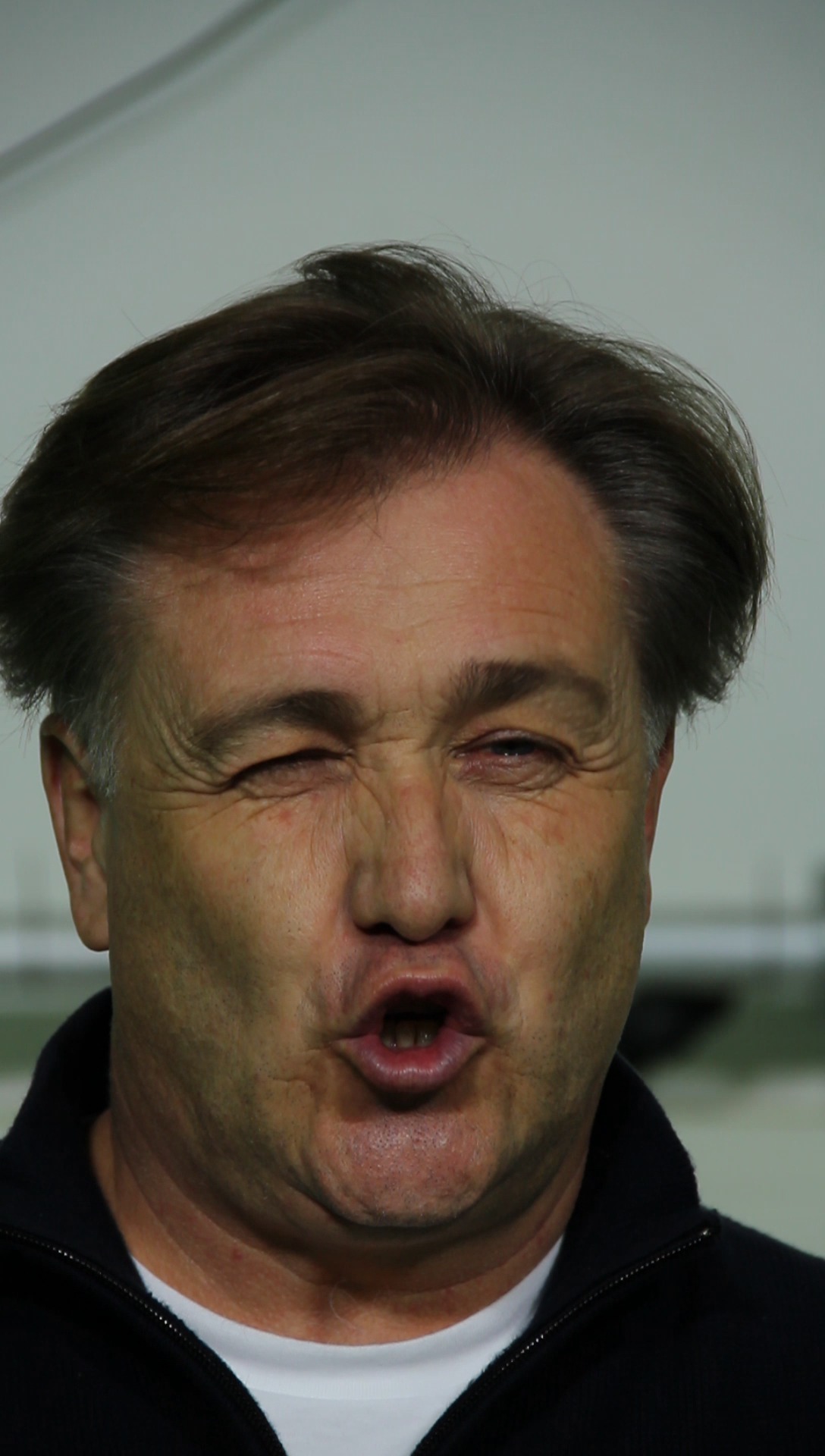}&
    \includegraphics[width=0.1525\columnwidth]{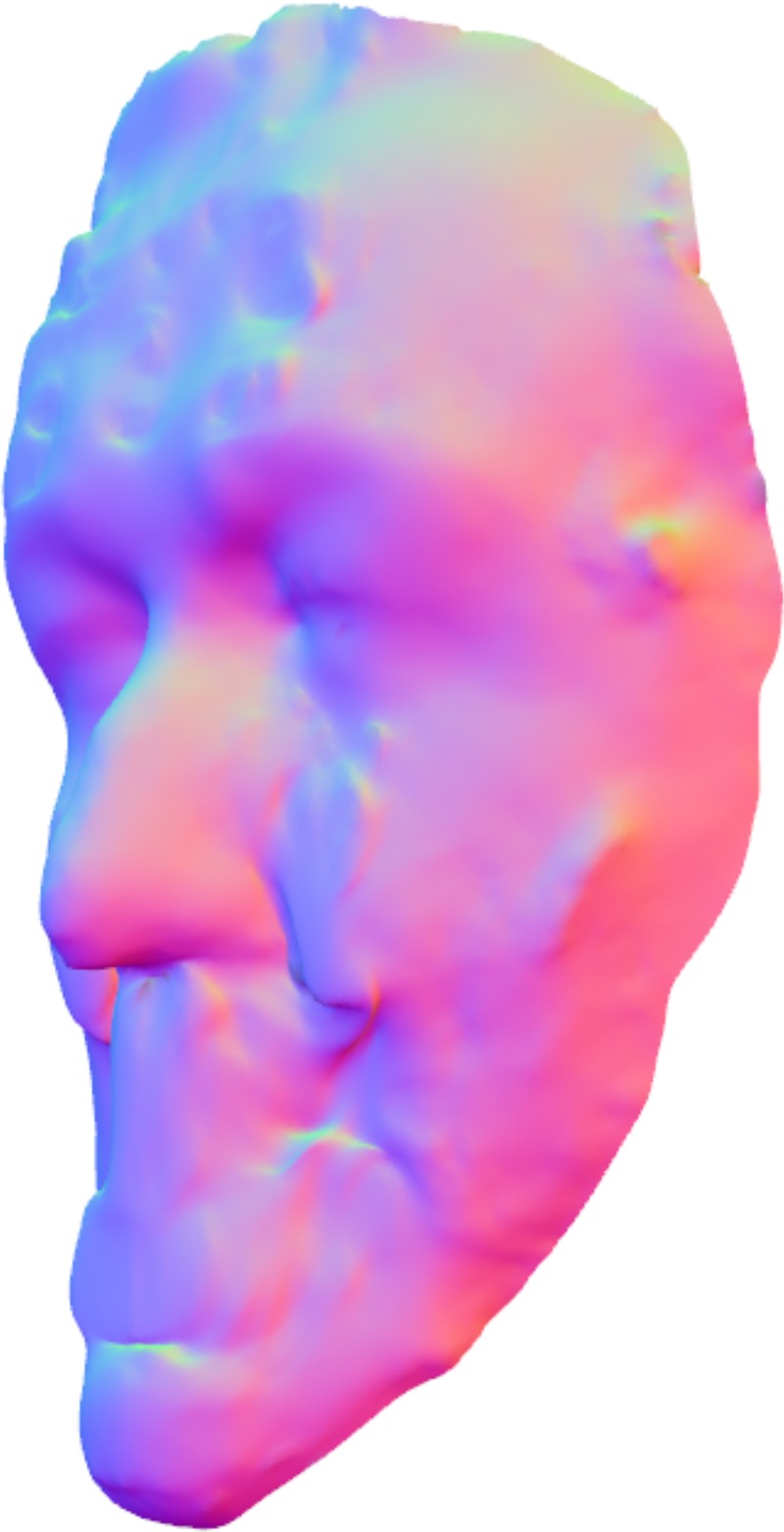}&
    \includegraphics[width=0.15\columnwidth]{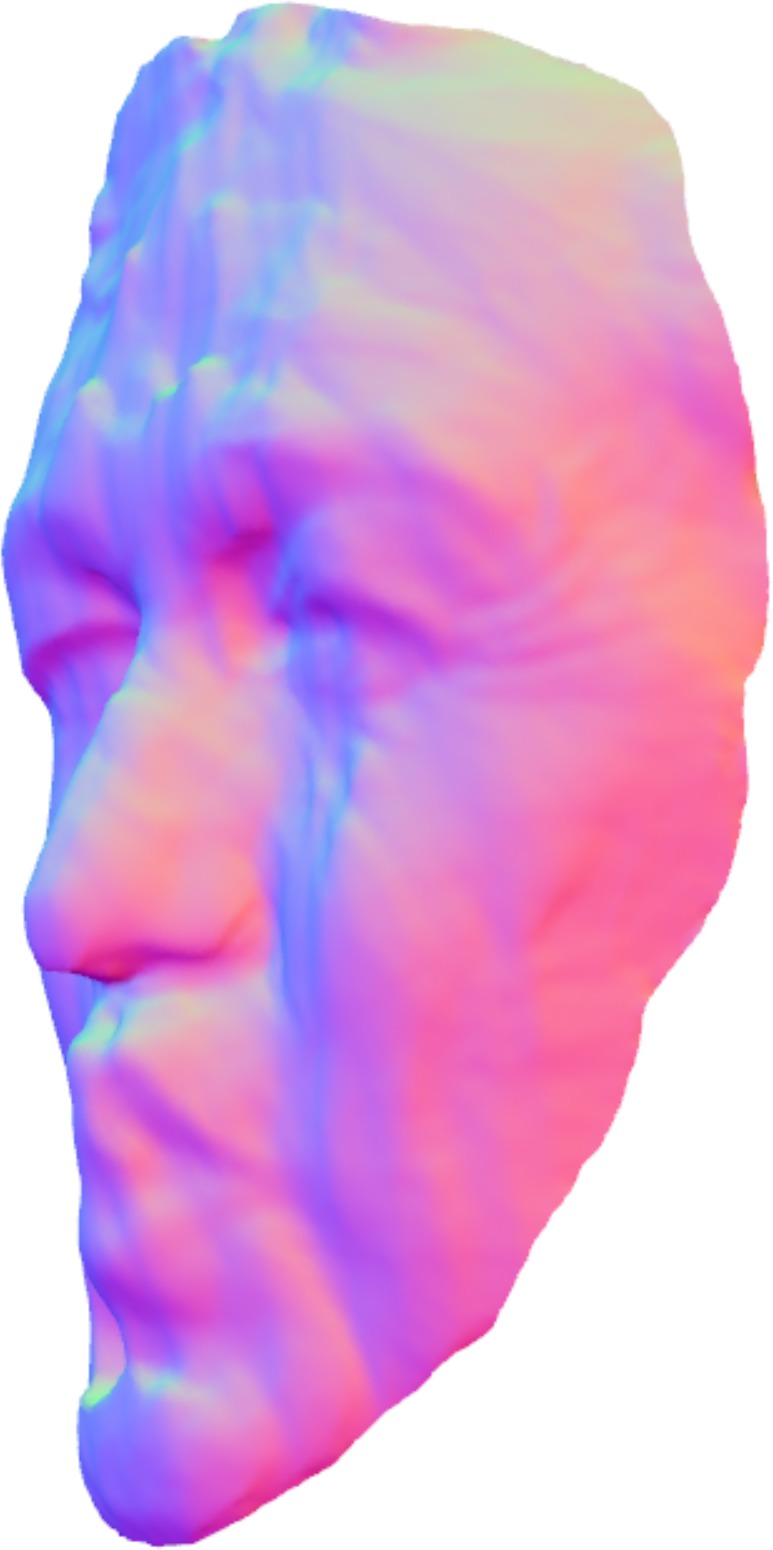}&
    \includegraphics[width=0.155\columnwidth]{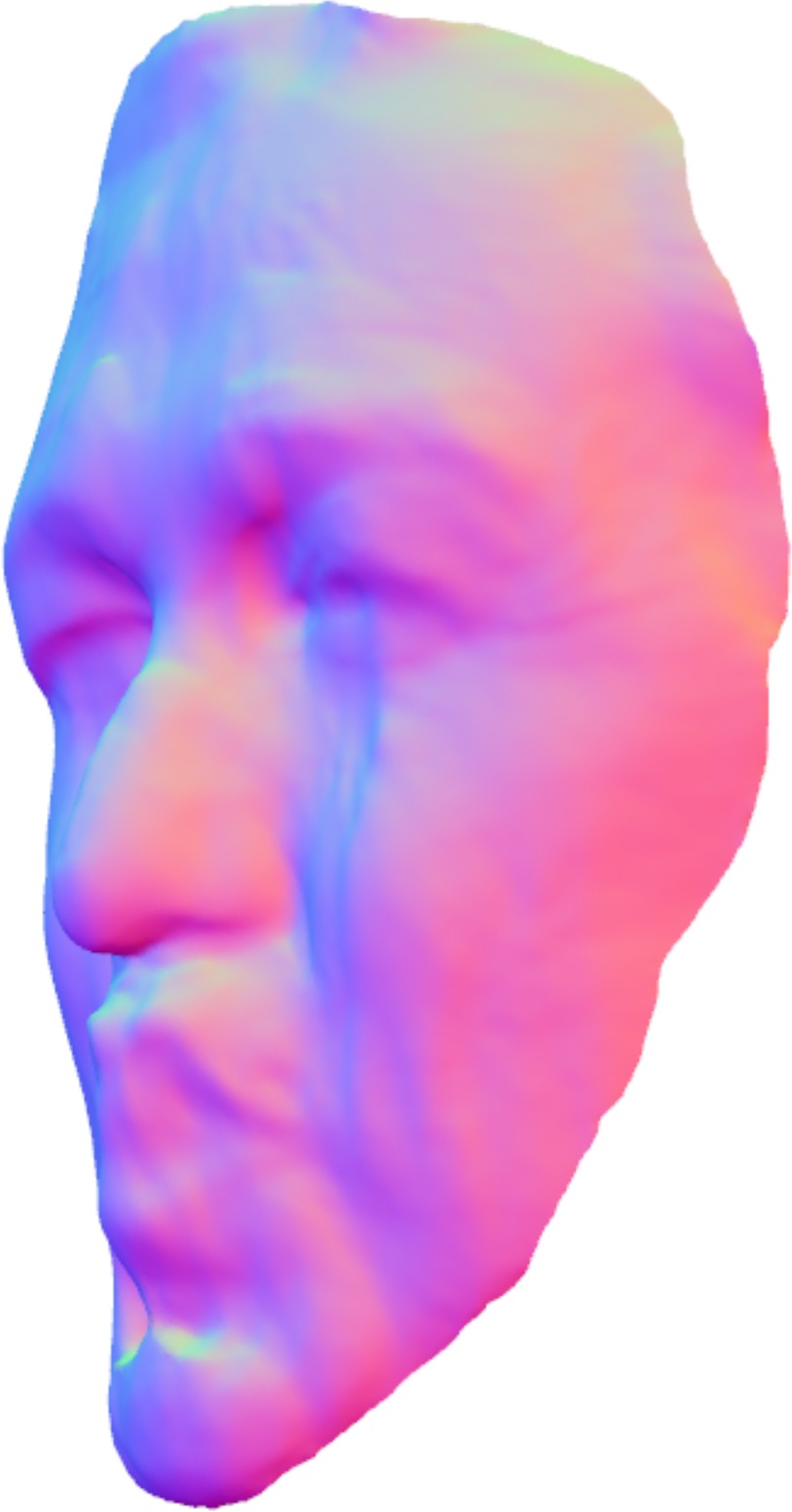}
  \end{tabular}}
  \egroup
  \caption{\label{fig:teaser}An illustration of our, and previous, approaches on
    a synthetic face sequence of a specular object under variable lighting. From
    Left to right: \emph{i)} Sample frame \emph{ii)} The direct reconstruction
    of ~\cite{Yu_2015_ICCV}, that does not consider shading artifacts.
    \emph{iii)} Our new approach that integrates \sfs~with non-rigid
    reconstruction but does not consider specularities. \emph{iv)} Our unified
    framework for \sfs, non-rigid structure from motion, and specularity
    modeling. This framework improves the accuracy of our approach over Yu
    \etal~by a factor of nearly 240\% reducing the RMS error from 9.28mm to
    3.84mm. Use of joint refinement rather than alternating refinement reduces
    the error further to 3.01mm and additional depth data almost halves the
    remaining error to 1.77mm. See discussion in section \ref{sec:results}, and
    table \ref{table:comparison_with_rui} for
    details. 
  }
\end{figure*}

One of the main challenges in non-rigid 3D reconstruction lies in
evaluating the quality of reconstructions. It is particularly
challenging to capture dense deforming objects of interest with
sufficiently high fidelity under real world lighting conditions. For
example, depth data from an infra-red structured light
source. e.g. the Microsoft Kinect or the purpose built depth camera of
\cite{kinect_nr:siggraph2014}, can not be captured under strong
natural light, while multi-camera visible-light techniques such as
\cite{Valgaerts_2012_SIGGRAPH} require relatively uniform lighting to
maintain tracking. To validate our approach we  compare both on real
world sequences captured using the stereo setup  of
\cite{Valgaerts_2012_SIGGRAPH}, and use this data to generate
realistic synthetic sequences containing severe shading artifacts that
could not be tracked by \cite{Valgaerts_2012_SIGGRAPH}. Our method
displays a strong qualitative and quantitative improvement over these
previous methods. See figure~\ref{fig:teaser} and
section~\ref{sec:results} for details. Our code and new ground truth datasets for evaluation have been made publicly available.
\section{Related Work}
\label{sec:related-work}
All previous attempts to unify shape-from-shading with non-rigid 3D
reconstruction from RGB video have been pipeline approaches
\citep{total_moving_face,Malti:etal:IPCAI:2012,Valgaerts_2012_SIGGRAPH,Garrido:etal:SiggraphAsia:2013}
which first coarsely reconstruct these deformable objects and then apply
shape-from-shading to refine the initial reconstruction. Examples of this
include, the seminal work Face Reconstruction in the Wild
\citep{kemelmacher2011face}, which first made use of automatic point
correspondences to compute warps and align images of a variety of celebrities,
before reconstructing faces using {\sfs} to build dense face models. This was
followed by \cite{total_moving_face}, which refined a coarser intensity based
model using {\sfs}.


\cite{varol2012monocular} fused shape-from-shading with non-rigid
reconstruction, but only performed shape-from-shading on untextured regions of
the objects, and non-rigid reconstruction on the textured areas, before fusing
these reconstructions as a post-processing step. Moreover, they required a known
light field and could not reconstruct high-frequency details such as facial
creases.

Several works have made use of \sfs~in refining depth maps, either captured
directly using a depth scanner \citep{Or-el_2015_CVPR,Or-el_2016_CVPR,yu2013shading} or captured using
a multi-camera setup \citep{Valgaerts_2012_SIGGRAPH}. Of the RGB-D
approaches \cite{Or-el_2016_CVPR} is the most related to ours, and computes both
\sfs~and specularities in order to enhance their depth
maps.
Previous works have also used \sfs~to improve tracking: In
the multi-camera work of \cite{Beeler_2012_ECCV}, they used a pipeline approach
to improve the tracking and refine the shape of an initial reconstruction by both
estimating and removing ambient occlusions.
While \cite{Xu_2005_ICCV}  defined linear
equations for modeling changes of illumination and position that occur when tracking a rigid object in video.

Our work builds on the recent template-based approach to monocular and direct  non-rigid 3D
reconstruction of \cite{Yu_2015_ICCV}. This work made no use
of \sfs, but generated vivid reconstructions of objects by deforming a
known template to match direct photometric cost. We extend this direct
formulation by augmenting the direct photometric cost with terms that
capture the change in appearance that goes with shape and shading,
leading to more lifelike and plausible reconstructions.

\begin{algorithm}[tb]
\caption{Joint non-rigid 3D reconstruction and
  shape-from-shading \label{algo:non_rigid_tracking and shading}}
\SetKwInOut{Input}{Input}\SetKwInOut{Output}{Output}
\SetKwInput{Initialization}{Initialization}
\SetKwFunction{Median}{Median}
\Input{3D Template mesh ${\bf\widehat{S}}$ + template albedo ${\widehat{\bm{\rho}}}_i$ (obtained using Algorithm~\ref{algo:template_acquisition})\\ Current video frame $\mathbf{I}^t$ \\ Solution to previous frame $\{\bS^{t-1}, \bR^{t-1}, \bt^{t-1}, \mathbf{l}^{t-1}, {\bm \beta}^{t-1}\}$}
\Output{Deformed shape $\bS^t$, rotation $\bR^t$, translation $\bt^t$,\\ spherical harmonic coefficients $\mathbf{l}^t$ and specularities ${\bm \beta}^t$ for current frame $t$}
\For{each new image frame $\mathbf{I}^t$}{
{\bf Initialize}  $\{\bS^t, \bR^t, \bt^t, \mathbf{l}^t, {\bm \beta}^t\} \leftarrow \{\bS^{t-1}, \bR^{t-1}, \bt^{t-1}, \mathbf{l}^{t-1}, {\bm \beta}^{t-1}\}$\\
Minimize \eqref{eq:energy_general_form} w.r.t. rigid alignment $\{\bR^t,\bt^t\}$ holding $\{\bS^{t},\mathbf{l}^{t}, {\bm \beta}^{t}\}$ constant\\
Minimize  \eqref{eq:energy_general_form} w.r.t. deformations and lighting  $\{\bS^{t}, \mathbf{l}^{t}\}$ holding $\{\bR^t, \bt^t, {\bm \beta}^{t}\}$ constant\\
Minimize  \eqref{eq:energy_general_form} w.r.t. specularities  ${\bm \beta}^{t}$ holding $\{\bS^{t}, \mathbf{l}^{t}, \bR^t,\bt^t\}$ constant\\
{\bf Joint refinement} Minimize  \eqref{eq:energy_general_form} w.r.t. all variables: $\bR^t,\bt^t$, $\bS^{t}$, $\mathbf{l}^{t}$ and  ${\bm \beta}^{t}$. \\
}
\end{algorithm}

\section{Problem Formulation}
\label{sec:problem-formulation}
Consider a single RGB perspective camera, of known internal
calibration, observing a non-rigid object. We propose a sequential,
frame-by-frame, approach to capture both the 3D geometry and the
reflectance properties of the non-rigid object. We parameterize the
object at time-step $t$ as a mesh $\bS^t$ with $N$ vertices
with associated 3D coordinates $\bS^t=\{\bs^t_i\}$, $i = 1..N$.

Our proposed approach is summarized in Algorithm~\ref{algo:non_rigid_tracking and shading}. The goal for
each incoming frame at time $t$ is to estimate the current 3D
coordinates of the vertices of the mesh, the light field --
parameterized in terms of spherical harmonics -- and specularities, as
well as the overall rigid rotation and translation $(\bR^t, \bt^t)$
that align the deformed shape and a reference 3D template. The only
inputs to our method are: the current image frame $\mathbf{I}^t$, the solution
to the previous frame, and a 3D template of the object (including its geometry $\bf{\widehat{S}}$ and albedo map ${\widehat{\bm{\rho}}}_i$)
acquired in a preliminary stage described in
Section~\ref{sec:template-capture} and
Algorithm~\ref{algo:template_acquisition}. Note that all variables related to the template are denoted with  $\boldsymbol{\hat{~}}$.

\begin{figure*}
\resizebox{\linewidth}{!}{
  \begin{tabular}{@{\hspace{2pt}}c@{\hspace{6pt}}cccccc}
    \rotatebox{90}{
      ~~~~~~~~~~~~~Input
    }&
    \includegraphics[width=0.13\columnwidth]{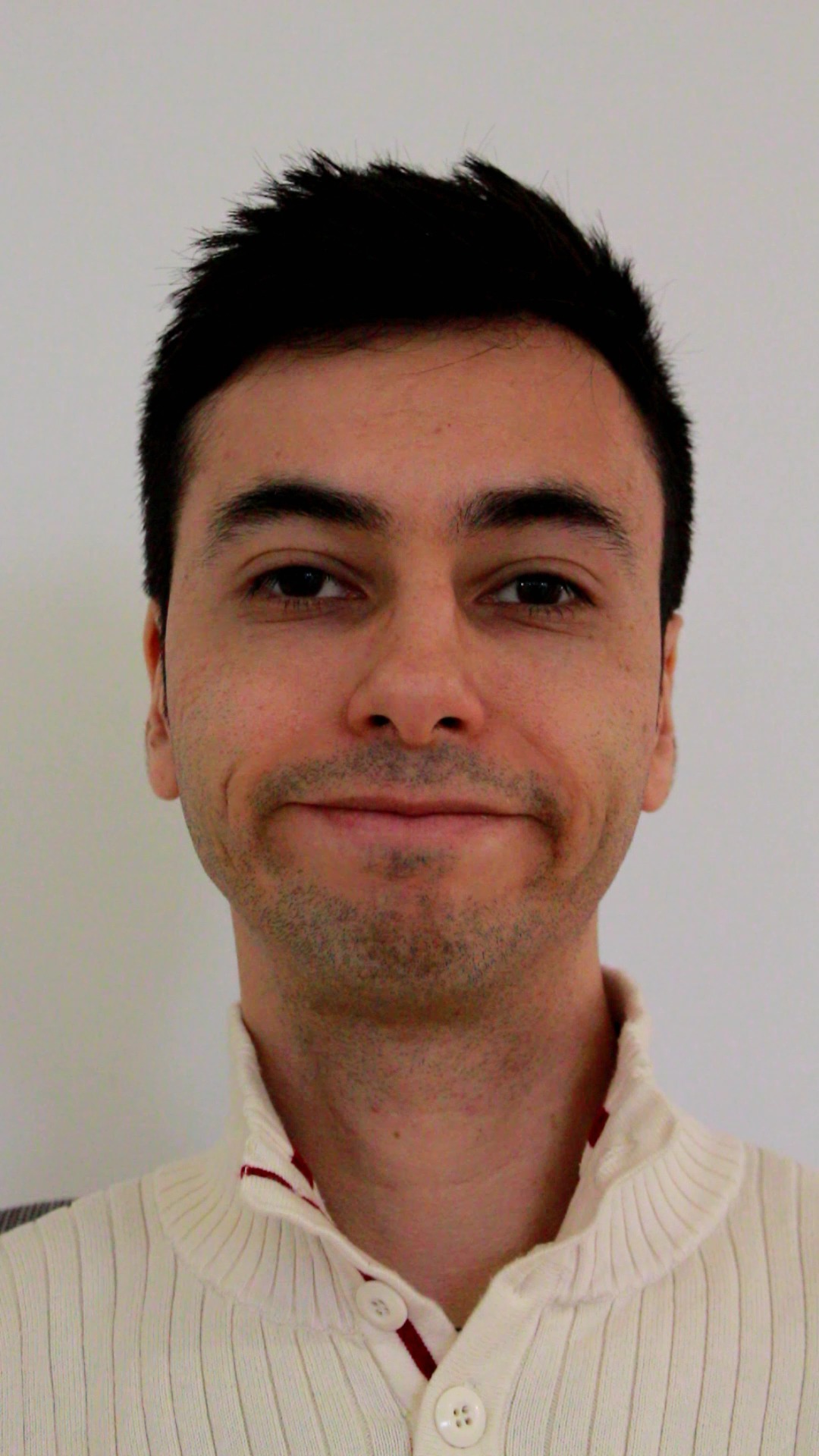}&
    \includegraphics[width=0.13\columnwidth]{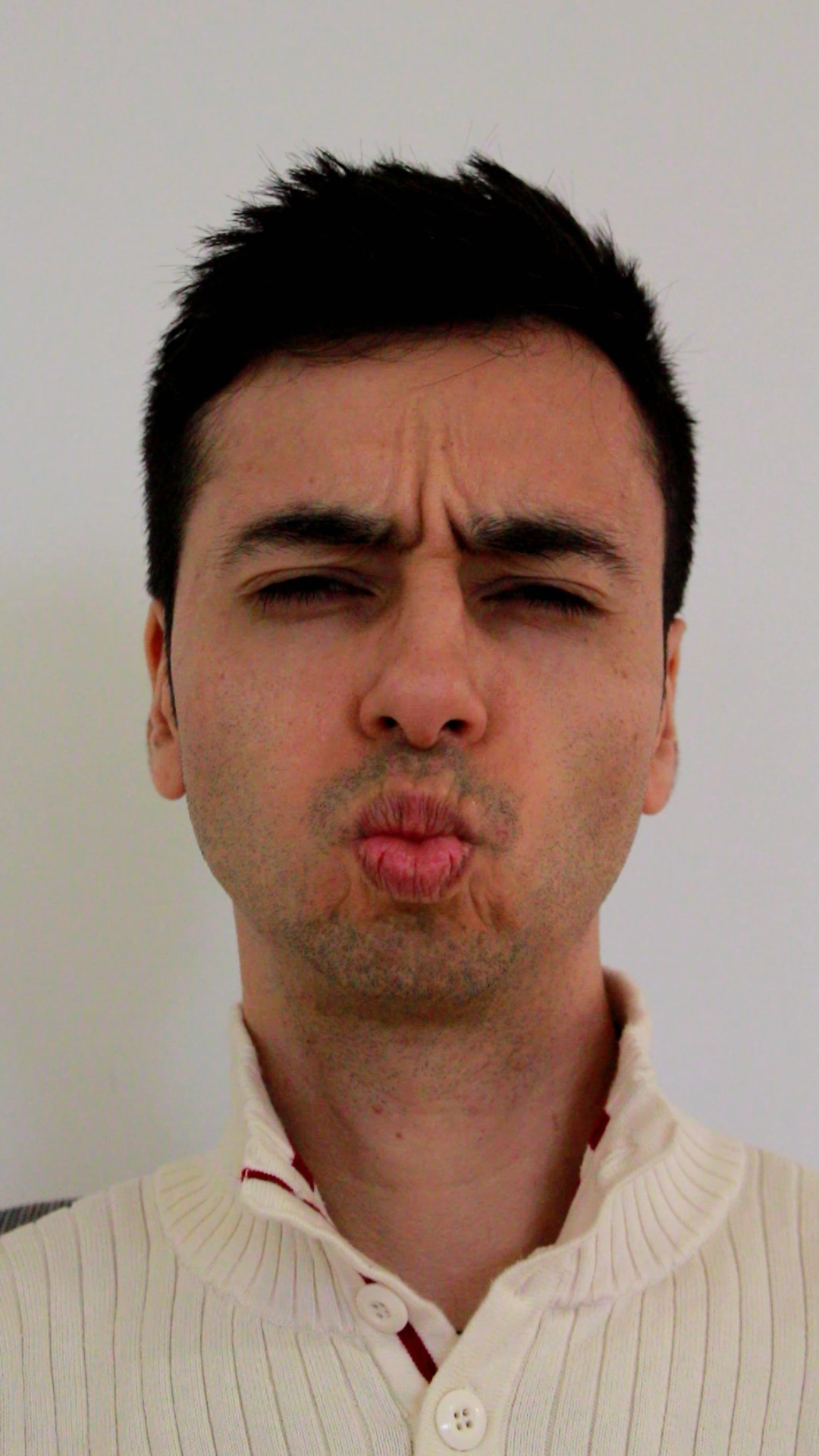}&
    \includegraphics[width=0.13\columnwidth]{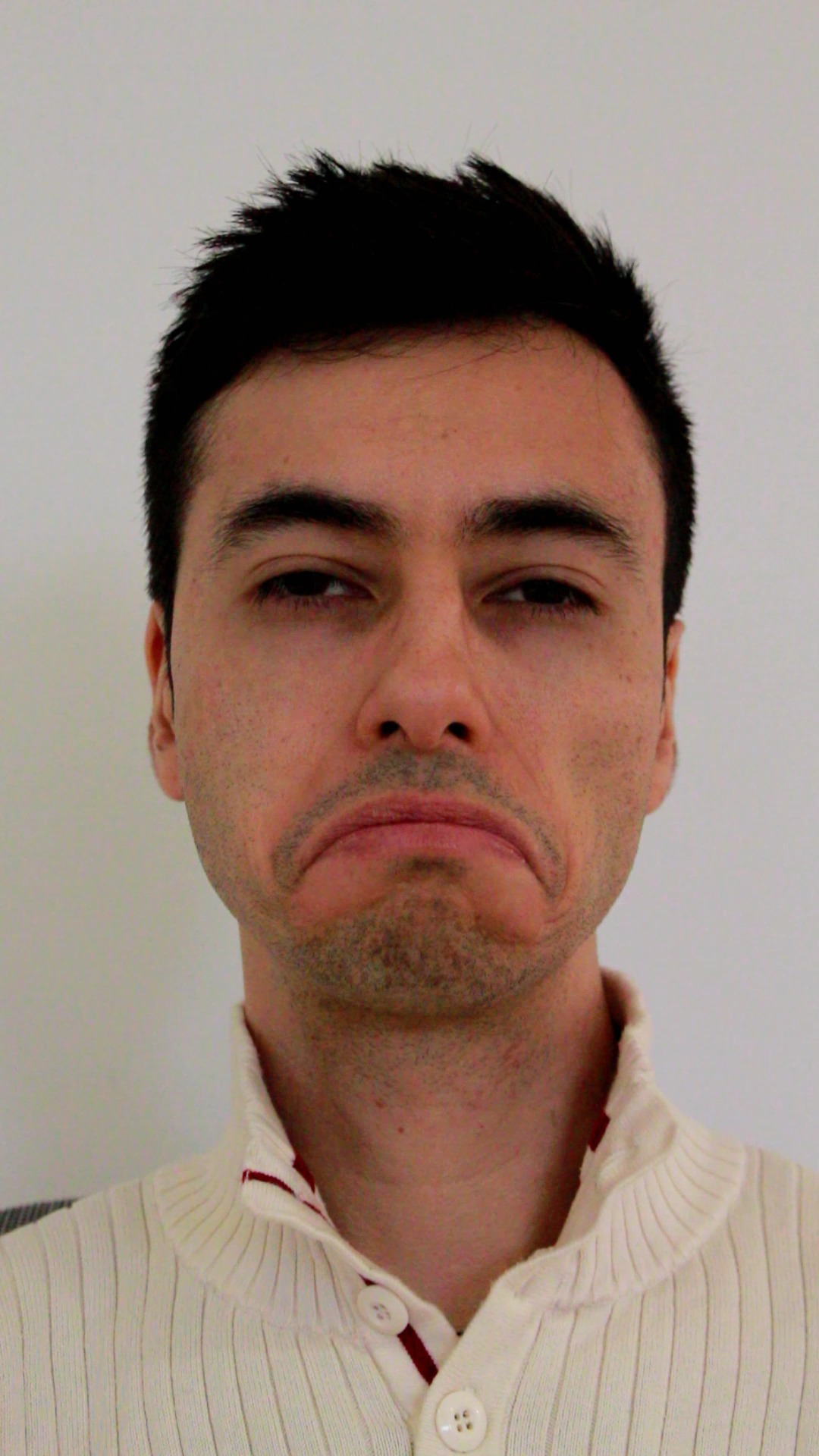}&
    \includegraphics[width=0.13\columnwidth]{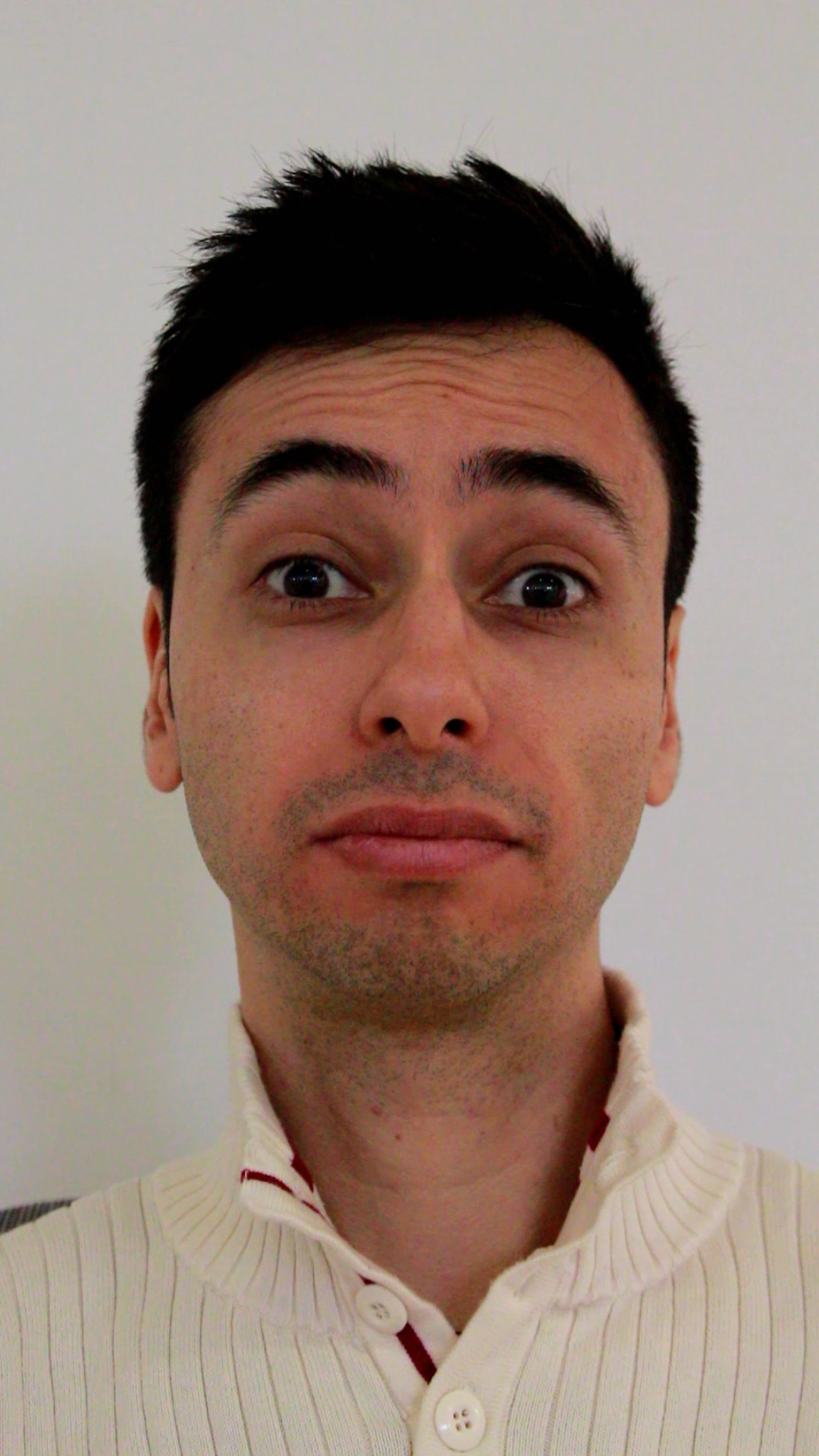}&
    \includegraphics[width=0.13\columnwidth]{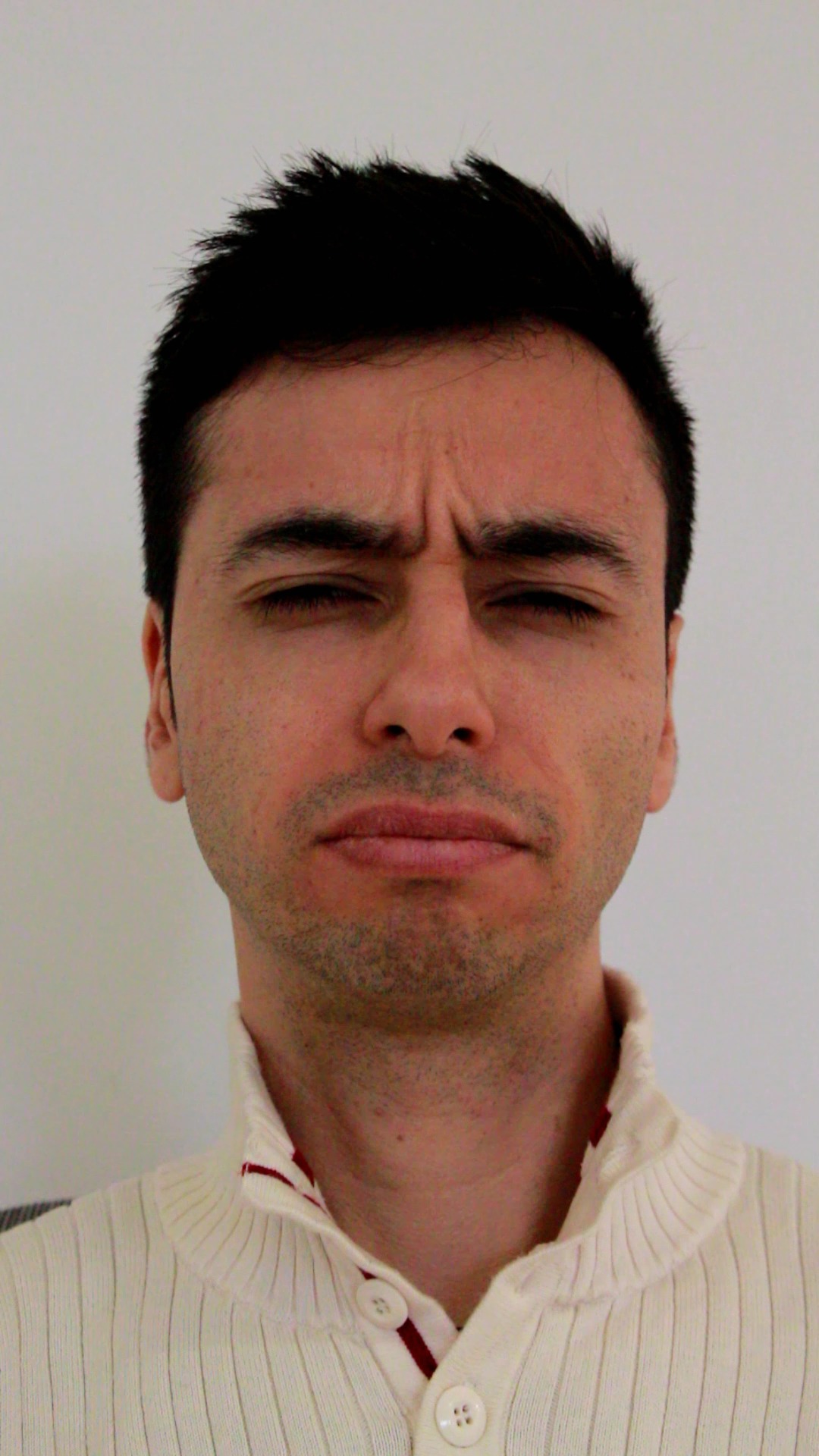}&
    \includegraphics[width=0.13\columnwidth]{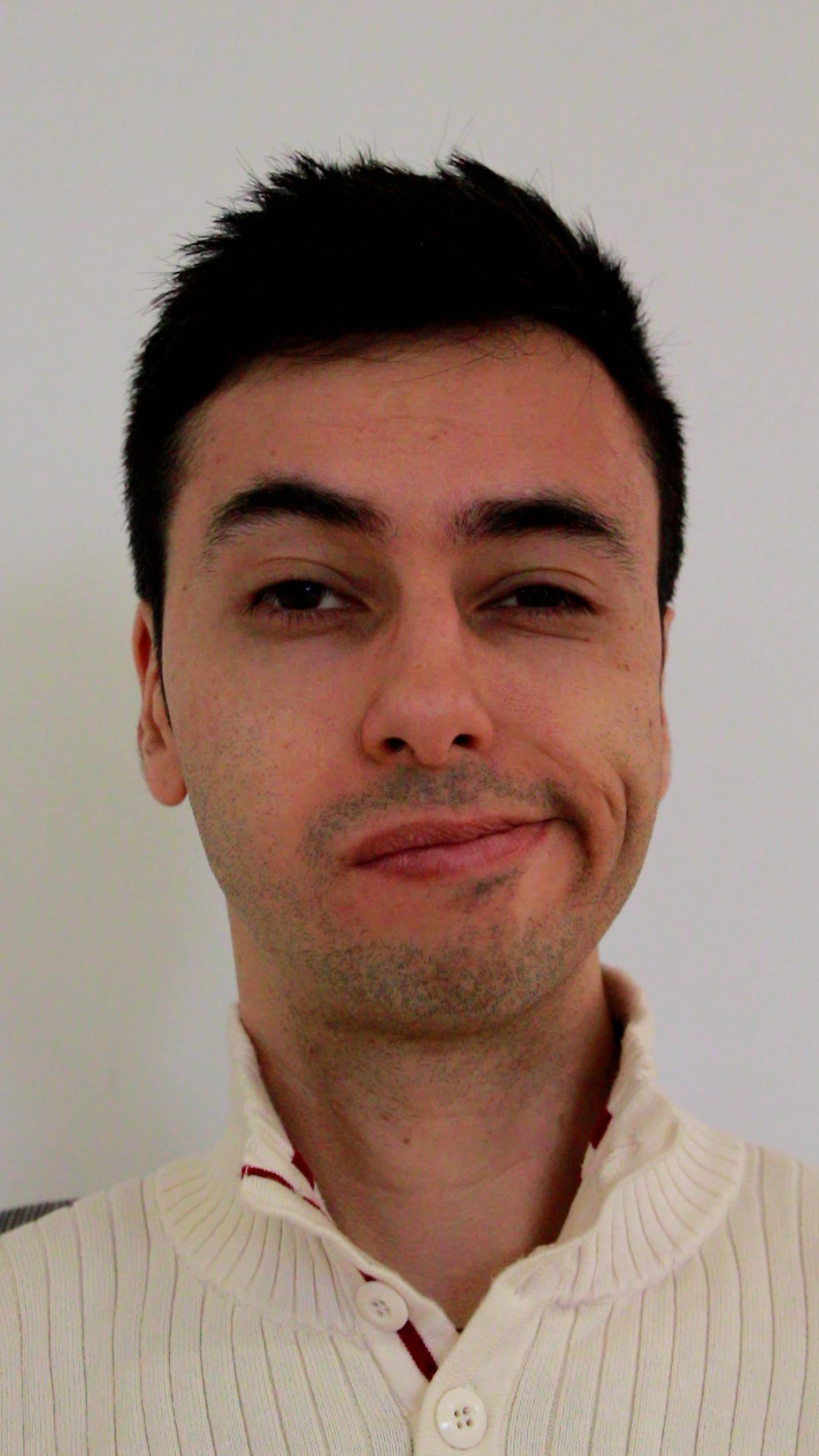}\\
    \rotatebox{90}{
      ~~~~Yu \etal
    }&
    \includegraphics[width=0.13\columnwidth]{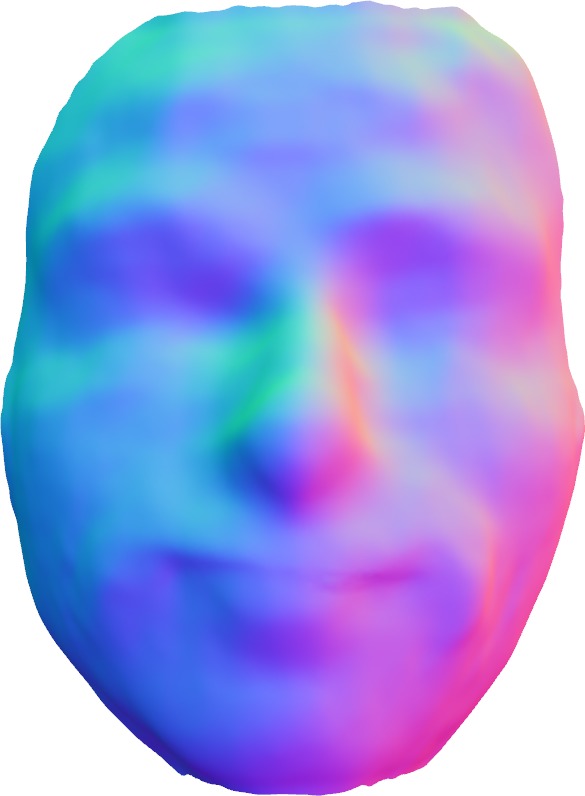}&
    \includegraphics[width=0.13\columnwidth]{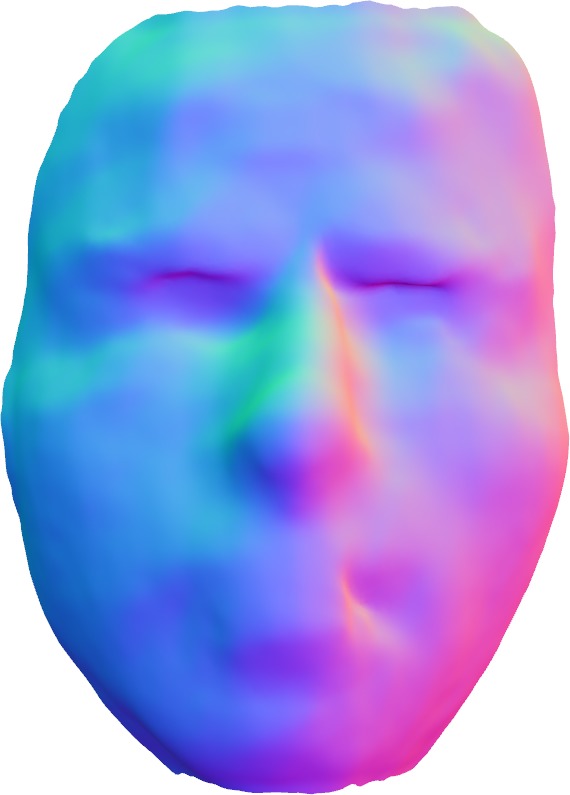}&
    \includegraphics[width=0.13\columnwidth]{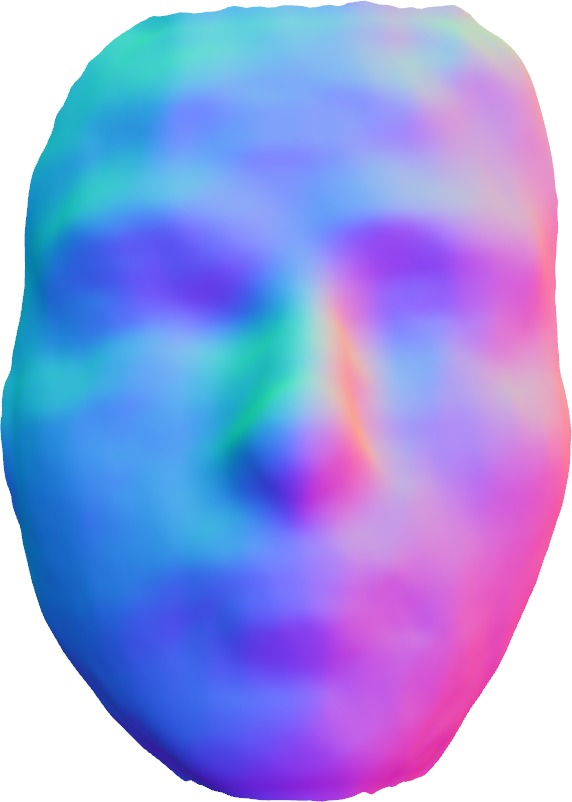}&
    \includegraphics[width=0.13\columnwidth]{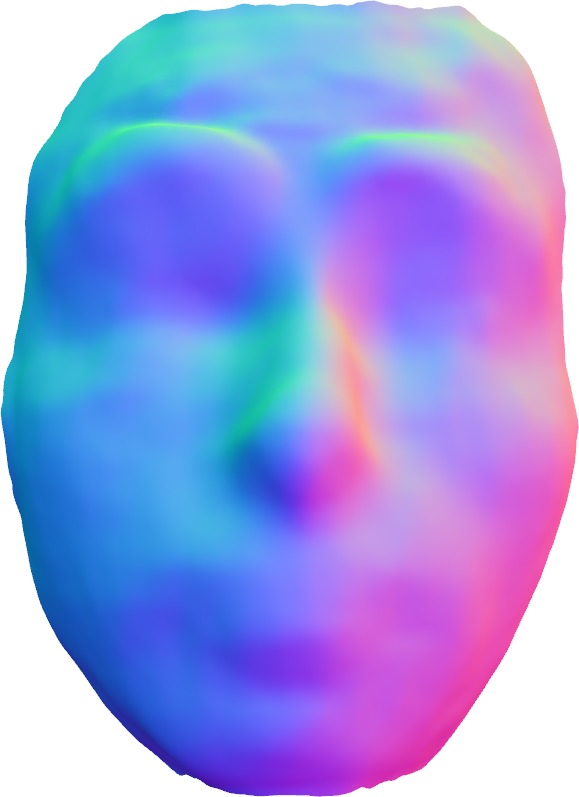}&
    \includegraphics[width=0.13\columnwidth]{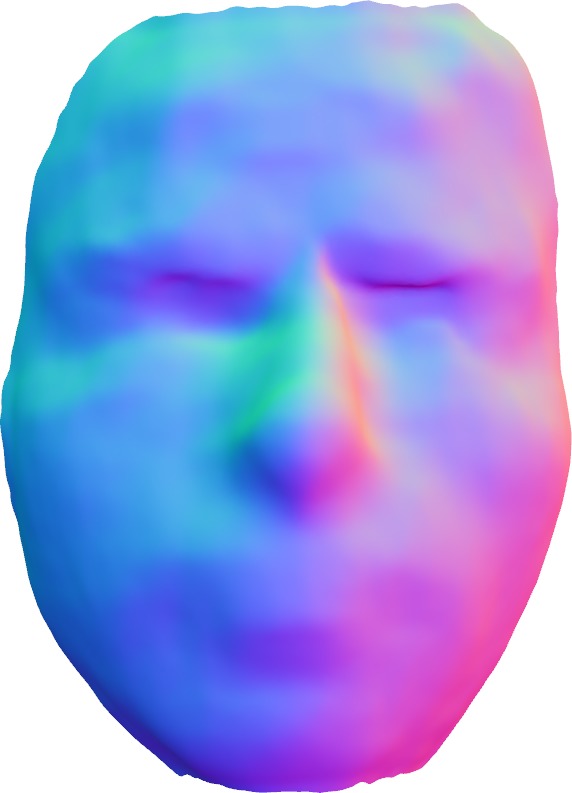}&
    \includegraphics[width=0.13\columnwidth]{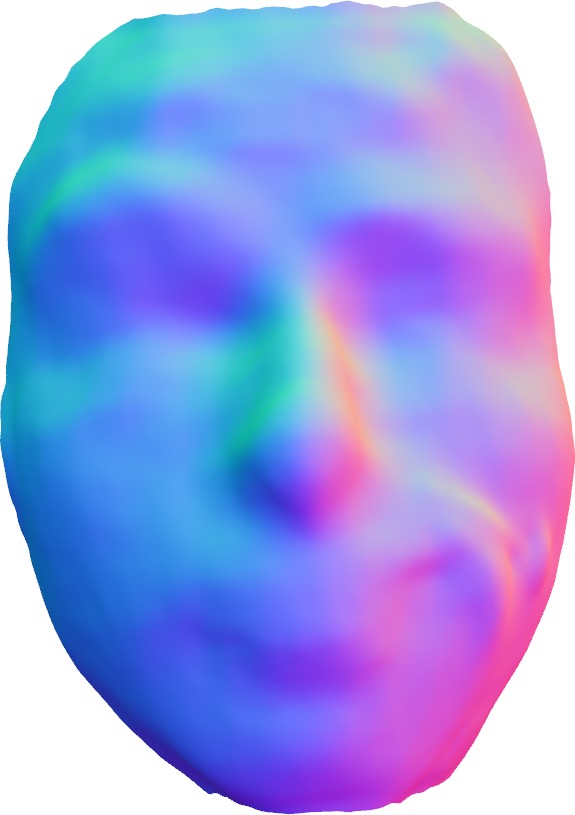}\\
    \rotatebox{90}{
      ~~~~~~Ours
    }&
    \includegraphics[width=0.13\columnwidth]{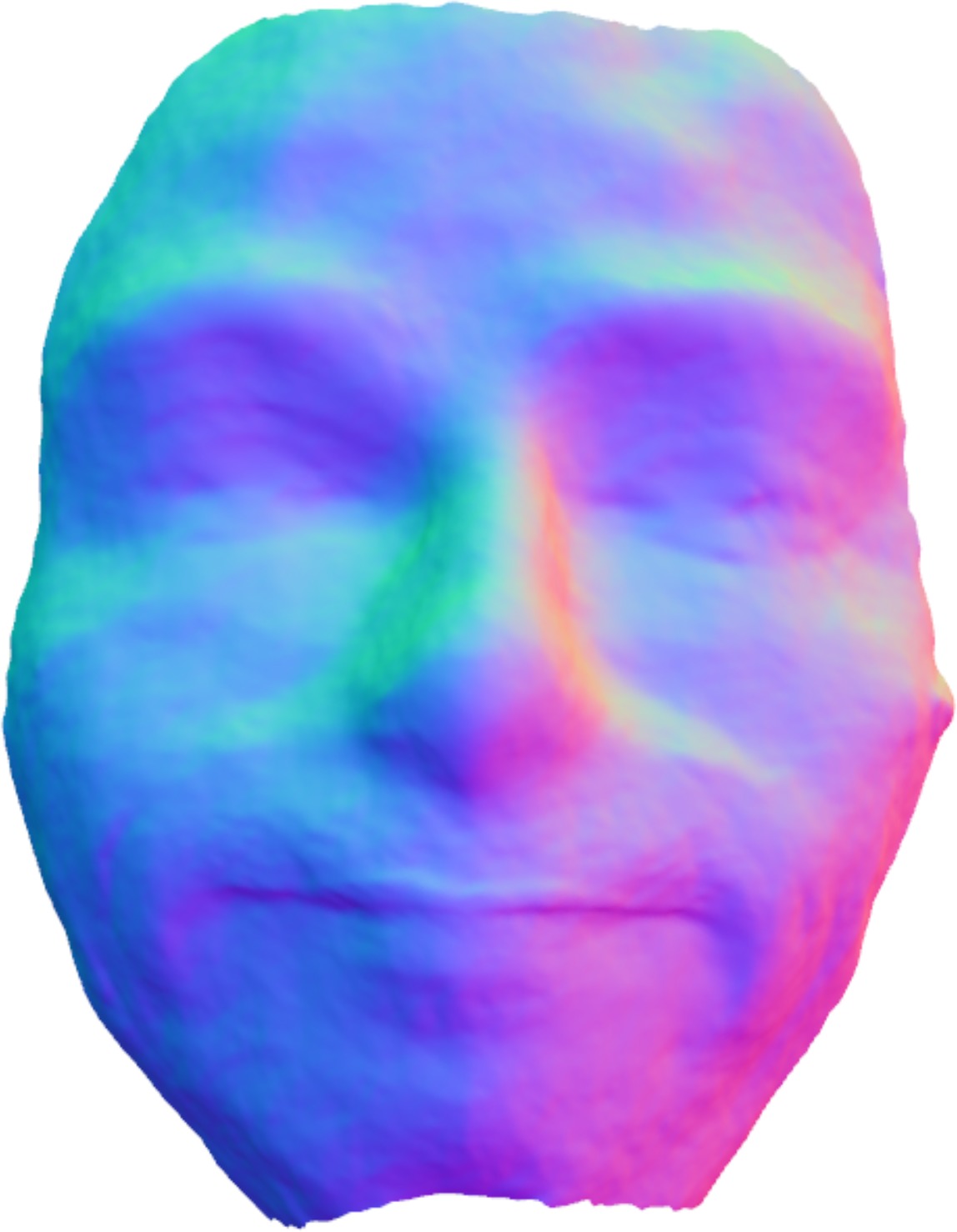}&
    \includegraphics[width=0.13\columnwidth]{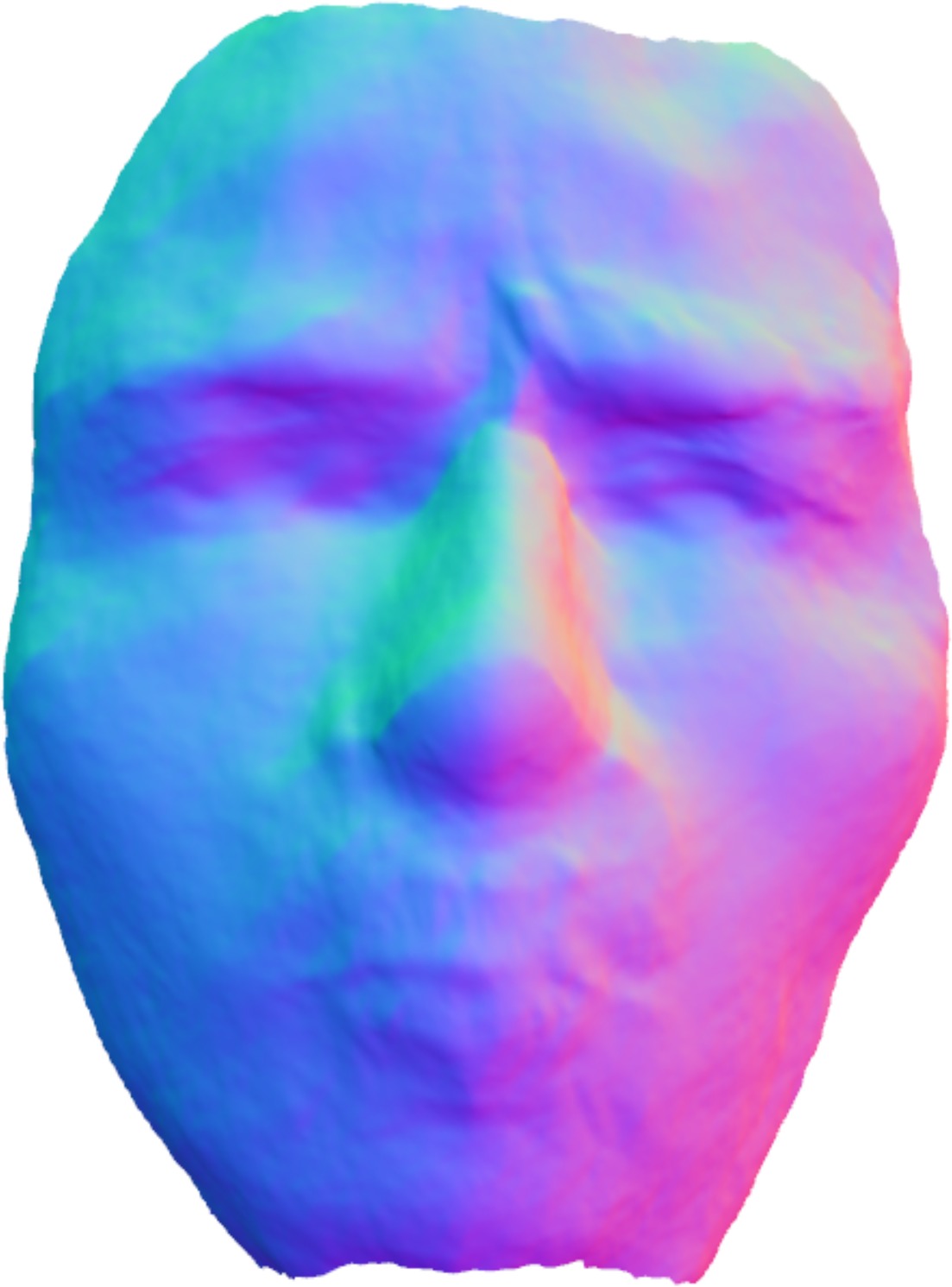}&
    \includegraphics[width=0.13\columnwidth]{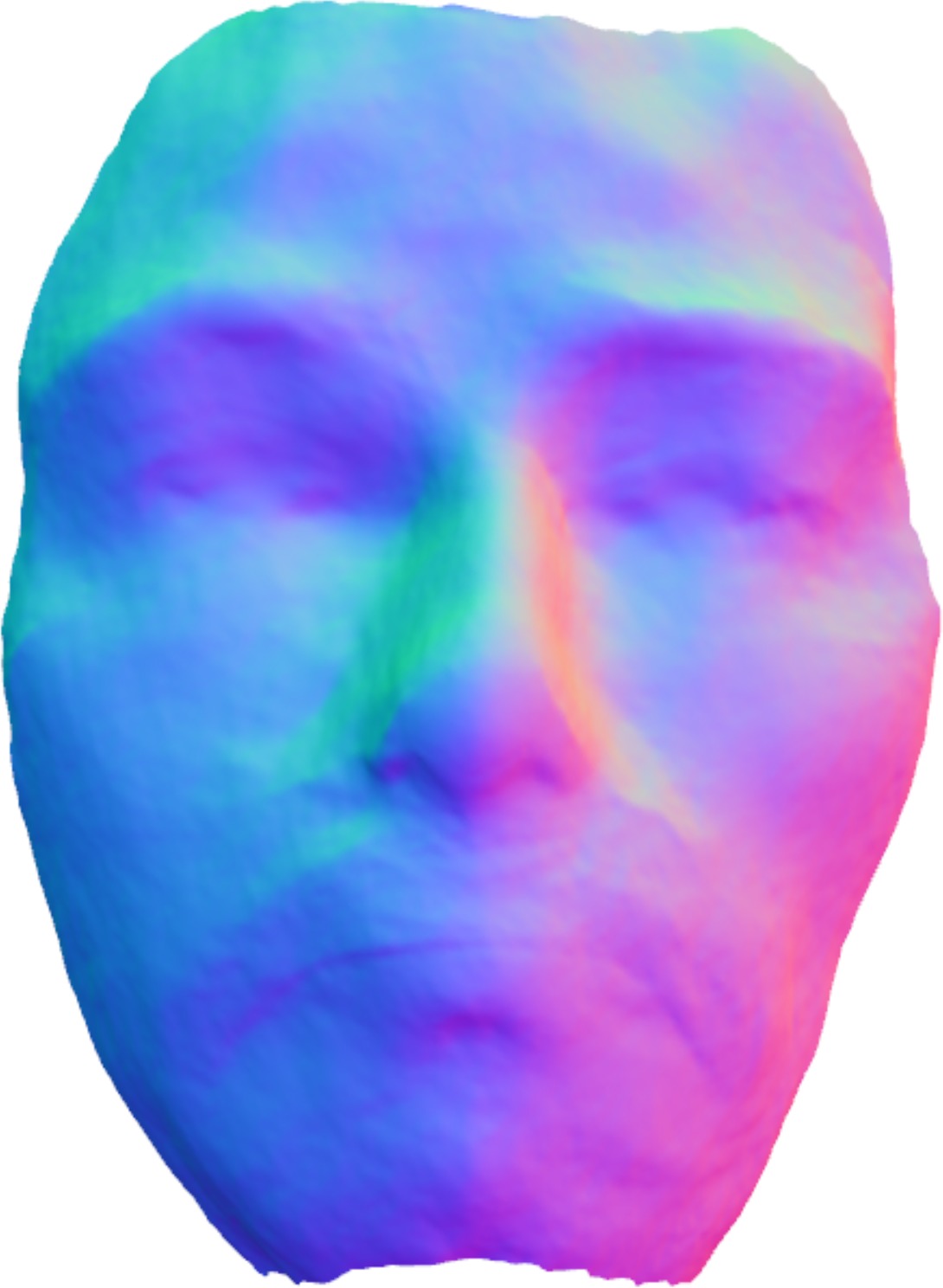}&
    \includegraphics[width=0.13\columnwidth]{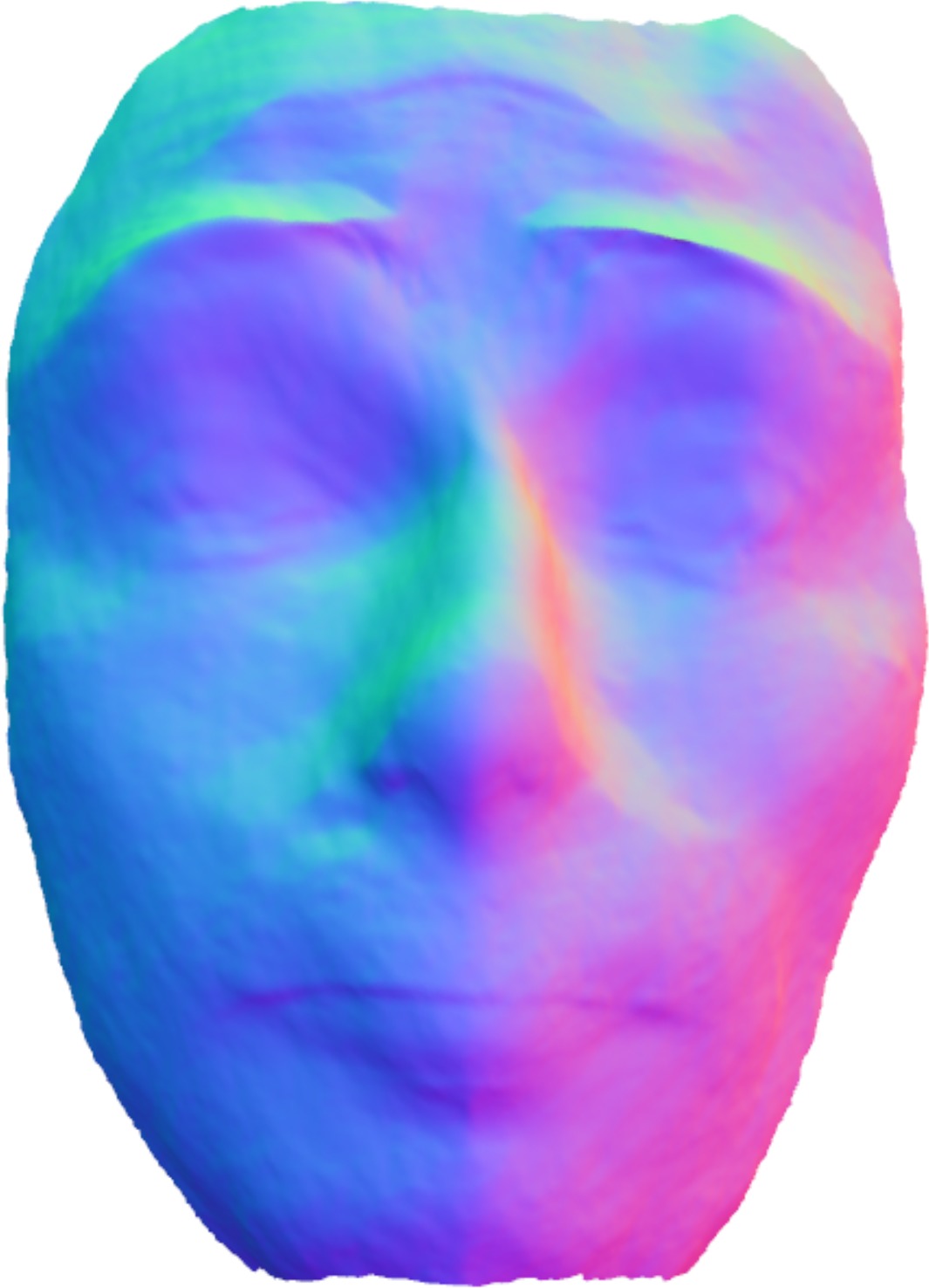}&
    \includegraphics[width=0.13\columnwidth]{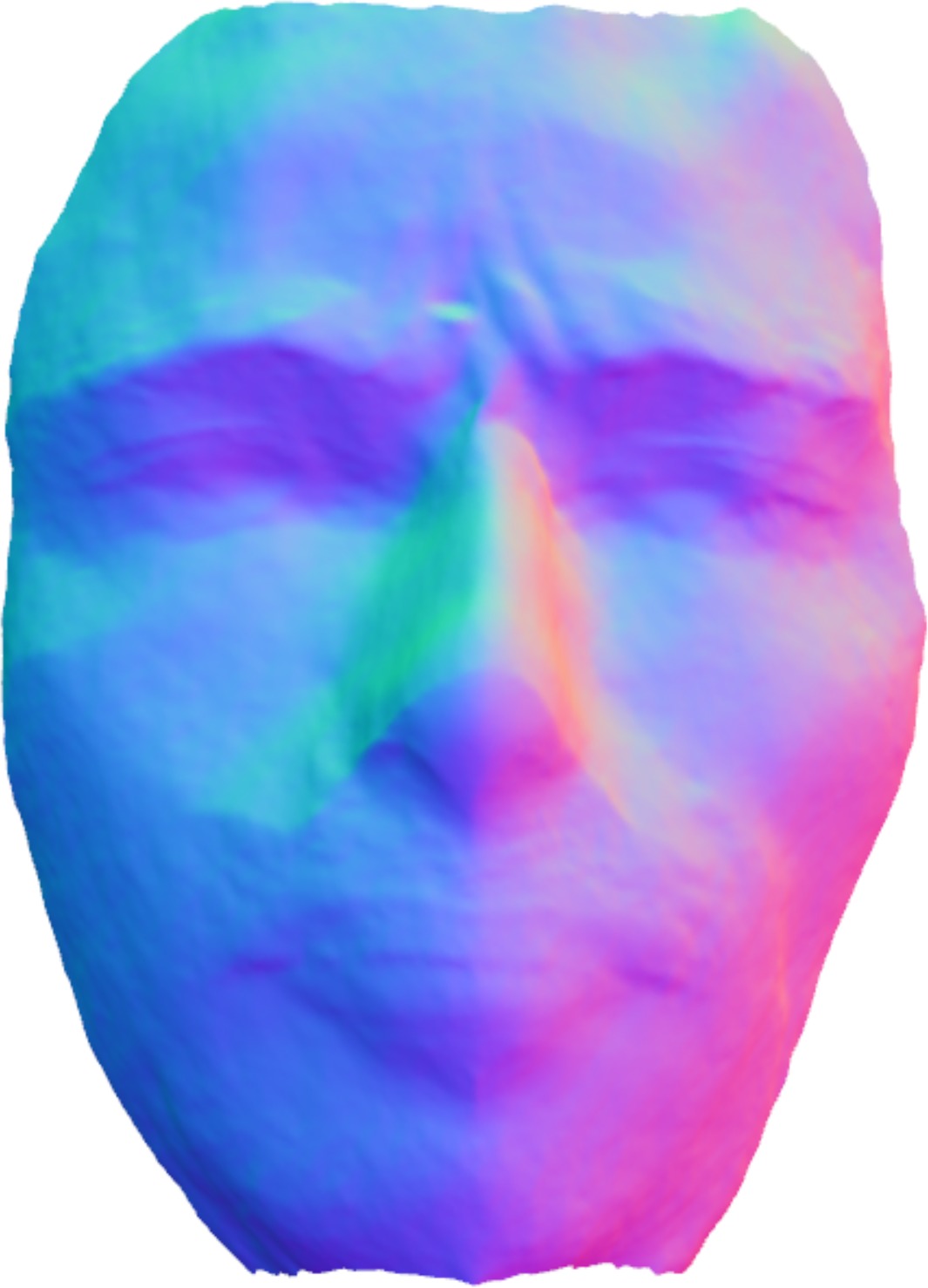}&
    \includegraphics[width=0.13\columnwidth]{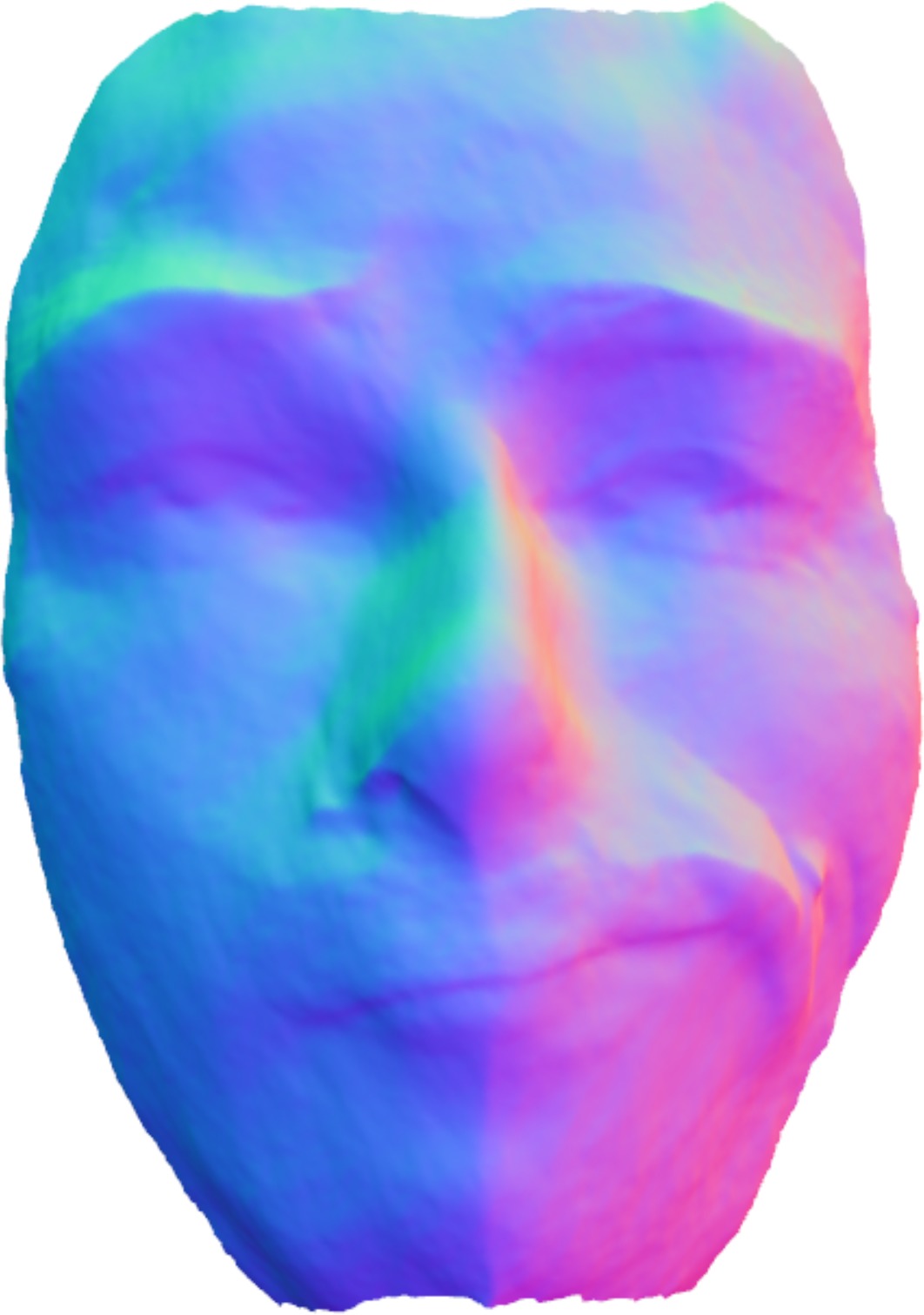}\\
  \end{tabular}}\textbf{}
  \caption{\label{fig:ivan}Best viewed in color: A qualitative comparison of our method against that of
    \cite{Yu_2015_ICCV}, on their face dataset.}
\end{figure*}

\section{Reflectance Model}
\label{sec:reflectance-model}
Modern
solutions, such as \cite{Yu_2015_ICCV,total_moving_face,Valgaerts_2012_SIGGRAPH},
to \emph{direct} 3D reconstruction of non-rigid objects from RGB video
adopt an energy optimization approach that minimizes a robust
photometric cost based upon \emph{brightness constancy}. In other
words, they jointly estimate dense correspondences alongside non-rigid
deformations, by penalizing differences in intensity between
images and the new deformed shape, which is assumed to be
the same color and brightness of a reference template. As the points
on the object change color in response to differences in illumination
or in shading caused by strong deformations these methods need to use
 robust costs to cope with deviations from the model.

In contrast, our method explicitly models the reflectance properties of
non-Lambertian objects and can handle  materials which exhibit a
mixture of specular and diffuse reflection properties.
In practice we adopt an approximation of the Phong
reflection model  
which models light leaving the
non-rigid object at point $i$ as the sum of two additive terms: a
viewpoint-independent diffuse term and a view-dependent specular term:
\begin{equation} \label{eq:full_reflection_model}
\bI_i =  \bI_i^\text{diff}+ \bm{\beta}_i.
\end{equation}
In other words, the intensity at point $i$  can be
explained as the sum of the specularity-free diffuse component
$\bI_i^\text{diff}$ and the specular component $\bm{\beta}_i$.

To increase our robustness to changes in lighting and shading and to recover
high frequency details of the object geometry, we decouple the diffuse
component into the product of object albedo and the object irradiance or
shading (see figure \ref{fig:decomposition}). While the albedo is independent of the surface orientation, the
shading is a function of the surface normal at each vertex $i$
\begin{equation} \label{eq:lambertian_reflectance}
\bI_i^\text{diff} = \bm{\rho}_i r(\mathbf{n}_i(\bS)).
\end{equation}
Here
$\bm{\rho}_i$ is the RGB reflectance or albedo, $\mathbf{n}_i(\cdot)$ is
a function that returns the direction of the surface normal at vertex
$i$, and $r(\cdot)$ is an irradiance map function that returns the
shading value given the surface orientation vector. We assume white
illumination so $r(\cdot)$ returns a single scalar value greater than 0.

Following Basri and Jacobs \cite{Basri_2003_PAMI} we model the
irradiance map using a spherical harmonic basis
\begin{equation}\label{eq:irradiance_map}
r(\mathbf{n}_i(\bS))= \sum_{n=0}^{N} \sum_{m=-n}^{n} l_{nm} Y_{nm}(\mathbf{n}_i(\bS)) = \mathbf{l} \cdot Y(\mathbf{n}_i(\bS))
\end{equation}
where $l_{nm}$ is the coefficient associated with the spherical
harmonic function $Y_{nm}$. We limit our approximation to second order
spherical harmonics, i.e. $N=2$ giving $\mathbf{l}$  nine
coefficients.

If we consider a video of a non-rigid object evolving over time $\bS^t$,
our reflection model allows us to write the predicted image intensity of point
$i$ observed at time $t$ as
\begin{equation} \label{eq:full_reflection_model_final}
\bI_i^t =   \bm{\rho}_i \mathbf{l}^t \cdot Y(\mathbf{n}_i({\bS}^t))+ \bm{\beta}_i^t
\end{equation}

It is clear that our reflection model allows us to cope not only with varying
geometry ($\bS^t$) but also varying illumination coefficients ($\mathbf{l}^t$)
and specularities ($\bm{\beta}_i^t$). Notably, while the image brightness of
vertex $i$ might vary over time $t$ due to possible changes in illumination and
object surface normals $\mathbf{n}_i({\bS}^t)$ caused by the deformations, its
albedo $\bm{\rho}_i$ is constant over the entire sequence.

While previous approaches to non-rigid object reconstruction from RGB
video might include a post-processing step to add high frequency
details to the tracked objects by performing a similar
shape-from-shading decomposition \cite{total_moving_face} our work
incorporates this step directly into the 3D tracking. In essence, our
approach effectively takes advantage of changes in shading caused by
variations in the illumination and in the surface normals due to
deformations to improve 3D tracking while recovering high frequency
surface details.

Previous approaches based on \emph{brightness
  constancy} needed to adopt robust penalty terms to be able to
discard specularities as outliers or to be resilient to changes to
illumination. We will show that the benefits of modeling these effects
directly in our reflectance model are twofold: (\emph{i}) substantial
improvements in 3D non-rigid tracking and (\emph{ii}) direct recovery
of high frequency geometry details such as folds and wrinkles.

Our insight and the main contribution of this work is to track the
non-rigid deformations of the object based on \emph{albedo constancy}
instead of the more classical \emph{image brightness constancy}
constraint which does not hold for non-rigid objects or when the
illumination varies over time. In this way, we can take advantage of
the changes in illumination and shading to recover high frequency
details in non-rigid objects and by increasing 3D tracking accuracy.




\section{A Sequential Approach to Joint Non-Rigid 3D Reconstruction and Reflectance Estimation}
\label{sec:proposed-method}

\begin{figure*}[t]
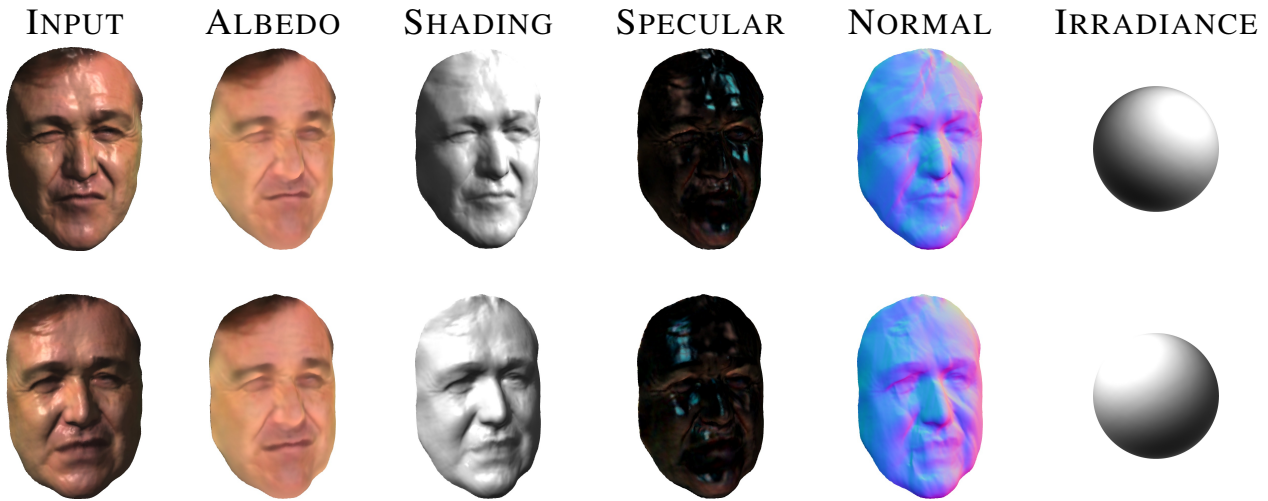

	\bgroup
	\setlength\tabcolsep{7pt}
	\begin{center}
\resizebox{\linewidth}{!}{
		\begin{tabular}{cccccc}
			\sc{Input} & \sc{Albedo} & \sc{Shading} & \sc{Specular} & \sc{Normal} & \sc{Irradiance} \\
			\adjustimage{width=0.13\columnwidth, valign=m}{rgb0073-trim}&
			\adjustimage{width=0.12\columnwidth, valign=m}{albedo73-trim}&
			\adjustimage{width=0.12\columnwidth, valign=m}{shading73-trim}&
			\adjustimage{width=0.12\columnwidth, valign=m}{specular73-trim}&
			\adjustimage{width=0.12\columnwidth, valign=m}{normal73-trim}&
			\adjustimage{width=0.12\columnwidth, valign=m}{sh_73-trim}\\ \\
			\adjustimage{width=0.13\columnwidth, valign=m}{rgb0120-trim}&
			\adjustimage{width=0.12\columnwidth, valign=m}{albedo120-trim}&
			\adjustimage{width=0.12\columnwidth, valign=m}{shading120-trim}&
			\adjustimage{width=0.12\columnwidth, valign=m}{specular120-trim}&
			\adjustimage{width=0.12\columnwidth, valign=m}{normal120-trim}&
			\adjustimage{width=0.12\columnwidth, valign=m}{sh_120-trim}
		\end{tabular}}
	\end{center}
	\egroup
	\caption{\label{fig:decompositionSC}Results of two different frames from the
    synthetic SC sequence (see section \ref{sec:results}) and the corresponding
    intrinsic decomposition (section \ref{sec:reflectance-model}). }
\end{figure*}
Much like \cite{Yu_2015_ICCV}, we use a template-based
approach to track and reconstruct non-rigid objects. However,
while \cite{Yu_2015_ICCV} assigned a fixed intensity to each vertex on
the template mesh, we decompose the intensity of each
vertex (see equation~\eqref{eq:full_reflection_model_final}) allowing us to take advantage of the
reflectance properties of the object to improve the resulting
reconstructions.

\subsection{Our Energy}
\label{sec:our-energy}

Our per-frame objective takes the form
%
\begin{align}\label{eq:energy_general_form}
  {E}(\bS,\bR,\bt,\mathbf{l},{\bm \beta})=&{E}_\text{data}(\bS,\bR,\bt,\mathbf{l},{\bm \beta}) + {E}_\text{smooth}(\bS,\bm \beta)\\
                                          &+E_\text{arap} (\bS)\ + E_\text{temp}(\bS,\bt,\mathbf{l},{\bm \beta}\nonumber)\\
                                          &+E_\text{sparse}({\bm \beta})\nonumber
\end{align}
\subsubsection{ Data Term $E_\text{data}$:} Our data term is a direct
photometric cost. Rather than minimizing the more commonly used
\emph{brightness constancy constraint} we use the more complex
reflectance model described in~\eqref{eq:full_reflection_model_final}
which decomposes the intensity of vertex $i$ into the product of a
constant albedo and a time-varying shading term (where variations can
be due to changes in illumination or to strong deformations) and
explicitly models specularities.
\begin{align}\label{eq:tracking_data_term}
E_\textnormal{data}\left(\mathbf{R},\mathbf{t},\mathbf{S},\mathbf{l},\bm{\beta}\right) = \sum_{i\in\mathcal{V}} \big\|
&\mathbf{I}\left( \pi\left(\mathbf{R}(\mathbf{s}_i) + \mathbf{t}\right) \right)\\
&\, -{\widehat{\bm{\rho}}}_i \mathbf{l} \cdot Y(\mathbf{R} (\mathbf{n}_i({\bS})))
-\bm{\beta}_i \big\|_\epsilon \nonumber
\end{align}
where
$\mathbf{I}$ is the current image frame,
$\cal V$ is the set of estimated visible vertices, \footnote{This is
  generated by realigning the deformed mesh of the previous frame to
  minimize photometric error, 
  and z-buffering.}  $\mathbf{R}$ and $\mathbf{t}$ are the rigid
rotation and translation of the object, $\{\mathbf{s}_i\}_1^N$ are the
3D vertices of the shape in the current frame, $\pi(\cdot)$ is the
projection from 3D points to image coordinates, known from camera
calibration, and ${\widehat{\bm{\rho}}}_i$ is the albedo of vertex $i$
on the template mesh.  and $\|\cdot\|_\epsilon$ is the Huber loss.
\subsubsection{ Spatial Smoothness Term $E_\text{smooth}$:}
\label{sec:spat-smoothn-term}
This regularization term encourages spatially smooth deformations of
the shape and specularities. In practice the spatial smoothness on the
shape is decoupled into two terms: a total variation term that
encourages smooth deformations of the shape $\bS$ with respect to the
template ${\bf\widehat{S}}$ and a \emph{Laplacian} smoothness term
\begin{align}\label{eq:ereg}
  E_\text{smooth}(\bS,\bm{\beta})
  &=E_\text{smooth}(\bS) + E_\text{smooth}(\bm{\beta}) \\
  &= E_\text{TV}(\bS)+ E_\text{Laplacian}(\bS) + E_\text{spec}(\bm{\beta})  \nonumber\\
  &=
    \sum_{i\in \mathcal V} \sum_{j \in \mathcal{N}_i}  \|(\bs_i -
    \bs_j) - ({\bf\widehat{s}}_{i} - {\bf\widehat{s}}_{j})\|_{\epsilon}\nonumber\\
  &+ \sum_{i\in \mathcal V} {\frac{1}{\left| \mathcal{N}_i\right|} \left\| \sum_{j\in\mathcal{N}_i} \left(\mathbf{s}_i - \mathbf{s}_j\right) \right\|_2^2} \nonumber\\
  &+\sum_{i\in \mathcal V}\sum_{j\in\mathcal{N}_i}{\left\| \bm{\beta}_i - \bm{\beta}_j\right\|_\epsilon}\nonumber
\end{align}
where $\mathcal{N}_i$ is the neighborhood of $i$.

\subsubsection{ ARAP Term $E_\text{arap}$:}
\label{sec:arap-term-e_text}
This \emph{as rigid as possible} cost \citep{Sorkine:ARAP} encourages
non-rigid objects to preserve locally rigidity while deforming. It
allows for local rotations to occur while preserving the relative
locations between neighboring points.
\begin{equation}\label{eq:earap}
E_\text{arap}(\bS,\{\bA_i\}) =  \sum_{i=1}^N \sum_{j \in \mathcal{N}_i} \| (\bs_i -
\bs_j) - \bA_i ({\bf\widehat{s}}_{i} - {\bf\widehat{s}}_{j})\|_2  ^2
\end{equation}
where the variables $A_i$ describe  per-point local rotations.

\subsubsection{ Temporal Smoothness $E_\text{temp}$:}
This set of temporal regularizers prevents flickering throughout the sequence
\begin{align}\label{eq:etemp}
  E_\text{temp}(\bS,\bt,\mathbf{l},\bm{\beta}_i) = &\left\|  \bS-\bS^{t-1}  \right\|^2_{\mathcal F} + \left\|  \bt-\bt^{t-1}  \right\|_2^2  \\
                                                   &+{\left\| \mathbf{l} - \mathbf{l}^{t-1} \right\|_2^2}+{ \left\| \boldsymbol{\beta} - \boldsymbol{\beta}^{t-1} \right\|_2^2}\nonumber
\end{align}
where $\bS^{t-1}$, $\bt^{t-1}$, $\mathbf{l}^{t-1}$, $\bm{\beta}^{t-1}$
are the shape, translation, spherical harmonic coefficients and
specularities in the previous frame and $\|\cdot\|_{\mathcal F}$
denotes the Frobenius norm of a matrix.

\subsubsection{ Sparsity Term $E_\text{sparse}$:} This prevents the entire
image being ``explained away'' as a specularity, by penalizing the use
of specularities.
\begin{equation}
E_\text{sparse}(\bm{\beta})=\sum_{i\in\mathcal{V}}{\left\| \bm{\beta}_i
  \right\|_\epsilon}
\end{equation}
\subsubsection{\bf Energy Optimization:}
For reasons of efficiency, we adopt a multi-stage optimization similar
to the approach taken by real-time SLAM approaches such as
\cite{Klein:Murray:ISMAR2007}. Starting from the solution given by the
previous frame, we hold all other coefficients fixed and optimize
first over rotations and translations, followed by jointly optimizing
shape and spherical harmonic coefficients, and finally re-estimating
the specularities.

The first two of these optimizations are performed coarse-to-fine over
a 3-layer spatial pyramid, providing robustness against sudden
movements and deformations of the object while for efficiency reasons,
specularities are only estimated at the finest level, and propagated
to the coarser levels of the pyramid, ready for the next
iteration. Algorithm~\ref{algo:non_rigid_tracking and shading}
summarizes our optimization strategy. We use the Levenberg-Marquardt
implementation from the Ceres solver \citep{ceres-solver} for all
continuous optimization, applying preconditioned conjugate gradient
for the linear solver. When handling color images, the pyramid is
formed by down-sampling using Gaussian filters, while for depth images
a median filter is used to avoid artifacts along the boundary of the object.

In the experimental section, particularly
table~\ref{table:additional_comparison}, we also compare against
performing joint optimization over all variables to convergence, after
initializing each frame using the procedure described above. This
leads to a noticeable improvement in performance.

\section{Reconstructing with RGB-D data}
We also consider sequences where depth cues have been provided by an
RGB-D camera such as the Microsoft
Kinect \citep{Newcombe:etal:CVPR2015}, or an Infra-red stereo setup
\citep{kinect_nr:siggraph2014}. Even with powerful depth cues, visual
tracking provides important information that allows the correct
tracking of smooth and deforming surfaces such as the cheeks and brow
of a person. As previously discussed these color cues can be polluted
by shading effects, making tracking based purely on color consistency
and depth challenging. Moreover, these shading cues provide important
information about the high-frequency changes in depth such as creases
about the brow and eyes that are too fine to be captured directly by
depth sensors.

As such, it is vital to make use of the entire pipeline we have
proposed as well as depth cues for vivid high-quality reconstructions
that are robust to the effects of lighting. In these cases, the energy
can be augmented by an additional term that encourages consistency
between the depth map and the generated reconstruction.  In this case,
we introduce an additional cost based on the distance between the
reconstruction and the depth map
\begin{equation}
  \label{eq:depth}
  E_\text{depth} ({\bf S},{\bf D})
  =\sum_{i=1}^N \argmin_{{\bf d} \in \bf D} \left\| {\bf s}_i -{\bf d} \right\|_\epsilon
\end{equation}

\begin{figure*}[t!]
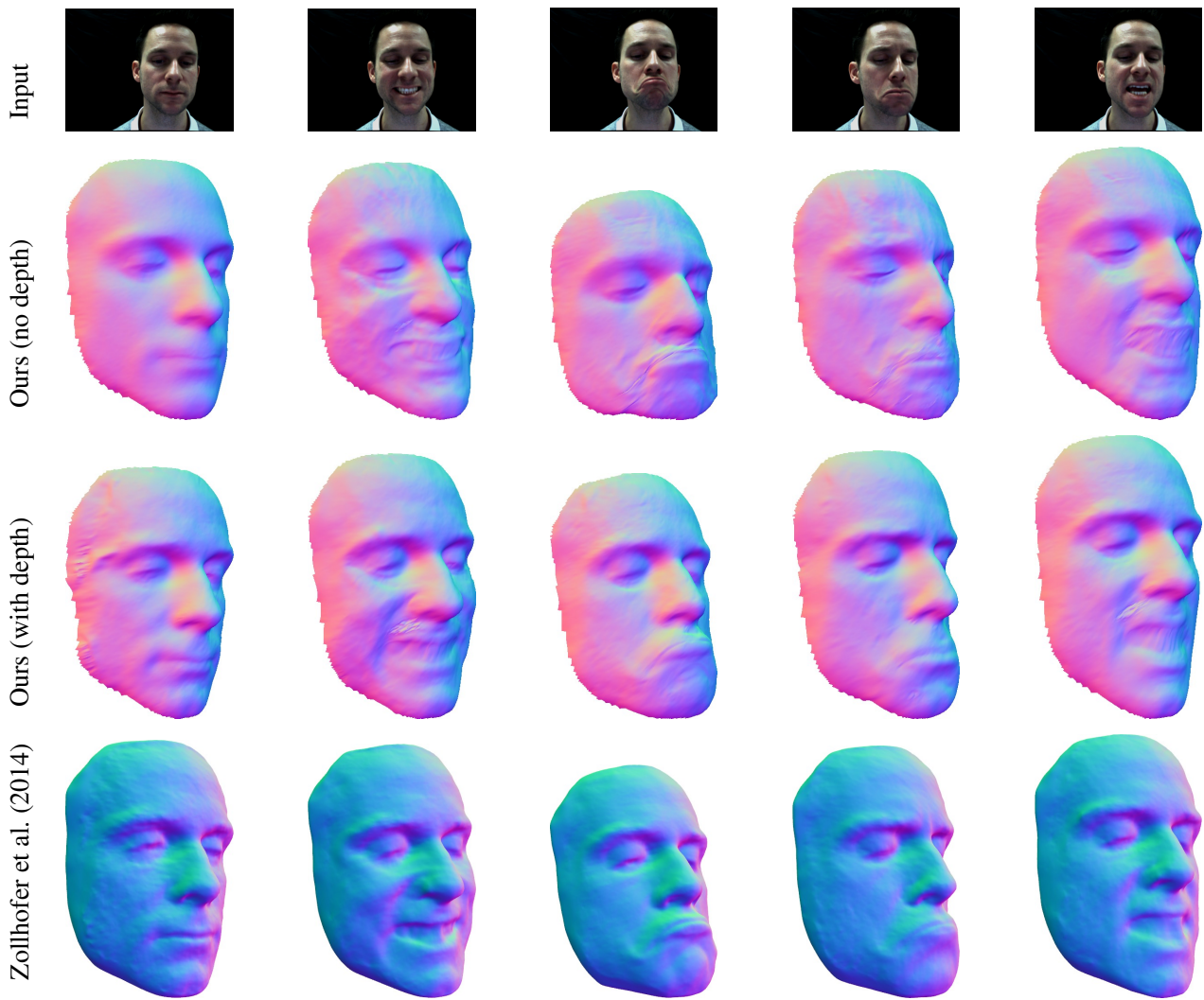

	\bgroup
	\setlength\tabcolsep{7pt}
	\begin{center}
\resizebox{\linewidth}{!}{
  \begin{tabular}{@{\hspace{2pt}}c@{\hspace{6pt}}ccccccc}
    \rotatebox{90}{
    \tiny Input
    }&
       \adjustimage{width=0.13\columnwidth }{color_0020}&
       \adjustimage{width=0.13\columnwidth }{color_0140}&
       \adjustimage{width=0.13\columnwidth }{color_0250}&
       \adjustimage{width=0.13\columnwidth }{color_0294}&
       \adjustimage{width=0.13\columnwidth }{color_0320}
\\
    \rotatebox{90}{
    \tiny Ours (no depth)
    }&
       \adjustimage{width=0.13\columnwidth }{without_depth_normal_0020}&
       \adjustimage{width=0.13\columnwidth }{without_depth_normal_0140}&
       \adjustimage{width=0.13\columnwidth }{without_depth_normal_0250}&
       \adjustimage{width=0.13\columnwidth }{without_depth_normal_0294}&
       \adjustimage{width=0.13\columnwidth }{without_depth_normal_0320}
    \\
    \rotatebox{90}{
    \tiny Ours (with depth)
    }&
       \adjustimage{width=0.13\columnwidth}{with_depth_normal_0020}&
       \adjustimage{width=0.13\columnwidth}{with_depth_normal_0140}&
       \adjustimage{width=0.13\columnwidth}{with_depth_normal_0250}&
       \adjustimage{width=0.13\columnwidth}{with_depth_normal_0294}&
       \adjustimage{width=0.13\columnwidth}{with_depth_normal_0320}
\\
    \rotatebox{90}{
    \tiny ~\cite{kinect_nr:siggraph2014}
    }&
       \adjustimage{width=0.13\columnwidth }{siggraph14_normal_0020}&
       \adjustimage{width=0.13\columnwidth }{siggraph14_normal_0140}&
       \adjustimage{width=0.13\columnwidth }{siggraph14_normal_0250}&
       \adjustimage{width=0.13\columnwidth }{siggraph14_normal_0294}&
       \adjustimage{width=0.13\columnwidth }{siggraph14_normal_0320}
\\
		\end{tabular}}
	\end{center}
	\egroup
	\caption{\label{fig:depth_face}{\bf Depth Fusion:} To illustrate that our
    approach to \nrsfm\ can be fused with other depth cues, we show results from
    our method both with and without data from a depth camera. The results show
    a comparison between the results of \cite{kinect_nr:siggraph2014} and our
    methods with and without depth data on a real face sequence taken from
    \cite{kinect_nr:siggraph2014}. Owing to the strong illumination effects,
    missing data and strong specularities, a direct comparison with
    \cite{Yu_2015_ICCV} was not possible, as quality of the reconstruction was
    very poor. In contrast, the results of our method using the same RGB input
    only \emph{(Ours (no depth))} provides good quality results. This justifies
    the need to reason jointly about shading, specularities and normals in the
    context of non-rigid tracking from RGB only, as we argue for in this paper.}
\end{figure*}

Qualitatively improved results from using depth, and a comparison
against the method of \citet{kinect_nr:siggraph2014}, which only makes
use of frame-to-frame color based tracking, rather than a \sfs model
can be seen in figure \ref{fig:depth_face}. The strong illumination
effects of this sequence prevent us from a direct comparison with any
method such as \citet{Yu_2015_ICCV} that makes use of sequence based
color consistency. More details about the experimental setup are given
in section \ref{sec:color-depth-camera}. With no better method
available (all methods make use of RGB-D) data, a quantitative
comparison between \cite{kinect_nr:siggraph2014} and our work is not
possible. However, section \ref{sec:color-depth-camera} discusses how depth images can improve our reconstructions on synthetic sequences against a known ground-truth.

\section{Template Capture}
\label{sec:template-capture}
This section describes how we capture the static geometry and the
reflectance properties of the object of interest -- or in other words
how we build the template model used for tracking. We achieve this by
moving a hand-held camera around the object while it remains rigid for
a few seconds, to observe it from different angles. During the template capture step, we assume that the illumination remains constant.

\begin{algorithm}[tb]
\caption{3D Template acquisition \label{algo:template_acquisition}}
\SetKwInOut{Input}{Input}\SetKwInOut{Output}{Output}
\SetKwInput{Initialization}{Initialization}
\SetKwFunction{Median}{Median}
\Input{ Rigid image subsequence $\{\mathbf{I^f_{rigid}}\}$ $f=1,\cdots,F$}
\Output{3D coordinates of template mesh vertices ${\bf\widehat{S}}=\{{\bf\widehat{S}}_i\}$ and \\Template albedo map ${\widehat{\bm{\rho}}} = \{{\widehat{\bm{\rho}}}_i\}$ where $i = 1 \cdots N$}
Obtain rigid camera poses for each frame $\{\mathbf{I^f_{rigid}}\}$ using VisualSFM  \cite{wu2011visualsfm}\\
Estimate 3D template mesh vertices ${\bf\widehat{S}}=\{\bs^t_i\}$ using \mvs\ \cite{Campbell:etal:ECCV2008,vogiatzis07pami,hernandez07}\\
Estimate diffuse component ${\bf\widehat{\mathbf{I}}^{diff}_i} ~~ \forall i$ template vertices as median color over all
frames\\
Solve for the illumination coefficients ${\bf\widehat{\mathbf{l}}}$ minimizing \eqref{eq:sh_coeff_estimation} assuming white albedo\\
Solve for albedo map of the template ${\widehat{\bm{\rho}}} = \{{\widehat{\bm{\rho}}}_i\}$ minimizing \eqref{eq:albedo_estimation}
\end{algorithm}

\subsubsection*{ Template geometry:}
For each frame we estimate the relative pose of the camera with
respect to the object using a standard off-the-shelf structure from
motion approach (VisualSFM \cite{wu2011visualsfm}). We then use the
multi-view stereo approach of \cite{Campbell:etal:ECCV2008} to produce
individual depth maps for each frame. Finally the depth maps are fused
using the volumetric technique of \cite{vogiatzis07pami} and the
probabilistic visibility approach of \cite{hernandez07} to produce as
output a watertight mesh of the template shape parameterized as the
set of 3D vertex coordinates
$\bf{\widehat{S}}=\{\bf{\widehat{s}_i}\}$, $i = 1..N$.

\subsubsection*{ Template reflectance properties:}
The next step of the template acquisition stage is to assign a color
value to each vertex $i$ on the mesh. Our implicit assumption is that
the light leaving the surface of the template is the sum of a
viewpoint-independent diffuse term and a view-dependent specular
term. We estimate the diffuse term ${\bf\widehat{\mathbf{I}}^{diff}_i}$
as the median color over all the frames in the rigid subsequence in
which the projected vertex is visible. While some previous approaches
favored the use of the minimum observed intensity
value \citep{nishino2001determining}, we choose to use the median as
proposed by \cite{wood2000surface} (also used in the color template generation of \cite{Yu_2015_ICCV}) since it provides
robustness to shadows and errors in the camera tracking.

We decompose the diffuse component of the template
${\bf\widehat{\mathbf{I}}^{diff}_i}$ further into the product of an
\emph{albedo map} and an irradiance function parameterized in terms of
\emph{spherical harmonics} to approximate the illumination and the
surface normals as described in~\eqref{eq:full_reflection_model_final}.
%
%
First we solve for the \emph{spherical harmonic coefficients} by optimizing the
following photometric objective function with respect to $\mathbf{\hat{l}}$:

\begin{equation}\label{eq:sh_coeff_estimation}
E_\textnormal{template}(\mathbf{\hat{l}}) =  \sum_{i\in\mathcal{V}} {\left \lVert {\bf\widehat{\mathbf{I}}^{diff}_i} - \boldsymbol{\hat{\rho}_i} \mathbf{l}\cdot Y \left(\mathbf{n}_i(\bf{\widehat{S}}) \right) \right \rVert_\epsilon}
 \end{equation}
where $\boldsymbol{\hat{\rho}_i}$ is an initial assumption of the albedo map
(e.g. white, uniform color, or the result from k-means).

%
%
%
The \emph{albedo map} is estimated by minimizing the same cost with a small
variant -- we give a lower confidence to points with low
shading. Also, a weighted local smoothing term is added based on the
difference in intensity.
\begin{align}\label{eq:albedo_estimation}
  E'_\textnormal{template}(\boldsymbol{\hat{\rho}}) = &\sum_{i\in\mathcal{V}}{w^a_i\left\|
                                             {\bf\widehat{\mathbf{I}}^{diff}_i} -
                                             \boldsymbol{\hat{\rho}_i}
                                             r(\mathbf{n}_i(\bf{\widehat{S}}))
                                             \right\|_\epsilon} \\
                                           &+ \sum_{i\in\mathcal{V}}\sum_{j\in\mathcal{N}_i}{w^{a^\prime}_{ij}\left\|  \boldsymbol{\hat{\rho}_i} -  \boldsymbol{\hat{\rho}_j}\right\|_2^2}\nonumber
\end{align}
where $w^a_i= r(\mathbf{n}_i)$ is chosen to decrease the importance placed on
regions of low shading and $w^{a^\prime}_{ij}= \exp \left( -
  \frac{\|\mathbf{I}_i - \mathbf{I}_j\|_2^2}{2\sigma_s^2} \right)$ to encourage
points with similar appearance to share the same albedo.





\section{Experimental Evaluation}
We consider the experimental evaluation on two separate problems:
\emph{(i)} Evaluating our results
when using only an RGB video sequence as input, and \emph{(ii)} Using input from an RGB-D
camera as input.
\subsection{Video Input}
\label{sec:results}
\begin{figure*}[t!]
\resizebox{\linewidth}{!}{
  \begin{tabular}{@{\hspace{2pt}}c@{\hspace{6pt}}cc||cccc}
  	&
  	\multicolumn{2}{c}{\sc{real sequence}}&
  	\multicolumn{4}{c}{\sc{rendered sequence}}\\
    \rotatebox{90}{
      ~~~~~~~~~~~~~Input
    }&
    \includegraphics[width=0.13\columnwidth]{rgb0010}&
    \includegraphics[width=0.13\columnwidth]{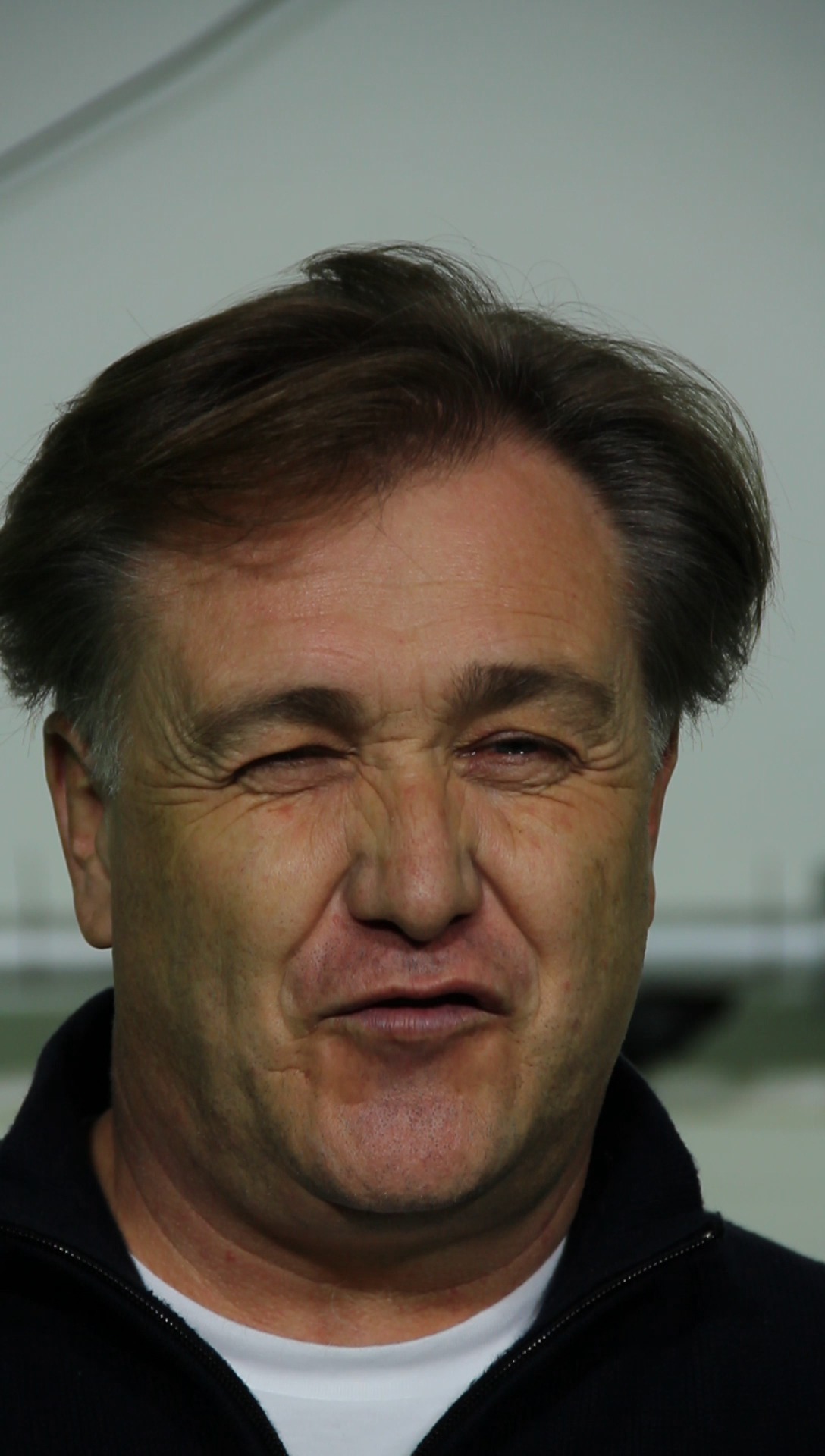}&
    \includegraphics[width=0.13\columnwidth]{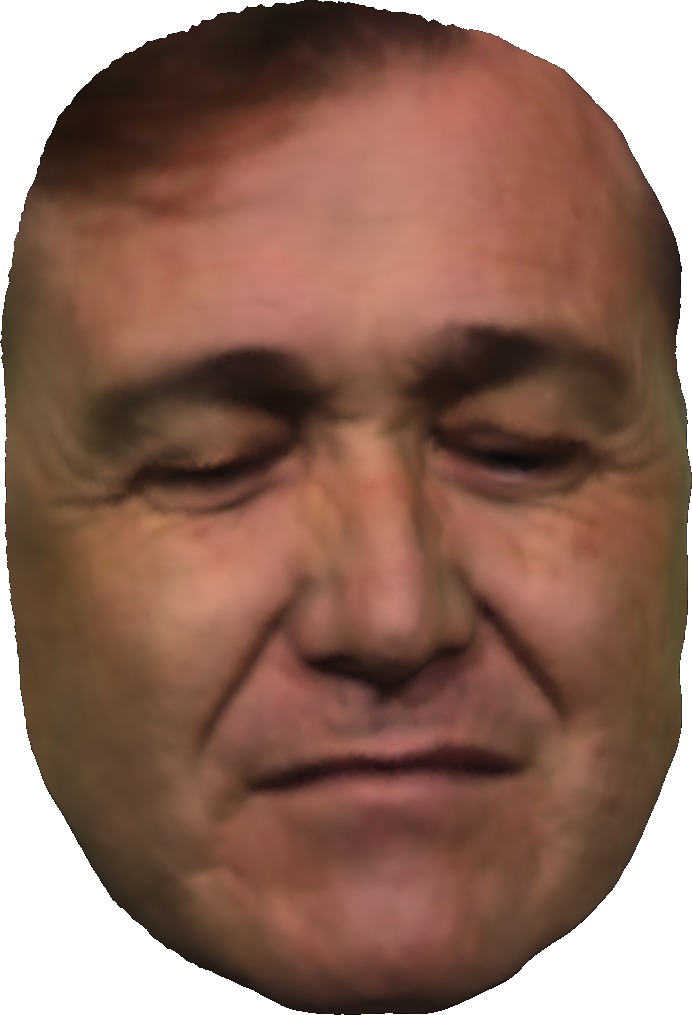}&
    \includegraphics[width=0.13\columnwidth]{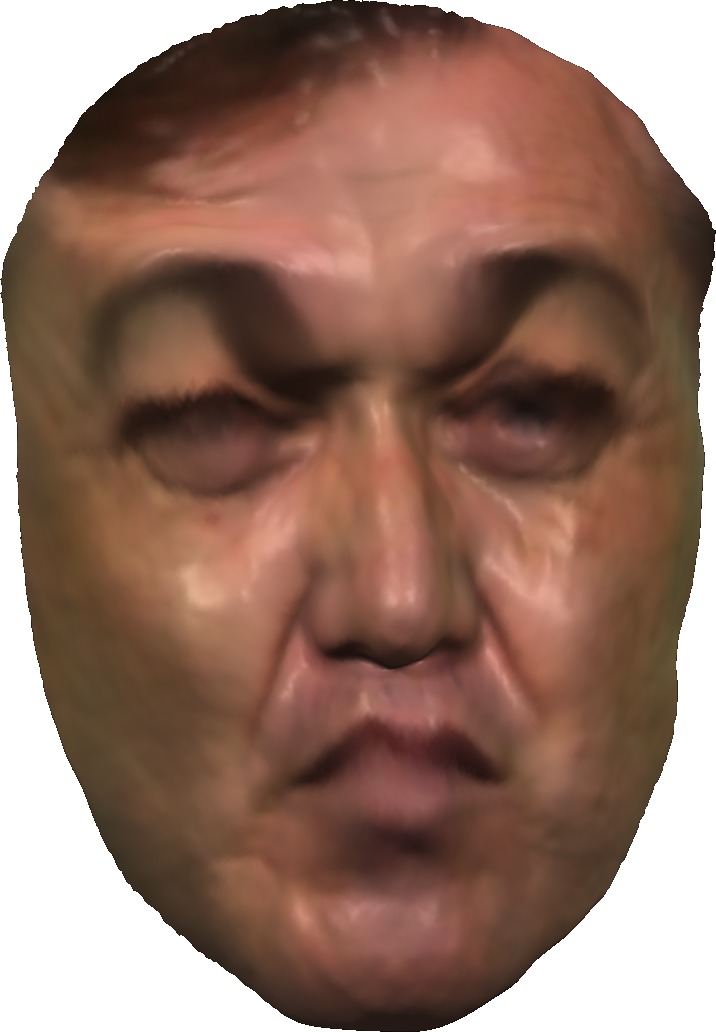}&
    \includegraphics[width=0.13\columnwidth]{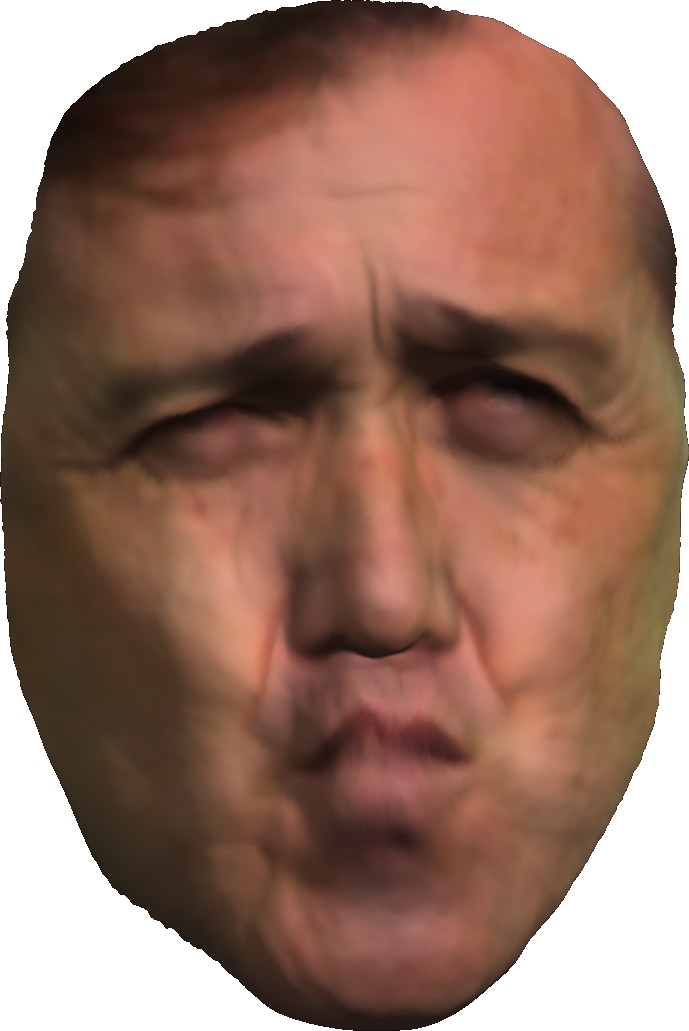}&
    \includegraphics[width=0.13\columnwidth]{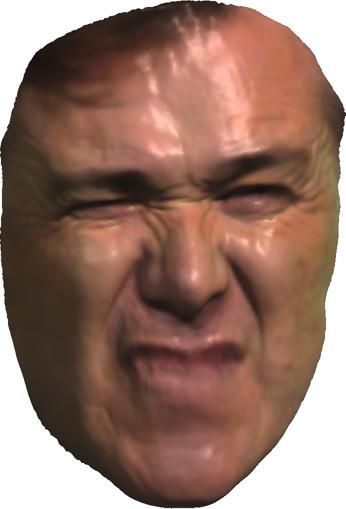}\\
    \rotatebox{90}{
      ~~~~~~~Yu \etal
    }&
\includegraphics[width=0.10\columnwidth]{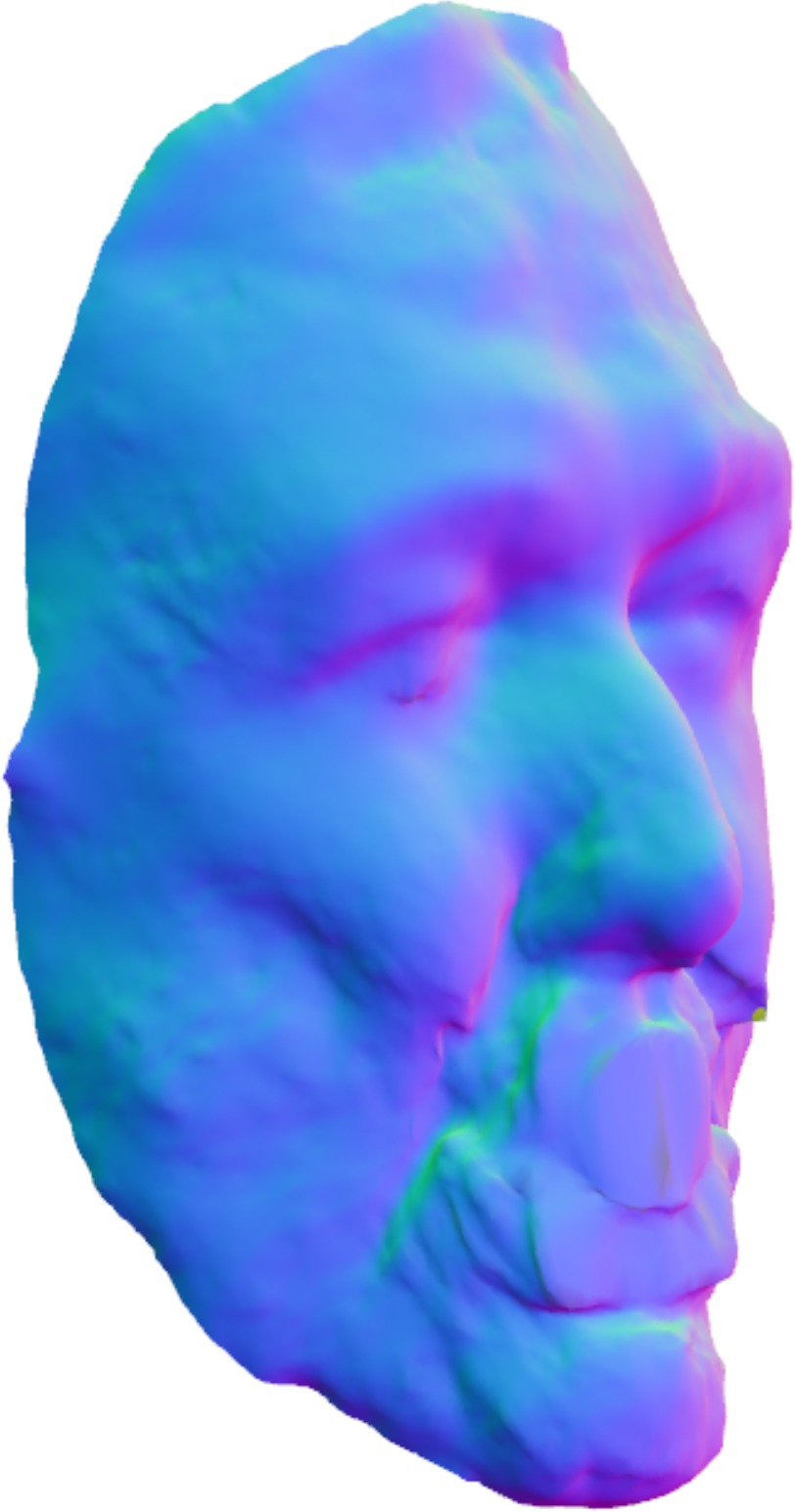}&
    \includegraphics[width=0.13\columnwidth]{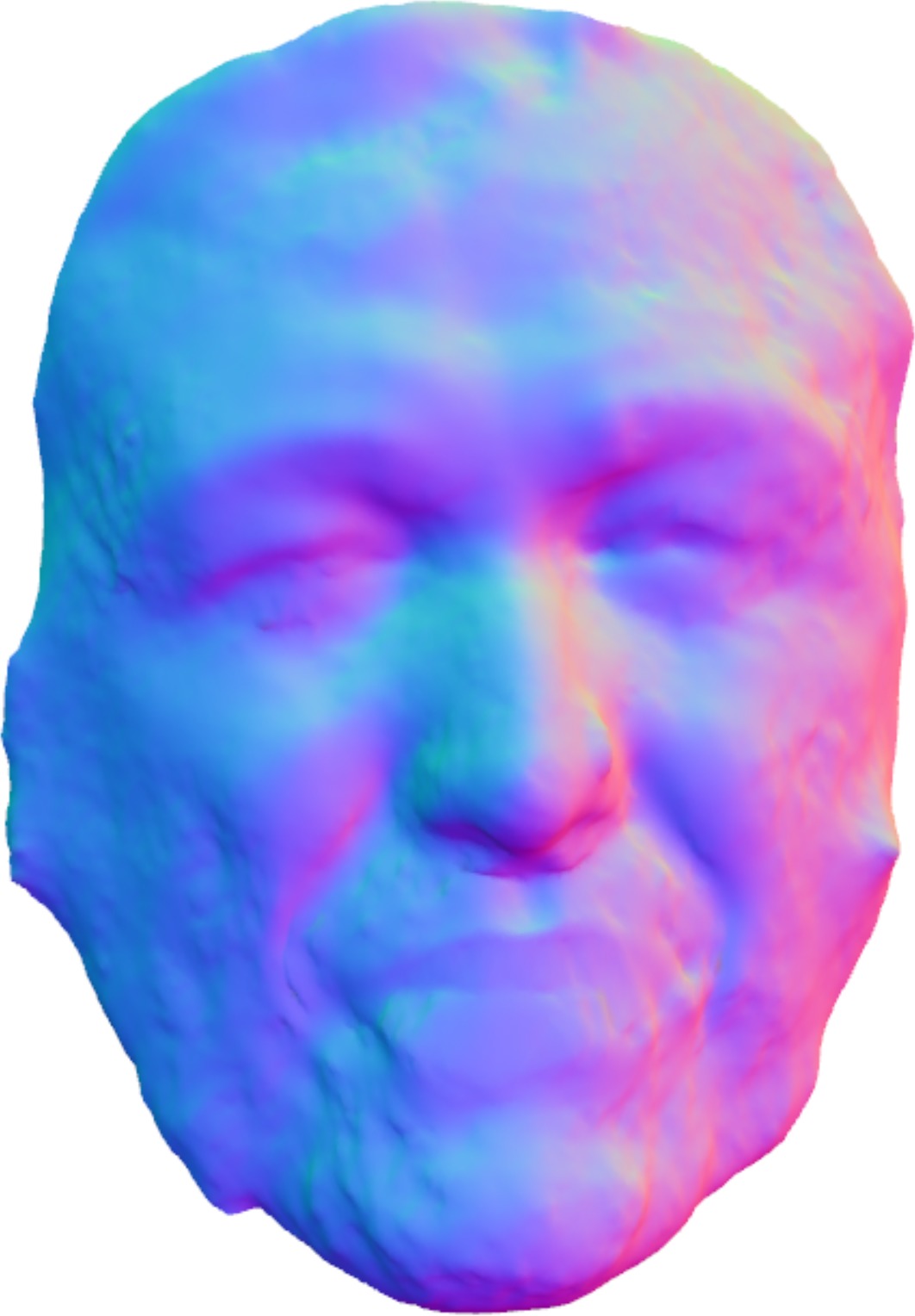}&
    \includegraphics[width=0.13\columnwidth]{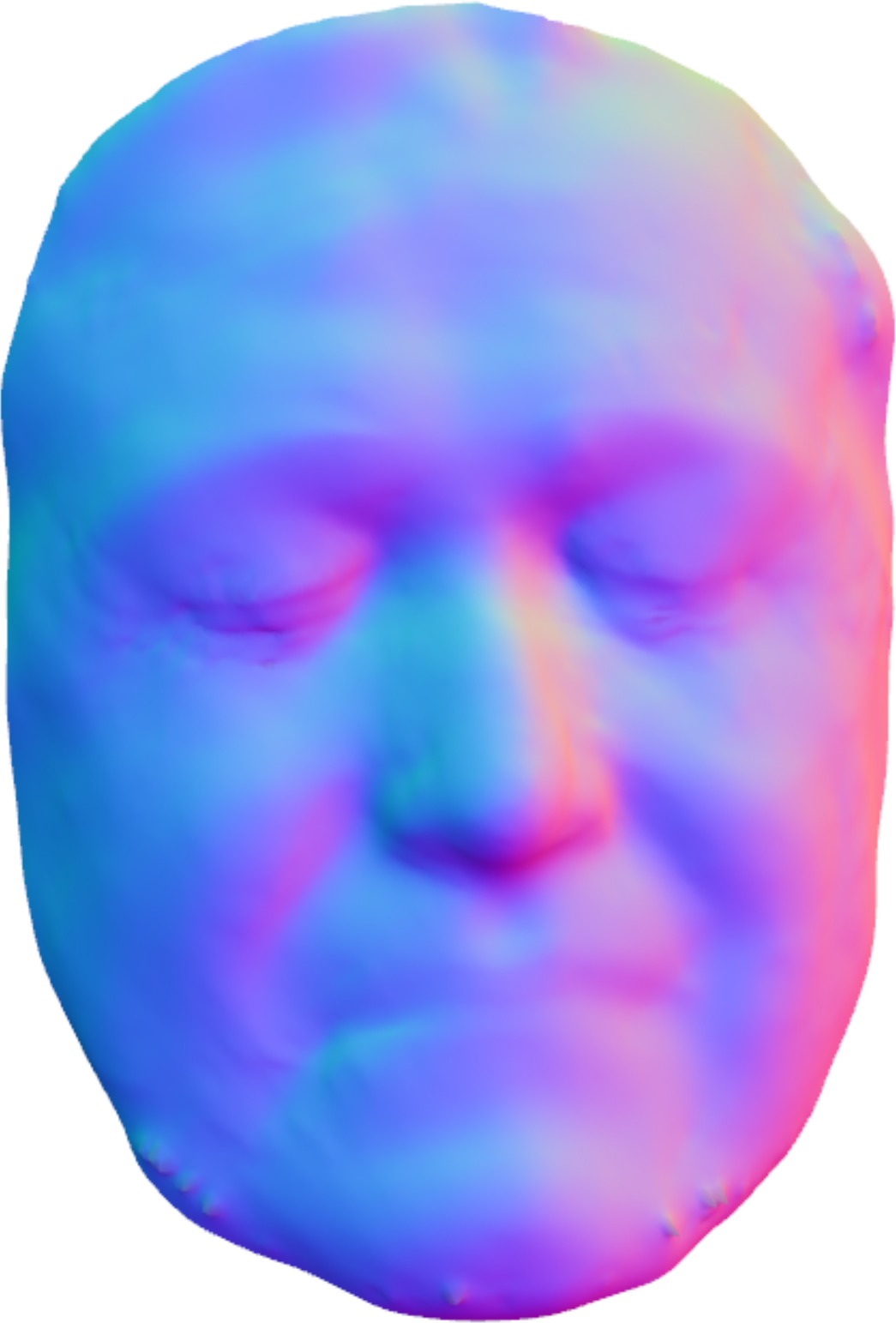}&
    \includegraphics[width=0.13\columnwidth]{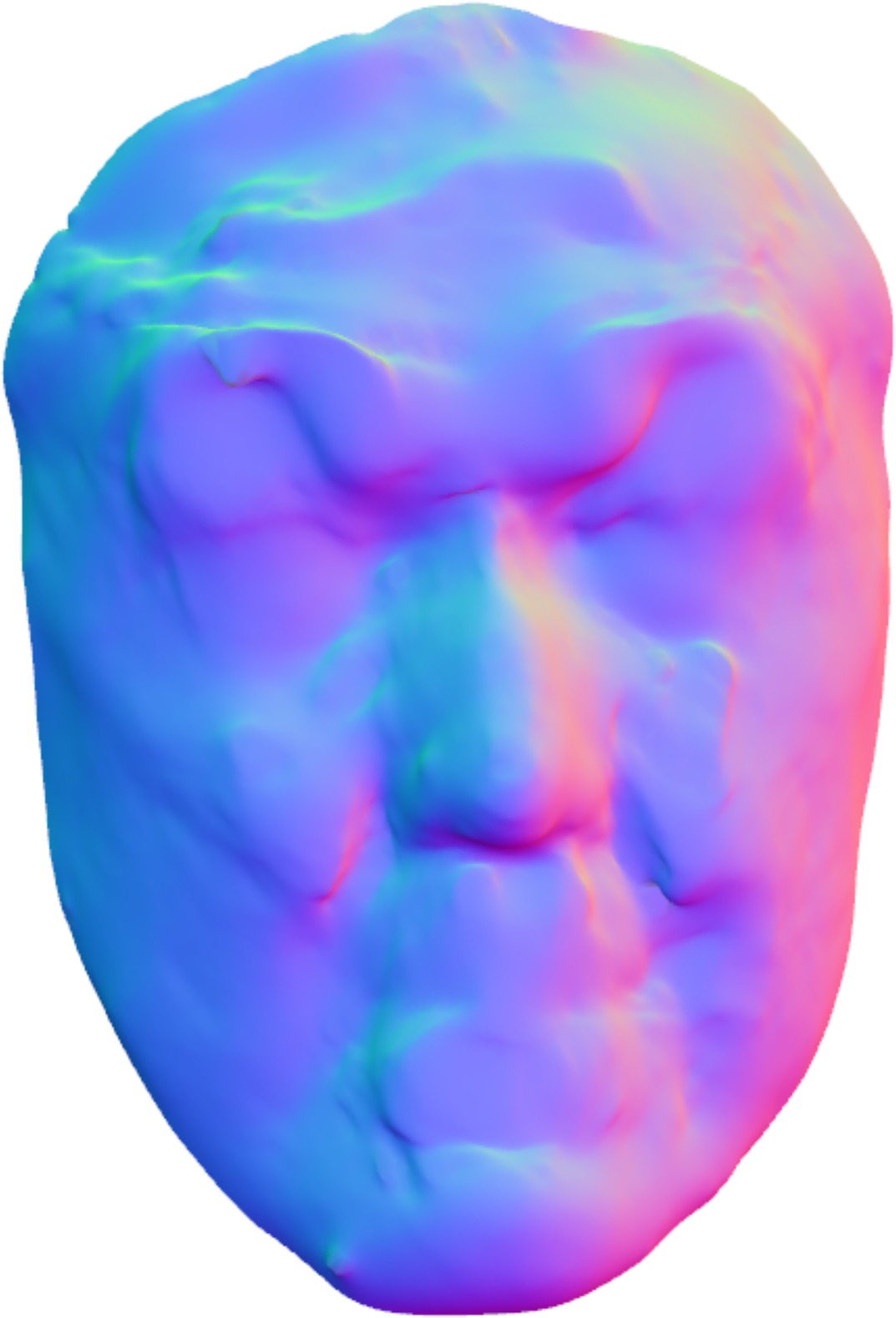}&
    \includegraphics[width=0.13\columnwidth]{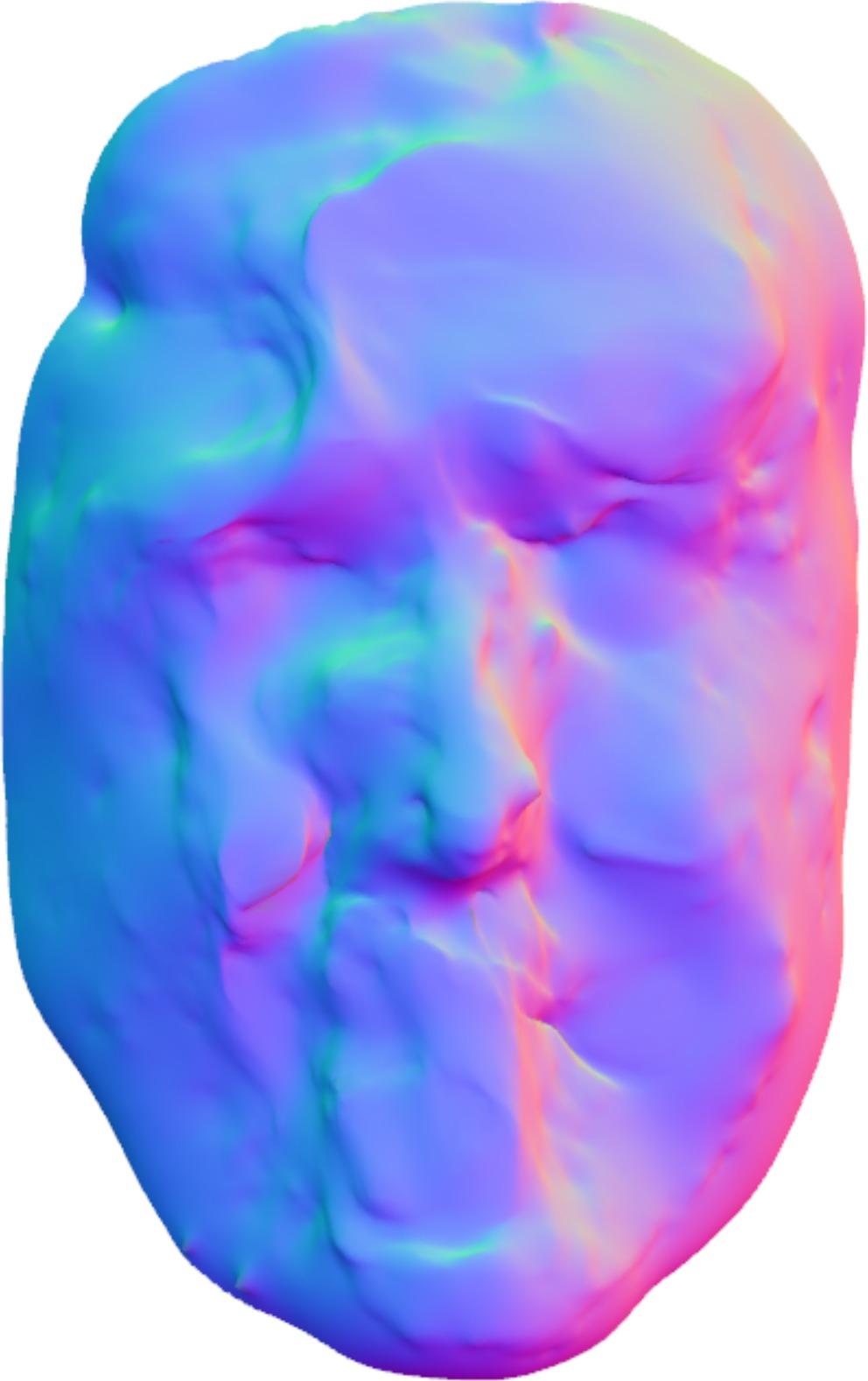}&
    \includegraphics[width=0.13\columnwidth]{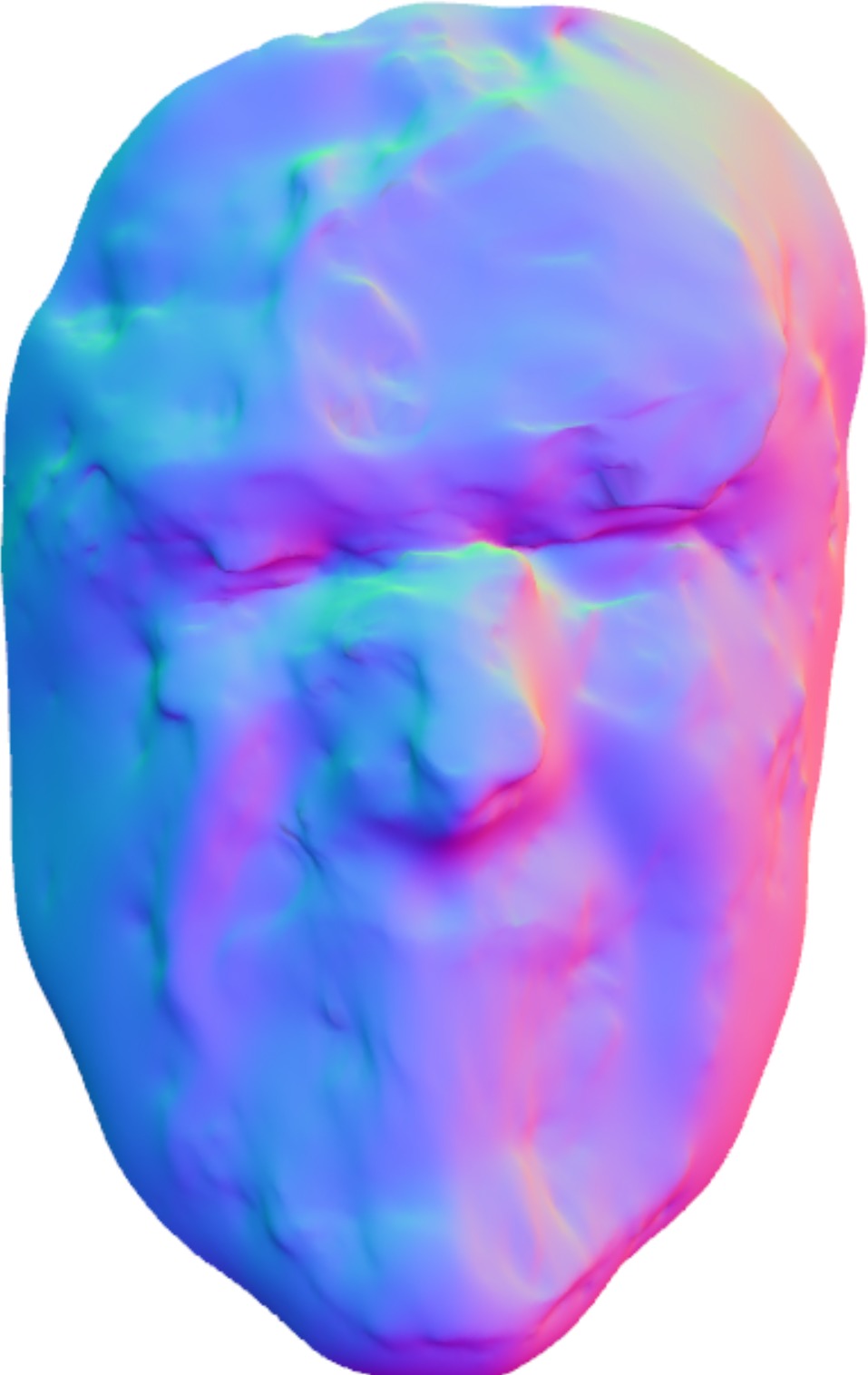}\\
    \rotatebox{90}{
      ~~~~~~~~~Ours
    }&
\includegraphics[width=0.10\columnwidth]{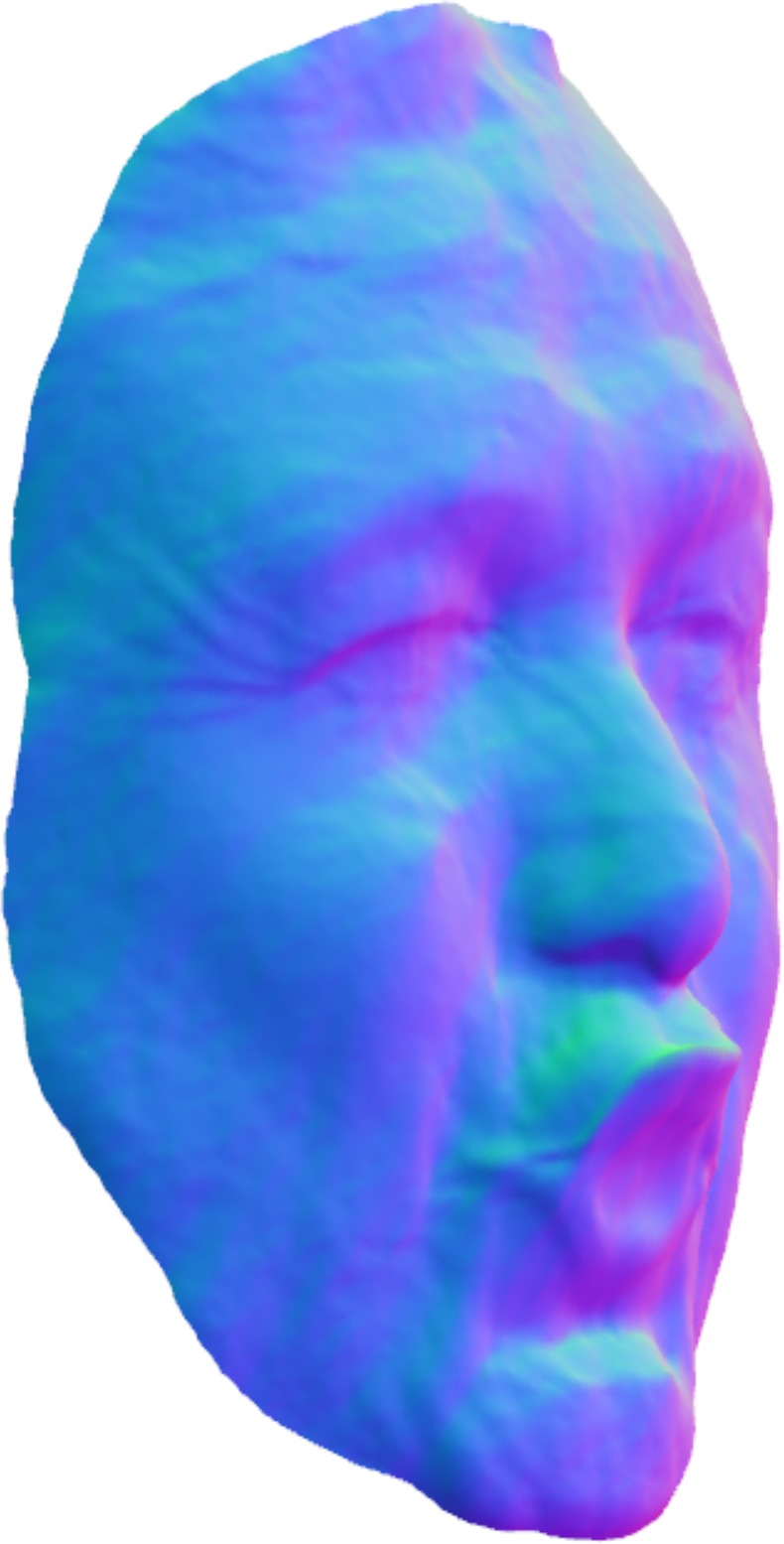}&
    \includegraphics[width=0.13\columnwidth]{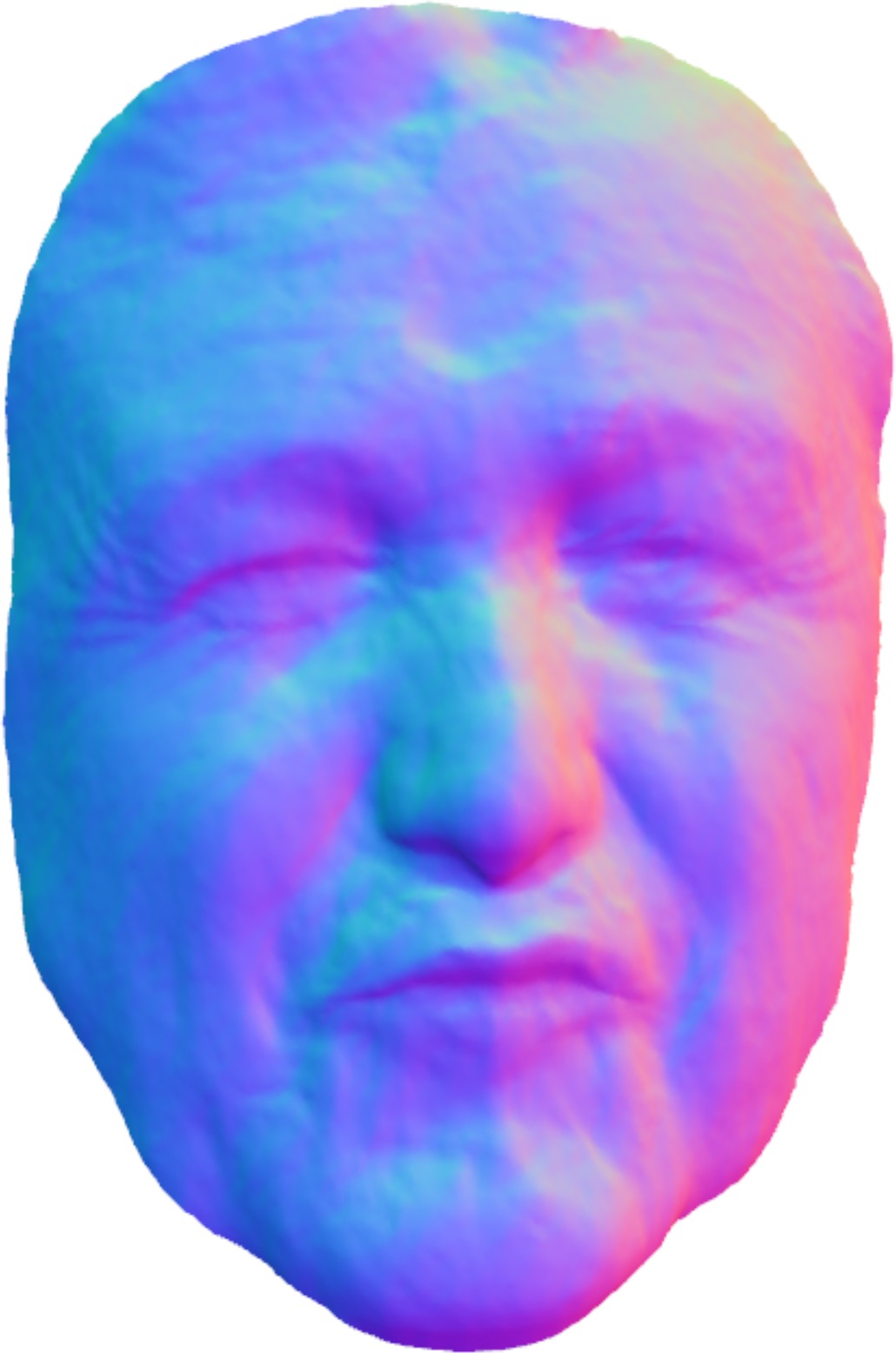}&
    \includegraphics[width=0.13\columnwidth]{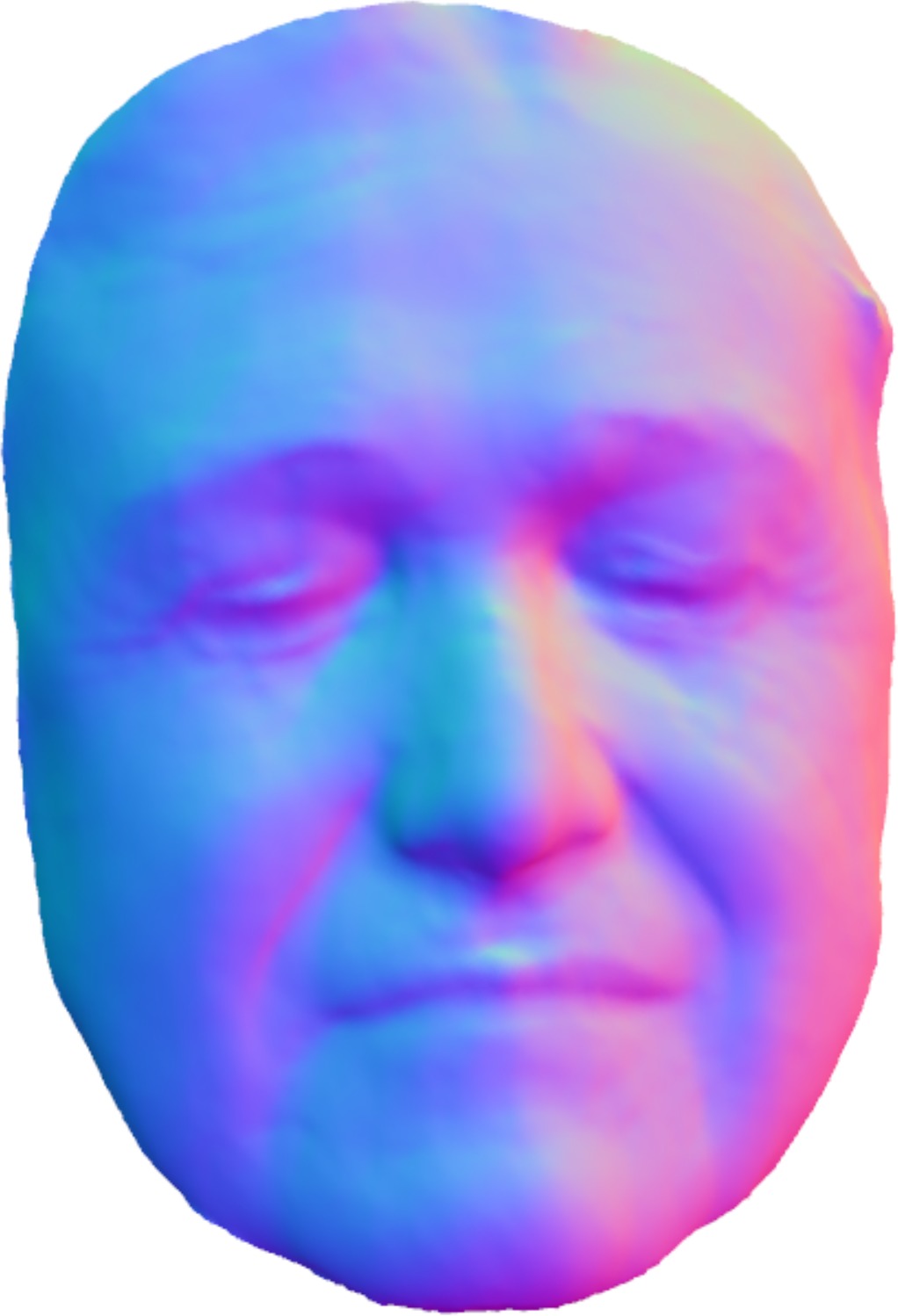}&
    \includegraphics[width=0.13\columnwidth]{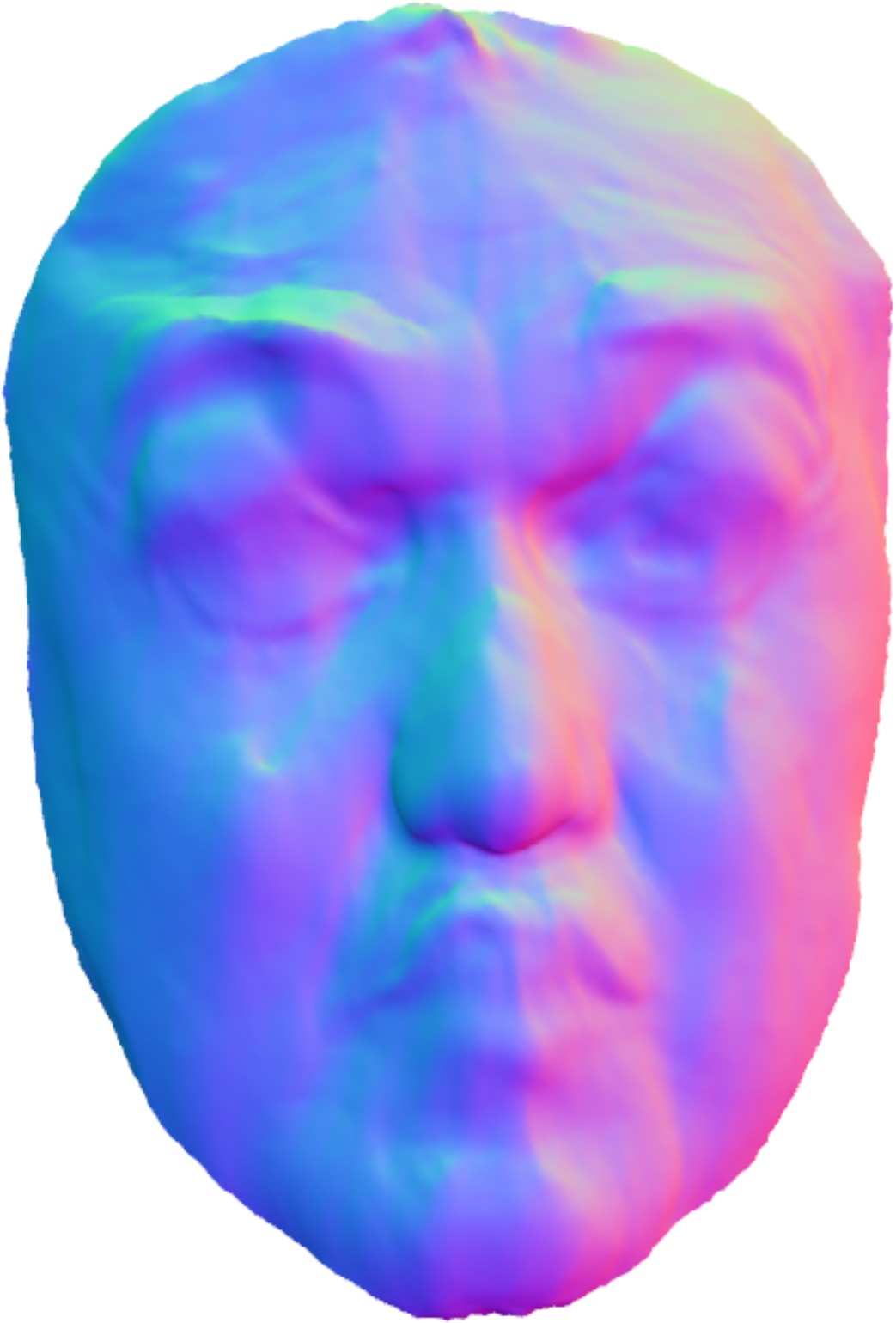}&
    \includegraphics[width=0.13\columnwidth]{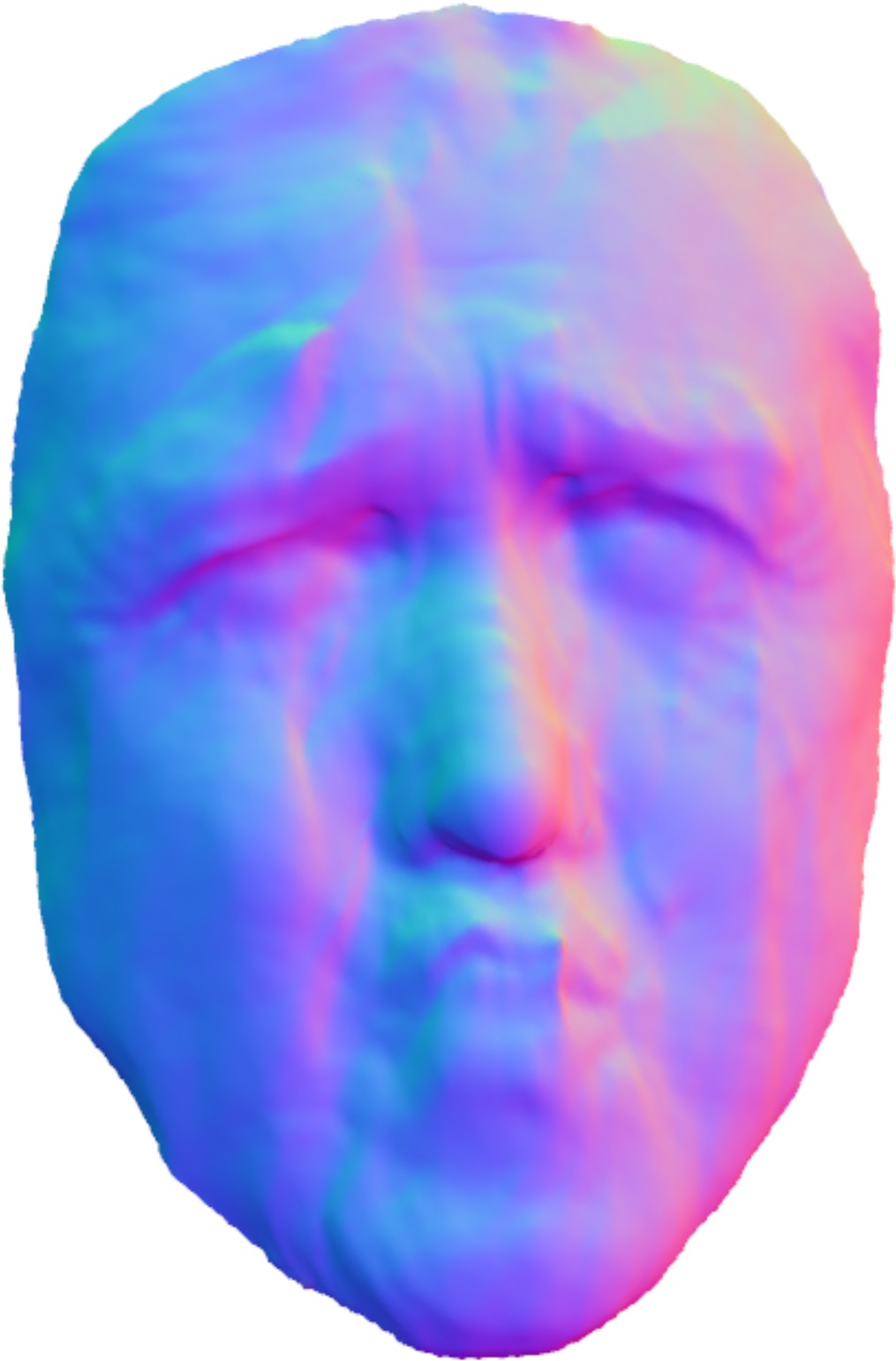}&
    \includegraphics[width=0.13\columnwidth]{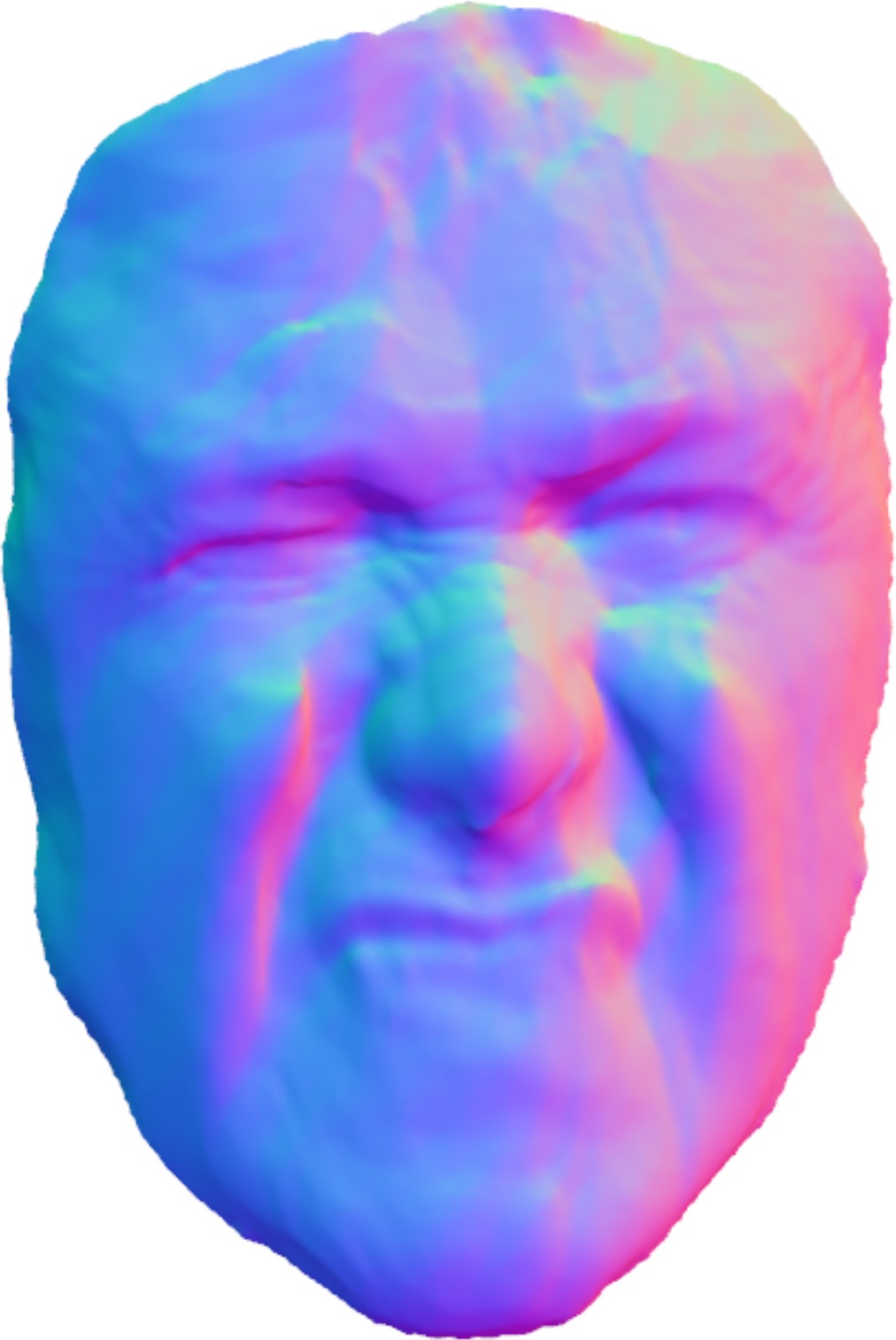}\\
  \end{tabular}}
\caption{\label{fig:orig}Best viewed in color: A qualitative comparison of our
  method against that of \cite{Yu_2015_ICCV}, on {\sc{(left)}} the real-world
  sequence of \cite{Valgaerts_2012_SIGGRAPH} and {\sc{(right)}} our four
  rendered sequences (from left to right: LF, SF, LC and SC).}
\end{figure*}

\subsubsection{Experiments on sequences with ground truth 3D}

We tested the proposed method on ground truth sequences that we
generated using the public dataset of 3D face shapes from
\cite{Valgaerts_2012_SIGGRAPH}, down-sampled ten times in order to
reduce the runtime of our method. These 3D shapes were reconstructed
by~\cite{Valgaerts_2012_SIGGRAPH}, using their high end stereo algorithm,
from a real video sequence of an actor and were made publicly
available. Using these 3D shapes as ground truth geometry in
conjunction with the original video we render various ground truth
sequences under different lighting conditions and surface properties.

We assign a constant albedo over time to each of the vertices of the
mesh, which we estimated using the 3D ground truth shape associated
with the first frame in a similar way as described in section
\ref{sec:template-capture}. Using this albedo map, we render four
different scenarios combining Lambertian or specular surfaces on a
scene lit by two white directional lights with constant or changing
intensity. Thus, we have sequences with a perfect Lambertian surface
with fixed (LF) or changing illumination (LC), and a specular surface
with fixed (SF) or changing illumination (SC) (see figures
\ref{fig:teaser} and \ref{fig:orig}). To test these ground truth
sequences we use a coarse-to-fine pyramid with a template mesh of
$\sim 6,000$ vertices at the coarsest level and, for the finest level,
one with the same amount of vertices as the ground truth ($\sim
24,000$ vertices). We run our method on a computer with an Intel Core
i7-5930K CPU, which takes around one minute to process each frame.

Table~\ref{table:comparison_with_rui} shows the comparison results
against the recent template-based method of \cite{Yu_2015_ICCV} using
their publicly available
code\footnote{\url{http://visual.cs.ucl.ac.uk/pubs/ddd/}}.  It can be
seen that our proposed approach is significantly better in all four cases,
both  when we do and when we do not model specularities. We reduce the baseline error
by a factor of 220\%-260\% when the specular component is not
estimated and around 240\%-280\% when it is. It should also be noticed
how estimating the specular component improves the results for pure
Lambertian sequences. This is due to the fact that the estimated
specularities are also compensating for the errors in the initially
computed albedo. Figure \ref{fig:decompositionSC} shows the
decomposition of two different input frames of the SC sequence (with a
specular surface and changing illumination) into \emph{albedo,
  shading, specularities, normals and illumination}. We show  how our
method can handle this challenging scenario with significant changes
in intensity. Our results can be best appreciated by watching the
accompanying video\footnote{Please visit
  \url{http://www0.cs.ucl.ac.uk/staff/Qi.Liu/bmvc16/better_together.html}
  to check video results and to access our publicly available code and
  datasets.}.
\begin{table*}[t]
  \caption{\label{table:comparison_with_rui} Comparison of different
    versions of our algorithm with \cite{Yu_2015_ICCV} on $4$
    different synthetic sequences.  We report average RMS error (in
    mm) over all frames w.r.t.  ground truth. The four experiments
    refer either to Lambertian (L) or Specular (S) surfaces with fixed
    (F) or changing (C) lighting.
  \label{table:additional_comparison} We also show a quantitative
  comparison between different versions of our own algorithm. Our
  results can be improved by joint refinement of all energy terms, and
  the use of depth maps. Note that joint refinement leads to a
  substantial increase in robustness, that can be seen not just in the
  better results but also in the slower rate of degradation, as we
  move to more challenging sequences. As expected, the use of RGB-D
  data as input gives a substantial improvement over RGB only.}
\begin{center}
\begin{tabular}{|c||C{1.5cm}|C{1.5cm}|C{1.5cm}|C{1.5cm}|}
  \hline
  & LF (mm) & SF (mm) & LC (mm) & SC (mm) \\ \hline
  \cite{Yu_2015_ICCV}   & 7.29 & 7.93 & 9.18  & 9.28   \\ \hline\hline
  Ours without specularities  & 2.91 & 3.28 &  3.50  & 4.21 \\ \hline
  Ours with specularities  & 2.73 & 2.89 &  3.42  & 3.84 \\ \hline
  Ours With Joint Refinement         & 2.71&2.80&3.11&3.01\\\hline\hline
  Ours With Depth                   &1.73&1.79&1.71&1.81\\\hline
  Ours With Depth \& Joint Refinement&1.64&1.74&172&1.77 \\ \hline
\end{tabular}
\end{center}
\end{table*}
\subsubsection{Experiments on real sequences}

\begin{figure}[t!]
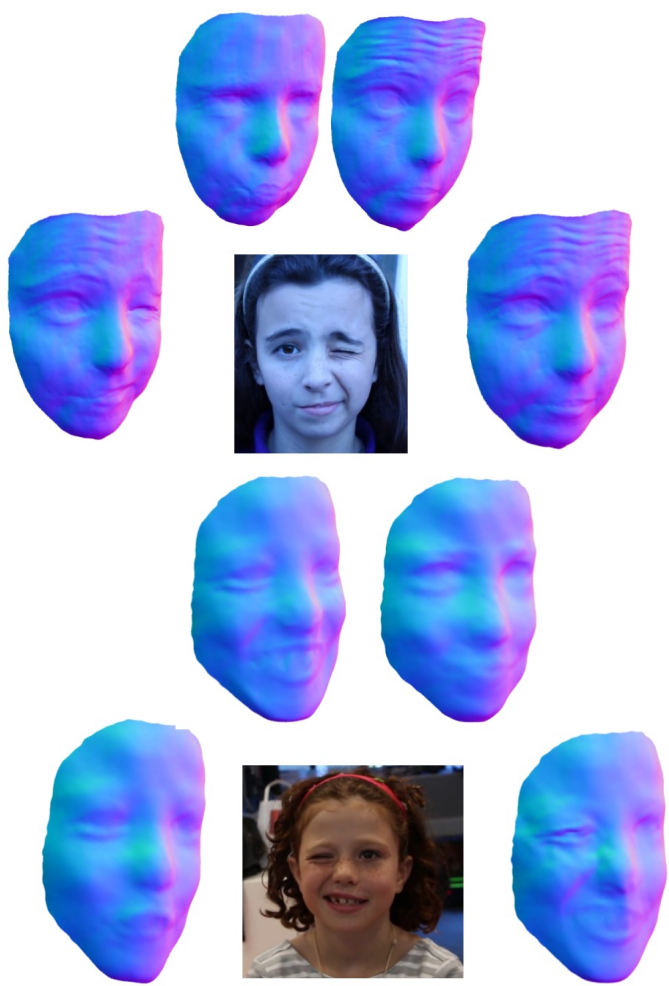

\centering
\adjustimage{width=\columnwidth, valign=m}{Blanca_reconstructions}
\adjustimage{width=1.1\columnwidth, valign=m}{Adela_reconstructions}
\caption{\label{fig:Blanca_y_Adela}Example reconstructions on real sequences with strong deformations.}
\end{figure}

\begin{figure*}[t!]
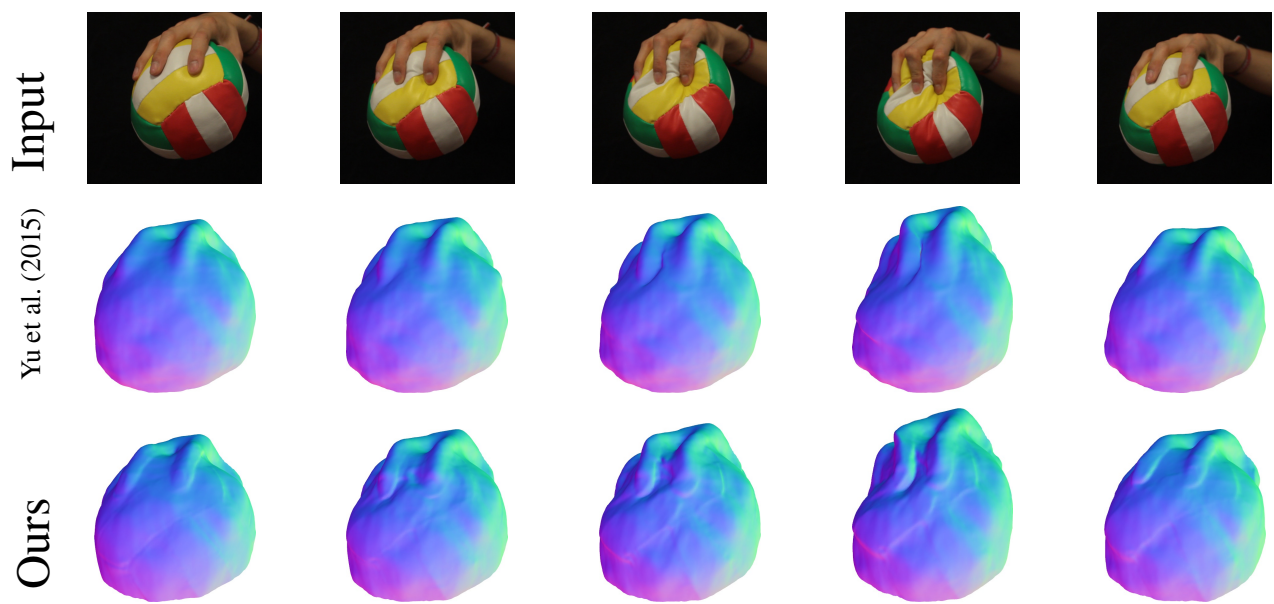

  \bgroup
  \setlength\tabcolsep{7pt}
  \begin{center}
    \resizebox{\linewidth}{!}{
      \begin{tabular}{@{\hspace{2pt}}c@{\hspace{6pt}}cccccc}
        \rotatebox{90}{
        Input
        }&
         \adjustimage{width=0.13\columnwidth}{rgb0001-trim}&
	\adjustimage{width=0.13\columnwidth}{rgb0055-trim}&
	\adjustimage{width=0.13\columnwidth}{rgb0065-trim}&
	\adjustimage{width=0.13\columnwidth}{rgb0080-trim}&
	\adjustimage{width=0.13\columnwidth}{rgb0100-trim}
	\\
        \rotatebox{90}{
        \tiny	\cite{Yu_2015_ICCV}
        }&
	\adjustimage{width=0.12\columnwidth}{ball_rui1_cropped-trim}&
	\adjustimage{width=0.12\columnwidth}{ball_rui55_cropped-trim}&
	\adjustimage{width=0.12\columnwidth}{ball_rui65_cropped-trim}&
	\adjustimage{width=0.12\columnwidth}{ball_rui80_cropped-trim}&
	\adjustimage{width=0.12\columnwidth}{ball_rui100_cropped-trim}
	\\
        \rotatebox{90}{
        Ours
        }&
	\adjustimage{width=0.12\columnwidth}{ball_qi1-trim}&
	\adjustimage{width=0.12\columnwidth}{ball_qi55-trim}&
	\adjustimage{width=0.12\columnwidth}{ball_qi65-trim}&
	\adjustimage{width=0.12\columnwidth}{ball_qi80-trim}&
	\adjustimage{width=0.12\columnwidth}{ball_qi100-trim}
      \end{tabular}}
  \end{center}
  \egroup
  \caption{\label{fig:ball}Comparison between the results of \cite{Yu_2015_ICCV}
    and our method on a real sequence of a hand holding and deforming a ball. }
\end{figure*}

We further evaluate qualitatively on  real sequences: the
original face from \cite{Valgaerts_2012_SIGGRAPH}\
(figure~\ref{fig:orig}), the face from \cite{Yu_2015_ICCV}\ (figure
\ref{fig:ivan}), and new sequences of a hand deforming a ball
(figure~\ref{fig:ball}) and two child faces (figure~\ref{fig:Blanca_y_Adela}). In the case the first sequence (figure~\ref{fig:orig}), notice the
improvement on the reconstructed deformation of the mouth thanks to
our diffuse shading model while \cite{Yu_2015_ICCV} only able to
recover a flat surface. We do not show a comparison with \cite{Yu_2015_ICCV} on the two sequences of the child faces (figure~\ref{fig:Blanca_y_Adela}) as the reconstructions were of poor quality given the specularities and strong deformations.    


\subsubsection{Influence of the template quality}
\label{sec:templ-qual-begint}

\begin{figure*}[t!]
\resizebox{\linewidth}{!}{
  \begin{tabular}{@{\hspace{2pt}}c@{\hspace{6pt}}ccccc}

$\sigma=0$ & $\sigma=0.001$ & $\sigma=0.005$ & $\sigma=0.008$ & $\sigma=0.01$ \\

       \includegraphics[width=0.2\columnwidth]{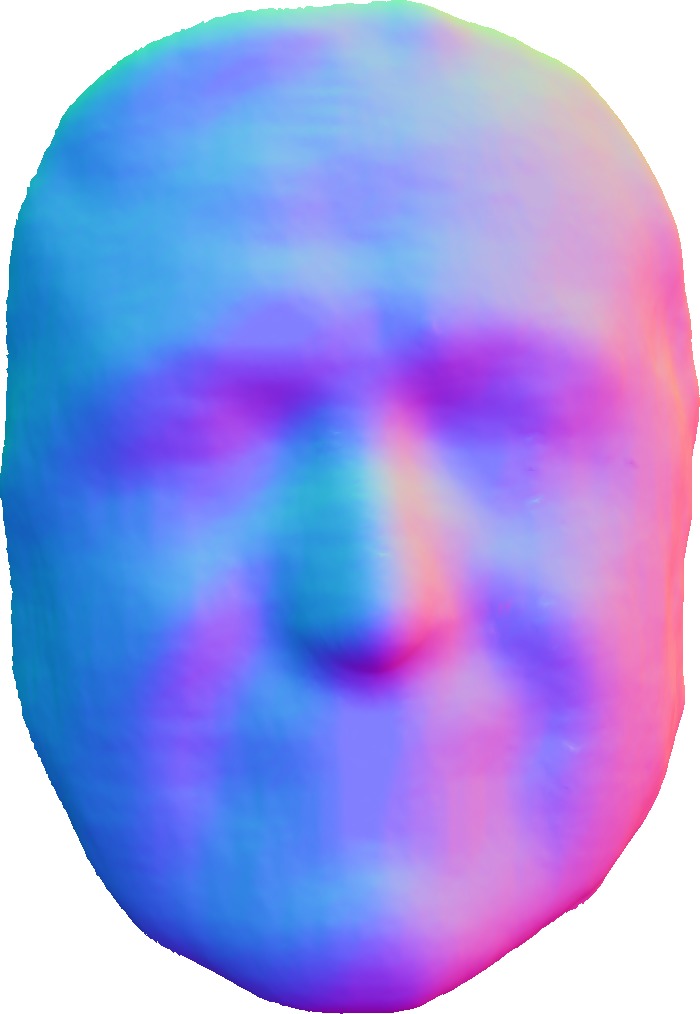}&
       \includegraphics[width=0.2\columnwidth]{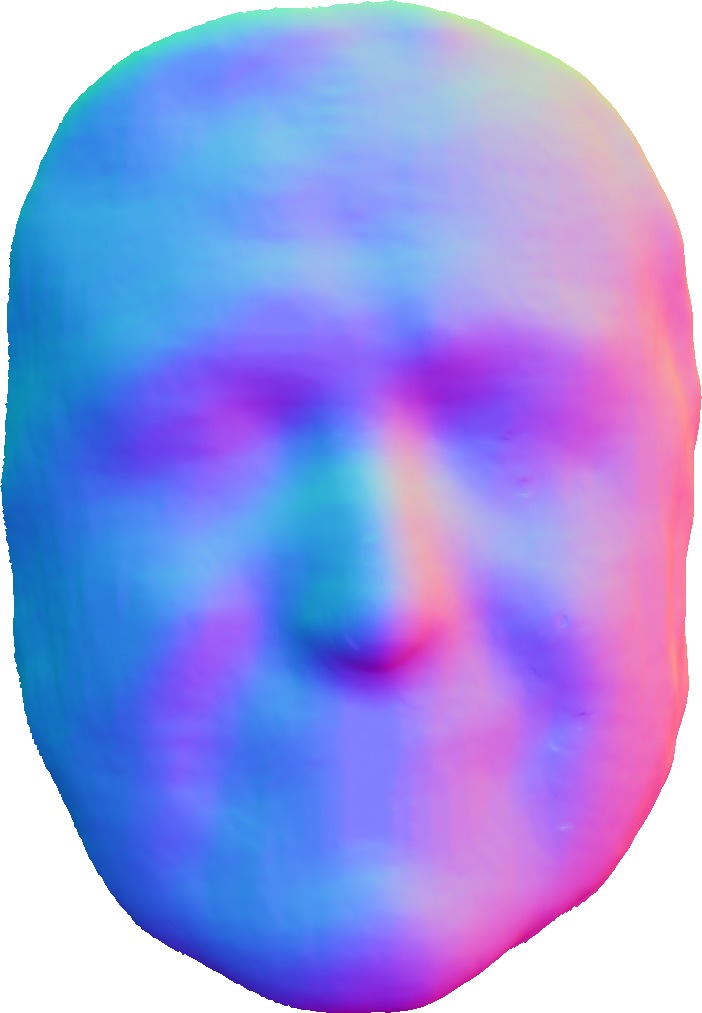}&
       \includegraphics[width=0.2\columnwidth]{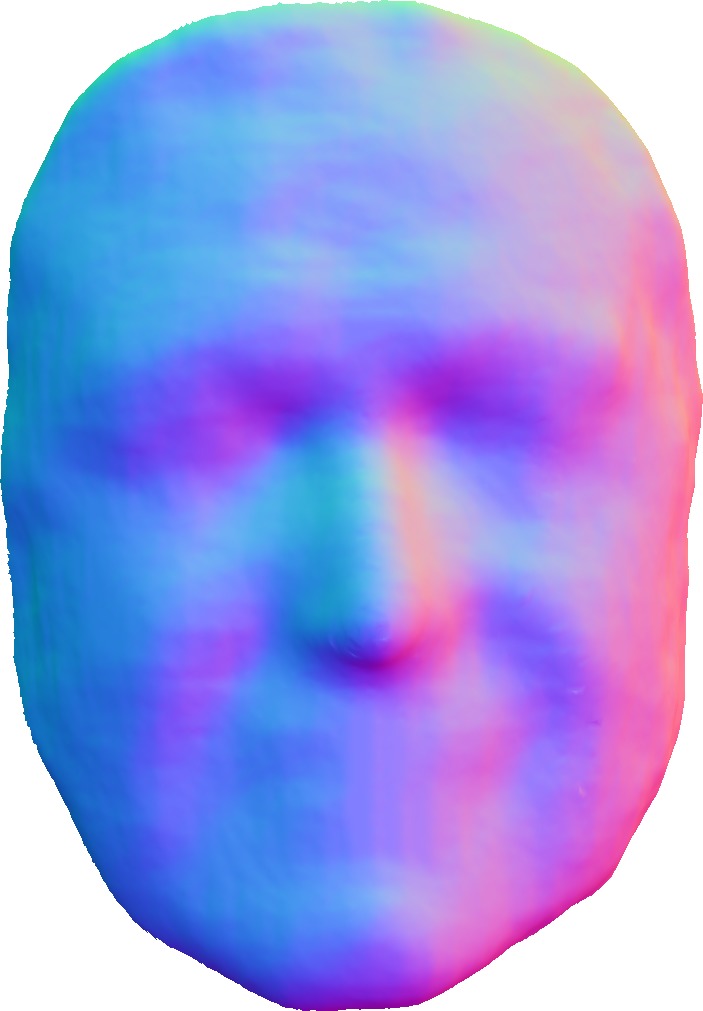}&
       \includegraphics[width=0.2\columnwidth]{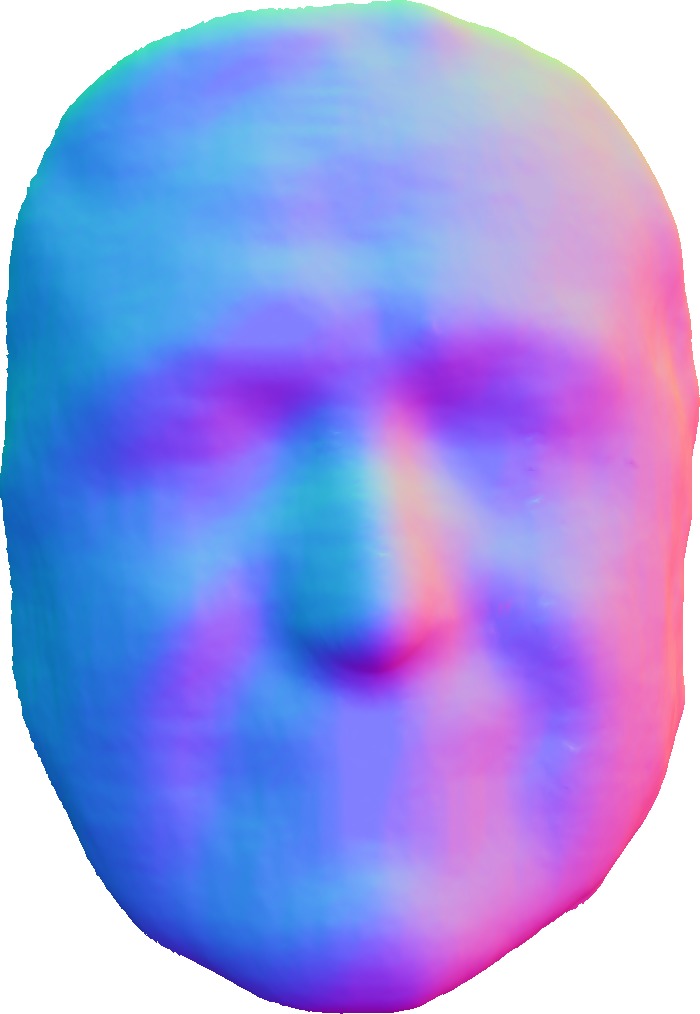}&
       \includegraphics[width=0.2\columnwidth]{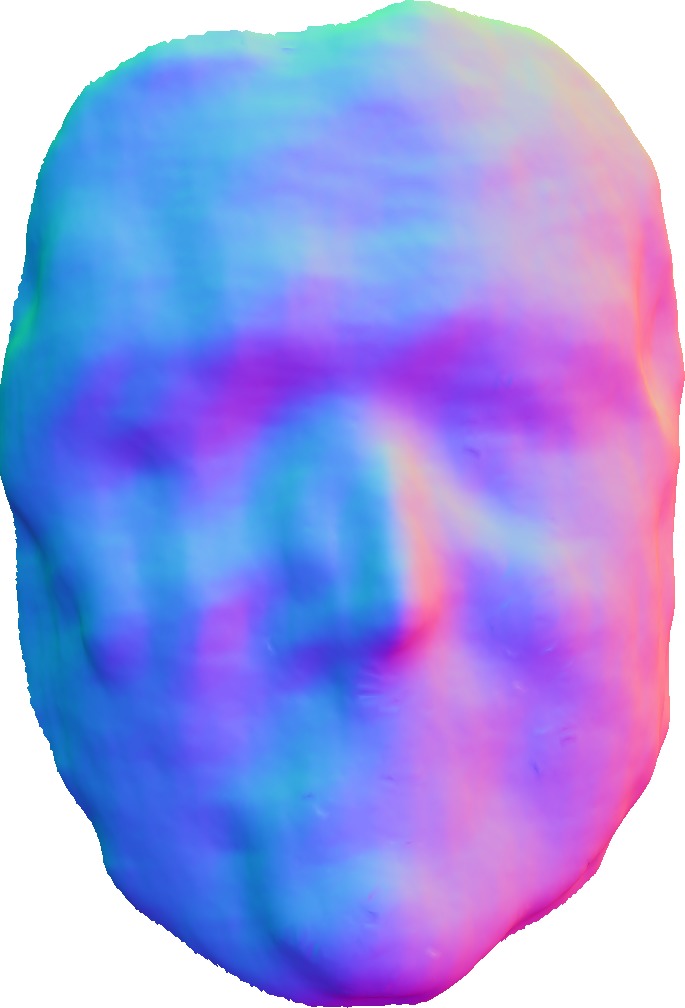}\\

\includegraphics[width=0.2\columnwidth]{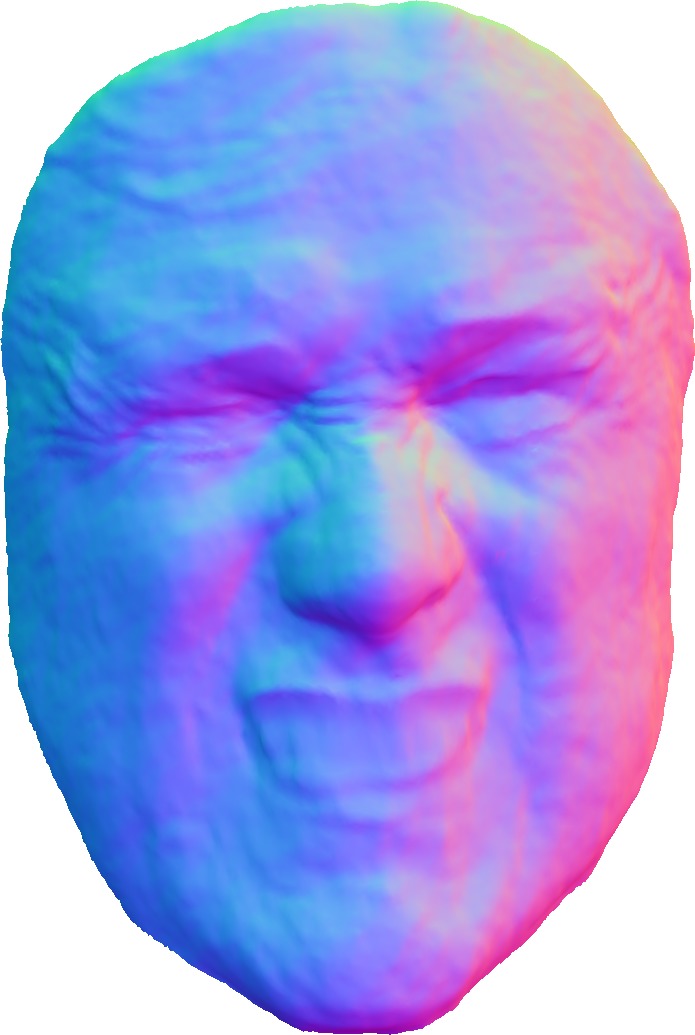}&
\includegraphics[width=0.2\columnwidth]{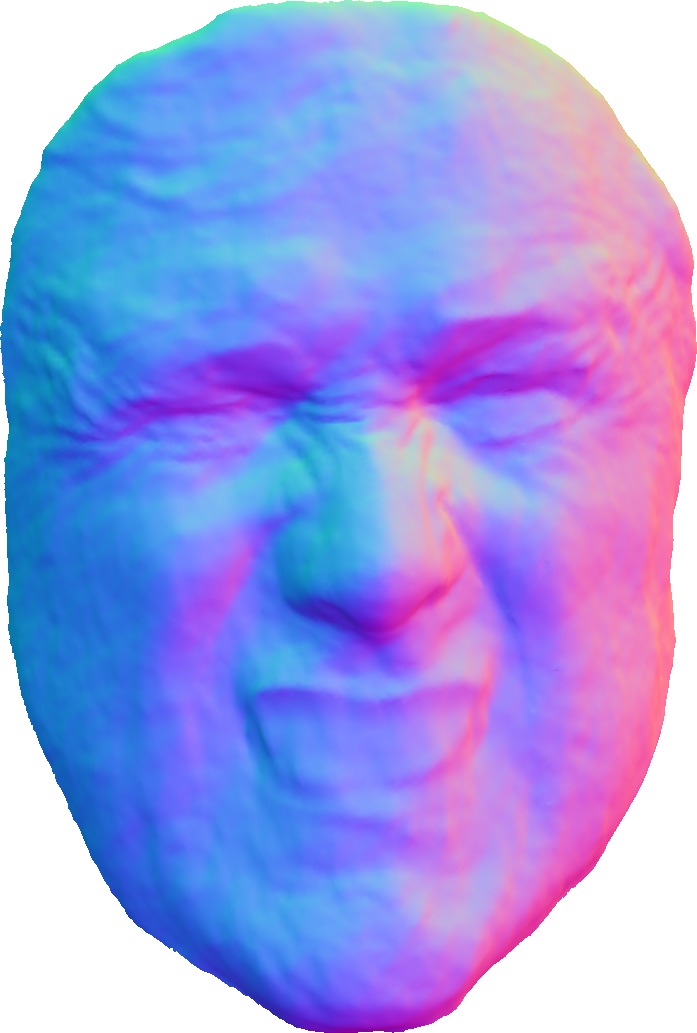}&
\includegraphics[width=0.2\columnwidth]{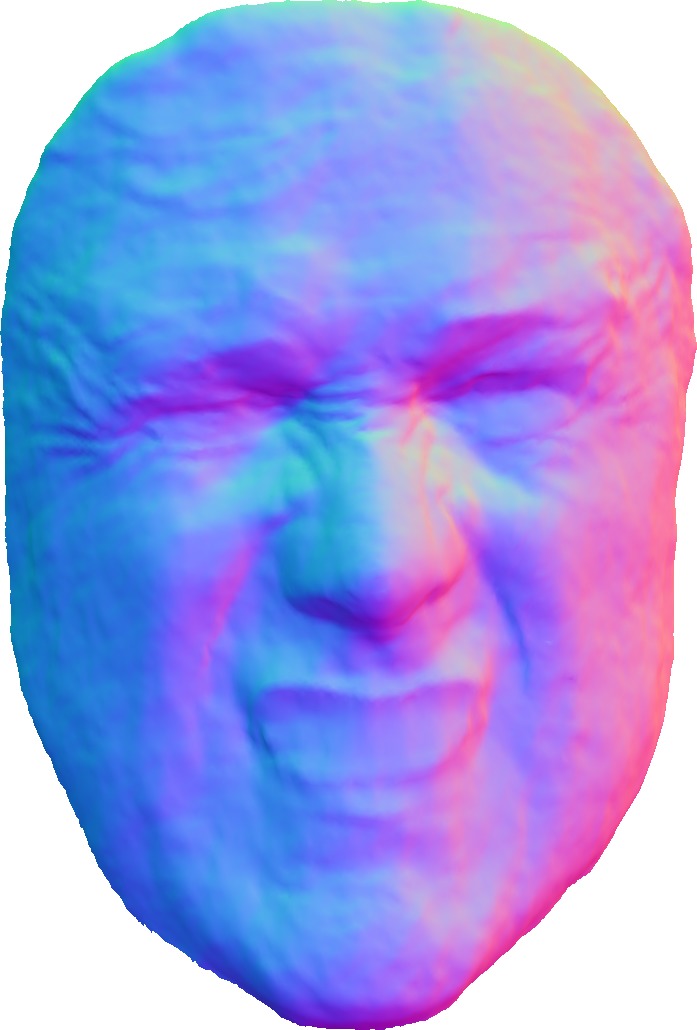}&
\includegraphics[width=0.2\columnwidth]{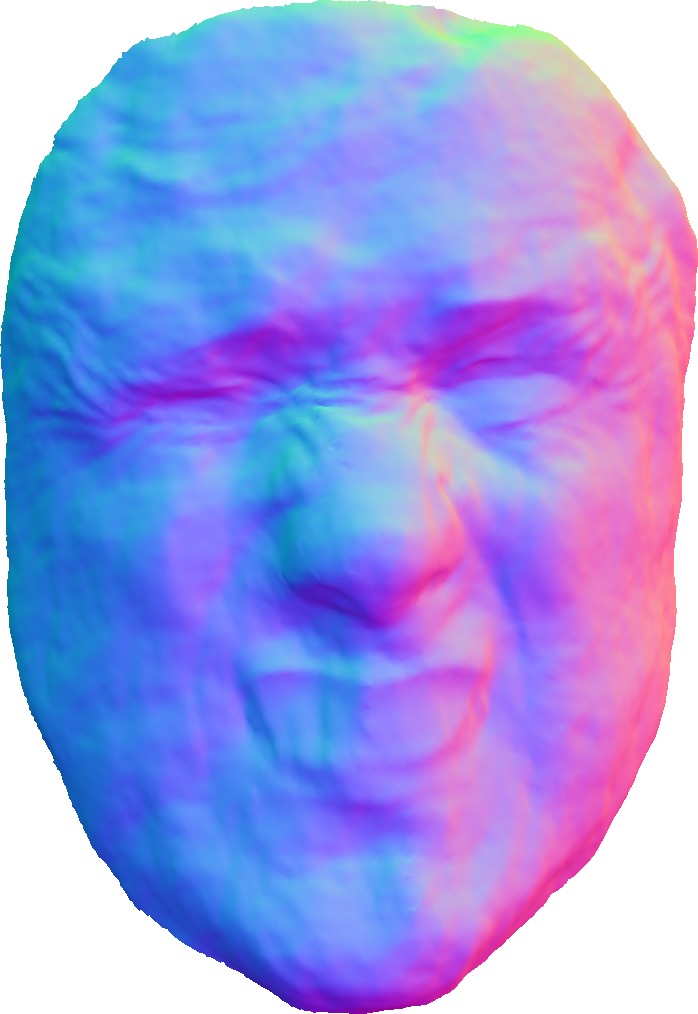}&
\includegraphics[width=0.2\columnwidth]{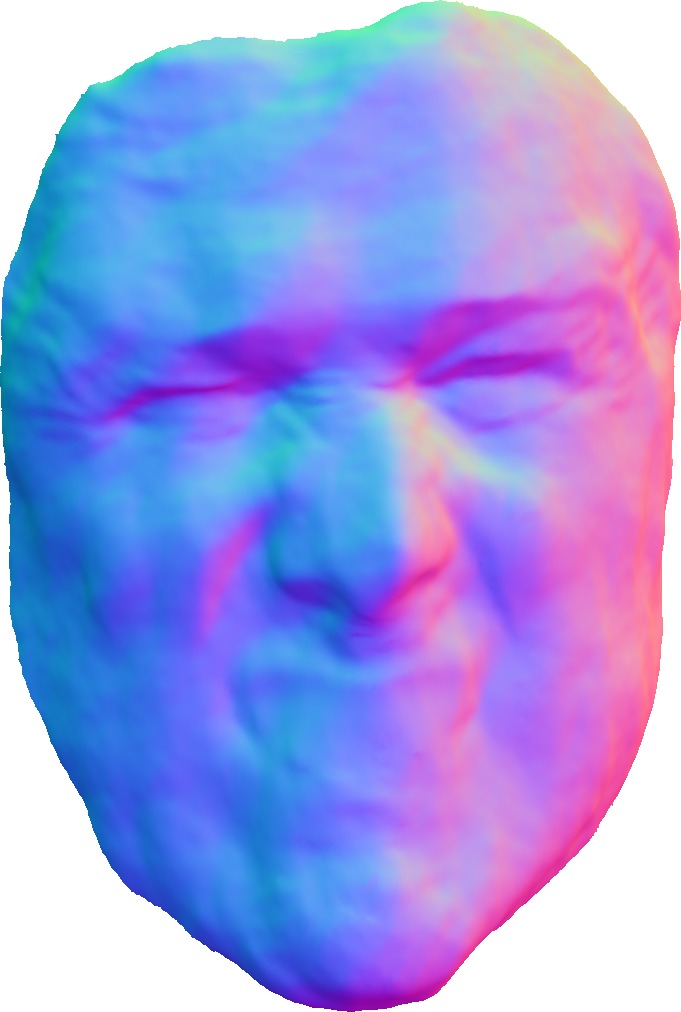}\\

\includegraphics[width=0.2\columnwidth]{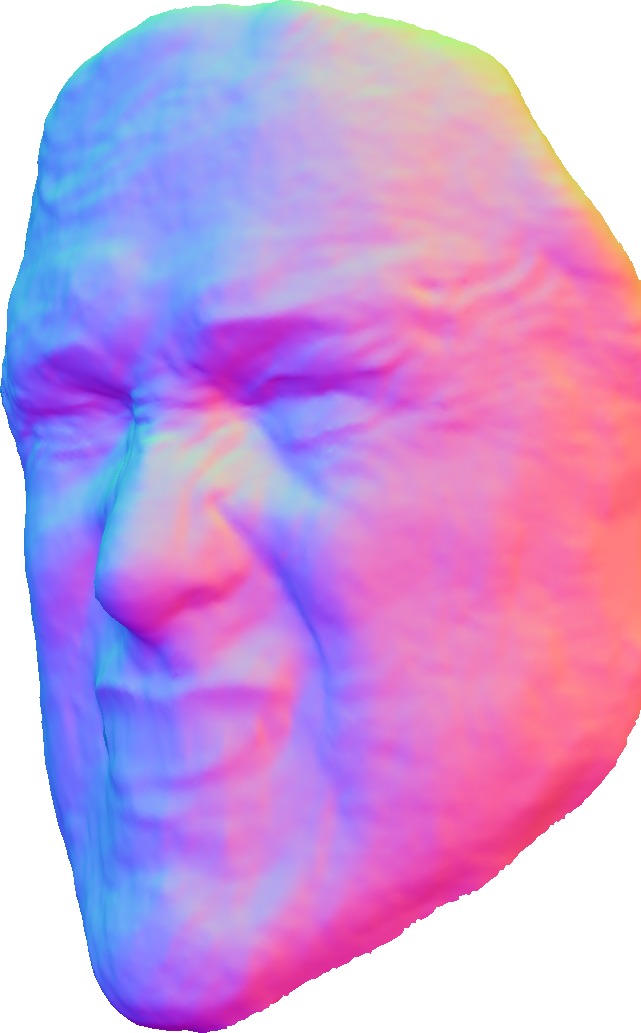}&
\includegraphics[width=0.2\columnwidth]{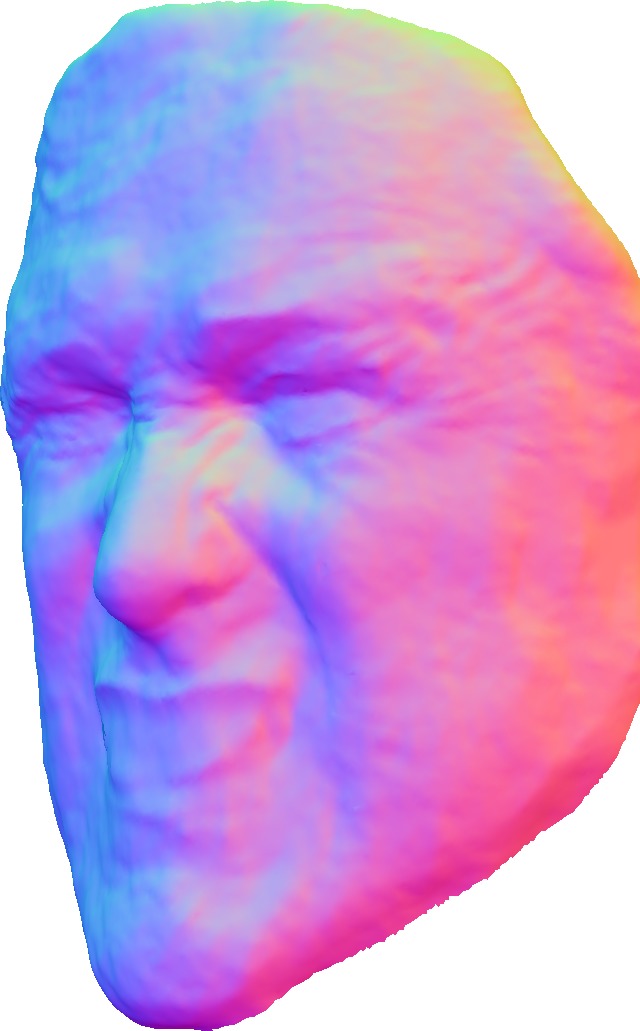}&
\includegraphics[width=0.2\columnwidth]{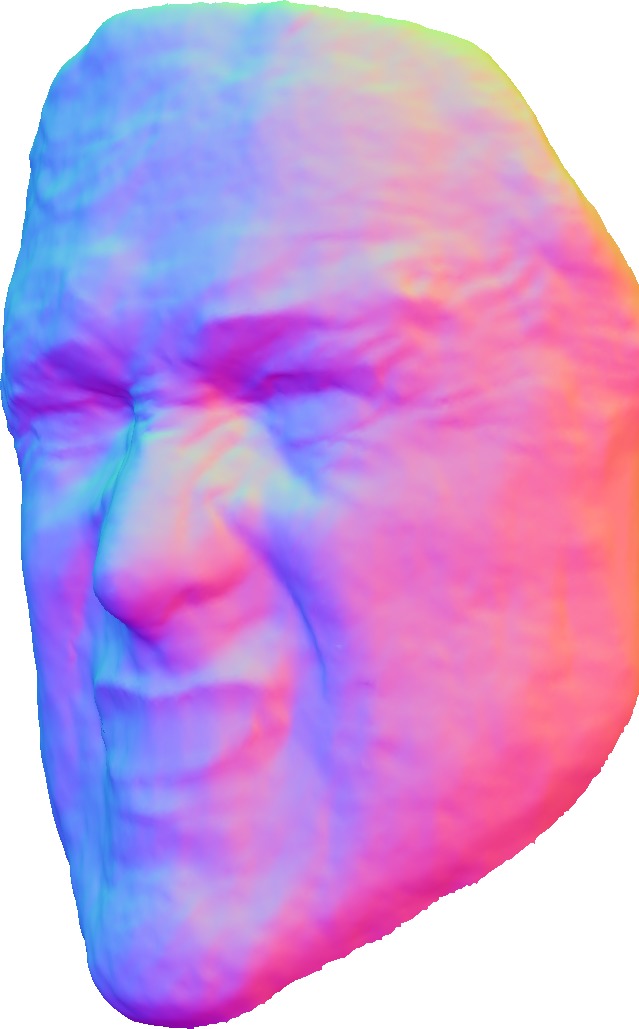}&
\includegraphics[width=0.2\columnwidth]{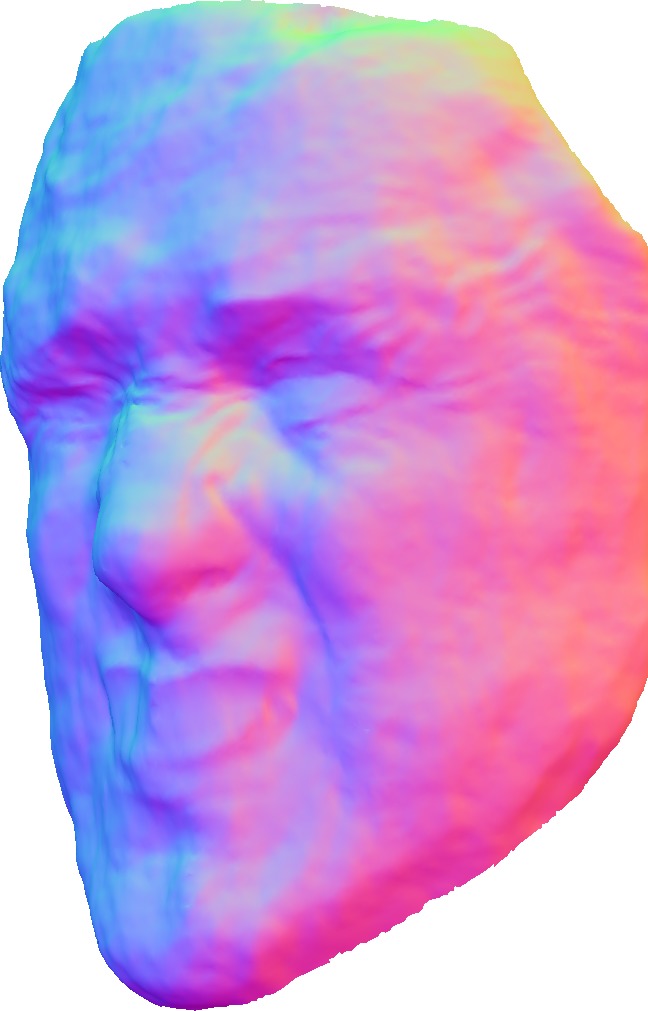}&
\includegraphics[width=0.2\columnwidth]{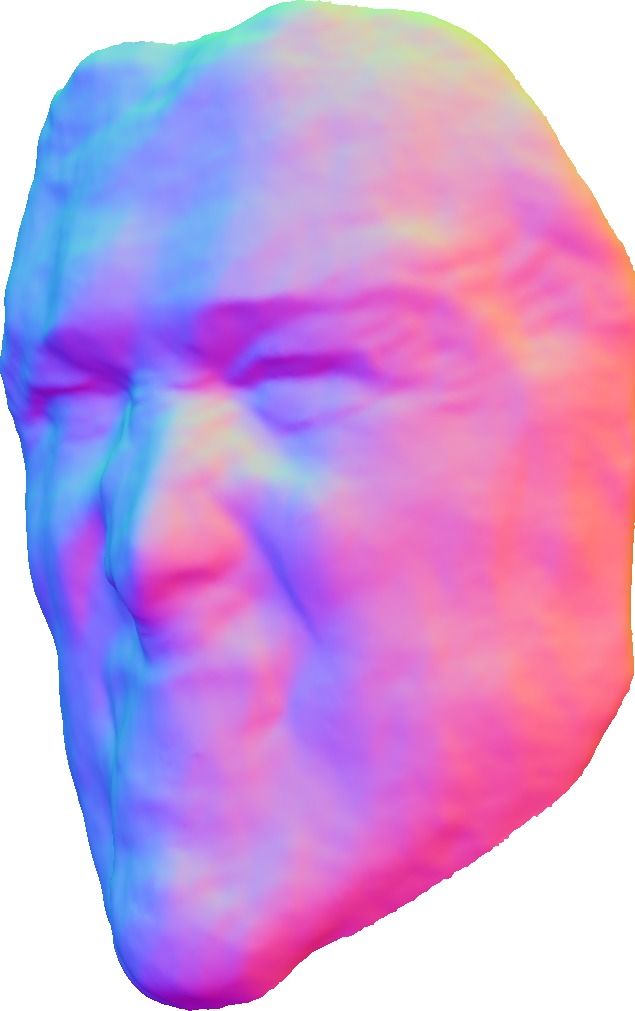}\\

  \end{tabular}}
\caption{\label{fig:noise_template}Best viewed in color: A qualitative
  comparison of our tracking results on real sequence using different templates,
  shown in the top row, with different levels of Gaussian noise. 
  The second and third row show the tracking results of one selected frame. 
}
\end{figure*}

One important question is how much do the quality of our results depend upon
high quality templates that we deform to track the object? Is our algorithm
robust to small discrepancies between the template and the ground-truth, or will
small mistakes in the template propagate leading to drifting reconstructions?
Figure \ref{fig:noise_template} and table \ref{tab:noise} show that our method
exhibits robustness to small amounts of noise with the error being as likely to
increase as to decrease as the template error increases. This can perhaps be
attributed to the first frame of the ground-truth being used as a template in
the synthetic sequence. The first frame is not a perfect example of the face at
rest, that should have zero spatial smoothness costs (i.e. total-variation as
defined in section \ref{sec:spat-smoothn-term}, and ARAP as defined in section
\ref{sec:arap-term-e_text}) associated with it, and small amount of noise have
little impact on the quality of the solution.

Note that this experiment was conducted on the \emph{real} sequence
from \cite{Valgaerts:etal:siggraph2013}. Although we generate the RMS
errors with respect to the difference between our results and geometry
given by the multi-camera reconstruction of
\cite{Valgaerts:etal:siggraph2013}, the reader should bear in mind
that the 3D reconstruction given by \cite{Valgaerts:etal:siggraph2013}
is not perfect ground-truth, just a better quality reconstruction
produced using more data than our approach has access to.

\begin{table*}
    \caption{\label{tab:noise} Template noise: The pose and 3D
      estimation of shapes does not require perfect templates and in
      fact is robust to different levels of Gaussian noise in the template
      reconstruction. See section \ref{sec:templ-qual-begint} for
      details.}
  \begin{center}
    \begin{tabular}{|c|ccccc|}
      \hline
      Template noise&$\sigma=0$&$\sigma=0.001$&$\sigma=0.005$&$\sigma=0.008$&$\sigma=0.01$ \\
      \hline    \hline
      No joint optimization
                    &3.1080&3.1339&3.1364&3.3118&3.5791\\
      \hline
      Joint optimization&3.0626&3.1081&3.1617&3.3012&3.5477\\
      \hline
    \end{tabular}
  \end{center}
\end{table*}
\subsection{Color and Depth Camera Input}
\label{sec:color-depth-camera}
To illustrate the versatility of our approach to non-rigid \sfm\ and
to show that it can also be combined with other important cues about
3D shape such as depth maps, we evaluate our method on the RGB-D face
sequence of \cite{kinect_nr:siggraph2014}. As can be seen in figure
\ref{fig:depth_face}, this is a highly challenging sequence that
contains extreme changes in appearance due to both Lambertian shading
and specular effects. As mentioned in section \ref{sec:related-work},
these effects are so strong that \citet{kinect_nr:siggraph2014} only
makes uses of color in its frame-to-frame tracking, as the effects of
shading and specularity are relatively constant over these short two
frame subsequences. However, by correctly formulating our color term
as a decomposition into variable lighting, specularities and
consistent albedo, we are able to track using a frame-to-model
approach and reliably reconstruct from RGB input only even without
using depth as a cue (see \emph{(Ours (no depth))} figure
\ref{fig:depth_face}).

Unfortunately, the images used to build the 3D templates
of~\cite{kinect_nr:siggraph2014} were no longer available at the time
of writing, and we were unable to estimate the albedo of the template
provided. Instead, we took the reconstruction in the first frame,
colored using the image data, as our template.

\cite{kinect_nr:siggraph2014} treats the output of an infra-red stereo
camera rig as a noisy depth mask and makes use of temporal
consistency, frame-to-frame color consistency and shape-based
regularization to track and denoise the output. For a fair comparison,
we also take the noisy depth map as a direct input and encourage agreement
between our reconstruction and the depth map using the additional cost
of \eqref{eq:depth}. Qualitative results can be seen in figure
\ref{fig:depth_face}.

We also evaluated the quantitative impact of using depth data on the
reconstructions conducted with synthetic sequences. Table \ref{table:additional_comparison} shows that the use of depth data gives a substantial improvement
over RGB only.

\section*{Acknowledgments}
This work was partly supported by the SecondHands project, funded from the
European Unions Horizon 2020 Research and Innovation programme under grant
agreement No 643950. Qi Liu-Yin was funded by a UCL/Microsoft Research
studentship. Chris Russell was partially supported by The Alan Turing Institute
under the EPSRC grant EP/N510129/1.


\section{Conclusion}
\label{sec:conclusion}
We have presented a principled approach to jointly reason about
non-rigid structure from motion and shape-from-shading, and provided
strong empiric evidence that it is required to avoid systematic
tracking failures, and that it significantly improves the
reconstruction quality of fine semantic details.  Although, we focused
upon the challenging problem of reconstruction from a single RGB
camera, we have also shown that joint reasoning about the motions of
deformable objects and shape-from-shading can readily be applied to
RGB-D and multi-camera based approaches, and the increased robustness
and detailed reconstructions it brings is likely to be of use to the
wider community.

\bibliographystyle{spbasic}

\bibliography{bibliography,piecewise2,mybib,humanis}

\begin{thebibliography}{28}
\providecommand{\natexlab}[1]{#1}
\providecommand{\url}[1]{{#1}}
\providecommand{\urlprefix}{URL }
\expandafter\ifx\csname urlstyle\endcsname\relax
  \providecommand{\doi}[1]{DOI~\discretionary{}{}{}#1}\else
  \providecommand{\doi}{DOI~\discretionary{}{}{}\begingroup
  \urlstyle{rm}\Url}\fi
\providecommand{\eprint}[2][]{\url{#2}}

\bibitem[{Agarwal et~al.(2012)Agarwal, Mierle et~al.}]{ceres-solver}
Agarwal S, Mierle K, et~al. (2012) Ceres solver. \url{http://ceres-solver.org}

\bibitem[{Basri and Jacobs(2003)}]{Basri_2003_PAMI}
Basri R, Jacobs DW (2003) Lambertian reflectance and linear subspaces. IEEE
  Trans Pattern Anal Mach Intell 25(2):218--233

\bibitem[{Beeler et~al.(2012)Beeler, Bradley, Zimmer, and
  Gross}]{Beeler_2012_ECCV}
Beeler T, Bradley D, Zimmer H, Gross M (2012) Improved reconstruction of
  deforming surfaces by cancelling ambient occlusion. In: 12th European
  Conference on Computer Vision, Springer Berlin Heidelberg, Berlin,
  Heidelberg, pp 30--43

\bibitem[{Campbell et~al.(2008)Campbell, Vogiatzis, Hern\'{a}ndez, and
  Cipolla}]{Campbell:etal:ECCV2008}
Campbell ND, Vogiatzis G, Hern\'{a}ndez C, Cipolla R (2008) Using multiple
  hypotheses to improve depth-maps for multi-view stereo. In: ECCV

\bibitem[{Dou et~al.(2015)Dou, Taylor, Fuchs, Fitzgibbon, and
  Izadi}]{Dou_2015_CVPR}
Dou M, Taylor J, Fuchs H, Fitzgibbon A, Izadi S (2015) 3d scanning deformable
  objects with a single rgbd sensor. In: The IEEE Conference on Computer Vision
  and Pattern Recognition (CVPR)

\bibitem[{El et~al.(2016)El, Hershkovitz, Wetzler, Rosman, Bruckstein, and
  Kimmel}]{Or-el_2016_CVPR}
El RO, Hershkovitz R, Wetzler A, Rosman G, Bruckstein A, Kimmel R (2016)
  Real-time depth refinement for specular objects. In: IEEE Conference on
  Computer Vision and Pattern Recognition (CVPR)

\bibitem[{Garrido et~al.(2013)Garrido, Valgaerts, Wu, and
  Theobalt}]{Garrido:etal:SiggraphAsia:2013}
Garrido P, Valgaerts L, Wu C, Theobalt C (2013) Reconstructing detailed dynamic
  face geometry from monocular video. In: ACM Trans. Graph. (Proc. SIGGRAPH
  Asia)

\bibitem[{Hern\'andez et~al.(2007)Hern\'andez, Vogiatzis, and
  Cipolla}]{hernandez07}
Hern\'andez C, Vogiatzis G, Cipolla R (2007) Probabilistic visibility for
  multi-view stereo. In: CVPR

\bibitem[{Kemelmacher-Shlizerman and Seitz(2011)}]{kemelmacher2011face}
Kemelmacher-Shlizerman I, Seitz SM (2011) Face reconstruction in the wild. In:
  Computer Vision (ICCV), 2011 IEEE International Conference on, IEEE, pp
  1746--1753

\bibitem[{Klein and Murray(2007)}]{Klein:Murray:ISMAR2007}
Klein G, Murray D (2007) Parallel tracking and mapping for small {AR}
  workspaces. In: ISMAR

\bibitem[{Malti et~al.(2012)Malti, Bartoli, and
  Collins}]{Malti:etal:IPCAI:2012}
Malti A, Bartoli A, Collins T (2012) Template-based conformal
  shape-from-motion-and-shading for laparoscopy. In: Information Processing in
  Computer-Assisted Interventions

\bibitem[{Newcombe et~al.(2011)Newcombe, Lovegrove, and
  Davison}]{Newcombe:etal:ICCV2011}
Newcombe R, Lovegrove S, Davison A (2011) {DTAM: Dense Tracking and Mapping in
  Real-Time}. In: ICCV

\bibitem[{Newcombe et~al.(2015)Newcombe, Fox, and
  Seitz}]{Newcombe:etal:CVPR2015}
Newcombe R, Fox D, Seitz S (2015) Dynamicfusion: Reconstruction and tracking of
  non-rigid scenes in real-time. In: CVPR

\bibitem[{Nishino et~al.(2001)Nishino, Zhang, and
  Ikeuchi}]{nishino2001determining}
Nishino K, Zhang Z, Ikeuchi K (2001) Determining reflectance parameters and
  illumination distribution from a sparse set of images for view-dependent
  image synthesis. In: Computer Vision, 2001. ICCV 2001. Proceedings. Eighth
  IEEE International Conference on, IEEE, vol~1, pp 599--606

\bibitem[{Or~El et~al.(2015)Or~El, Rosman, Wetzler, Kimmel, and
  Bruckstein}]{Or-el_2015_CVPR}
Or~El R, Rosman G, Wetzler A, Kimmel R, Bruckstein AM (2015) Rgbd-fusion:
  Real-time high precision depth recovery. In: The IEEE Conference on Computer
  Vision and Pattern Recognition (CVPR)

\bibitem[{Sorkine and Alexa(2007)}]{Sorkine:ARAP}
Sorkine O, Alexa M (2007) As-rigid-as-possible surface modeling. In:
  Proceedings of the Fifth Eurographics Symposium on Geometry Processing, SGP
  '07

\bibitem[{Sundaram et~al.(2010)Sundaram, Brox, and
  Keutzer}]{Sundaram:etal:ECCV2010}
Sundaram N, Brox T, Keutzer K (2010) Dense point trajectories by
  gpu-accelerated large displacement optical flow. In: ECCV

\bibitem[{Suwajanakorn et~al.(2014)Suwajanakorn, Kemelmacher-Shlizerman, and
  Seitz}]{total_moving_face}
Suwajanakorn S, Kemelmacher-Shlizerman I, Seitz SM (2014) Total moving face
  reconstruction. In: ECCV

\bibitem[{Valgaerts et~al.(2012)Valgaerts, Wu, Bruhn, Seidel, and
  Theobalt}]{Valgaerts_2012_SIGGRAPH}
Valgaerts L, Wu C, Bruhn A, Seidel HP, Theobalt C (2012) Lightweight binocular
  facial performance capture under uncontrolled lighting. In: ACM Transactions
  on Graphics (Proceedings of SIGGRAPH Asia 2012), vol~31, pp 187:1--187:11

\bibitem[{Valgaerts et~al.(2013)Valgaerts, Bruhn, Seidel, and
  Theobalt}]{Valgaerts:etal:siggraph2013}
Valgaerts L, Bruhn A, Seidel HP, Theobalt C (2013) Lightweight binocular facial
  performance capture under uncontrolled lighting. SIGGRAPH

\bibitem[{Varol et~al.(2012)Varol, Shaji, Salzmann, and
  Fua}]{varol2012monocular}
Varol A, Shaji A, Salzmann M, Fua P (2012) Monocular 3d reconstruction of
  locally textured surfaces. Pattern Analysis and Machine Intelligence, IEEE
  Transactions on 34(6):1118--1130

\bibitem[{Vogiatzis et~al.(2007)Vogiatzis, Hern\'andez, Torr, and
  Cipolla.}]{vogiatzis07pami}
Vogiatzis G, Hern\'andez C, Torr P, Cipolla R (2007) Multi-view stereo via
  volumetric graph-cuts and occlusion robust photo-consistency. PAMI

\bibitem[{Wood et~al.(2000)Wood, Azuma, Aldinger, Curless, Duchamp, Salesin,
  and Stuetzle}]{wood2000surface}
Wood DN, Azuma DI, Aldinger K, Curless B, Duchamp T, Salesin DH, Stuetzle W
  (2000) Surface light fields for 3d photography. In: Proceedings of the 27th
  annual conference on Computer graphics and interactive techniques, ACM
  Press/Addison-Wesley Publishing Co., pp 287--296

\bibitem[{Wu(2011)}]{wu2011visualsfm}
Wu C (2011) Visualsfm: A visual structure from motion system.
  \url{http://ccwu.me/vsfm/}

\bibitem[{Xu and Roy-Chowdhury(2005)}]{Xu_2005_ICCV}
Xu Y, Roy-Chowdhury AK (2005) Integrating the effects of motion, illumination
  and structure in video sequences. In: Computer Vision, 2005. ICCV 2005. Tenth
  IEEE International Conference on, IEEE, vol~2, pp 1675--1682

\bibitem[{Yu et~al.(2013)Yu, Yeung, Tai, and Lin}]{yu2013shading}
Yu LF, Yeung SK, Tai YW, Lin S (2013) Shading-based shape refinement of rgb-d
  images. In: Proceedings of the IEEE Conference on Computer Vision and Pattern
  Recognition, pp 1415--1422

\bibitem[{Yu et~al.(2015)Yu, Russell, Campbell, and Agapito}]{Yu_2015_ICCV}
Yu R, Russell C, Campbell N, Agapito L (2015) Direct, dense, and deformable:
  Non-rigid 3d reconstruction from rgb video. ICCV

\bibitem[{Zollhofer et~al.(2014)Zollhofer, Niessner, Izadi, Rehmann, Zach,
  Fisher, Wu, Fitzgibbon, Loop, Theobalt, and
  Stamminger}]{kinect_nr:siggraph2014}
Zollhofer M, Niessner M, Izadi S, Rehmann C, Zach C, Fisher M, Wu C, Fitzgibbon
  A, Loop C, Theobalt C, Stamminger M (2014) Real-time non-rigid reconstruction
  using an rgb-d camera. SIGGRAPH

\end{thebibliography}

%
%

\end{document}